\documentclass{article} 
\usepackage{iclr2026_conference,times}

\usepackage{amsmath,amsfonts,bm}

\def\eqref#1{equation~\ref{#1}}

\def\1{\bm{1}}

\DeclareMathAlphabet{\mathsfit}{\encodingdefault}{\sfdefault}{m}{sl}
\SetMathAlphabet{\mathsfit}{bold}{\encodingdefault}{\sfdefault}{bx}{n}

\usepackage{mathrsfs}
\usepackage{hyperref}
\usepackage{url}
\usepackage{amsthm}
\usepackage{algorithm}
\usepackage{algpseudocode}

\usepackage{graphicx}  

\graphicspath{{./figs/}}

\usepackage{booktabs}  

\usepackage{multirow}  
\usepackage{array}     
\usepackage{pifont}    
\usepackage{caption}   
\usepackage{adjustbox}
\usepackage{subcaption}
\usepackage{hyperref}
\usepackage{booktabs}
\usepackage[table]{xcolor}
\usepackage{tabularx}
\usepackage{enumerate}

\let\vec=\boldsymbol
\usepackage[capitalize,noabbrev]{cleveref}

\usepackage{tikz}
\usetikzlibrary{matrix,calc,positioning,fit,backgrounds,shadows.blur,arrows.meta}
\usepackage{lipsum}
\usetikzlibrary{matrix,calc,positioning,fit,backgrounds,shadows.blur,arrows.meta}
\usepackage{amsmath}
\usepackage[dvipsnames]{xcolor}
\newcommand{\learn}{\textsuperscript{\textcolor{orange!80!black}{\scriptsize L}}}
\newcommand{\pre}{\textsuperscript{\textcolor{blue!70!black}{\scriptsize PC}}}
\definecolor{sensorcolor}{RGB}{65,105,225}
\definecolor{framecolor}{RGB}{220,20,60}
\definecolor{shapecolor}{RGB}{34,139,34}
\definecolor{navcolor}{RGB}{255,140,0}
\definecolor{band}{RGB}{246,246,246}
\definecolor{Crimson}{rgb}{0.86, 0.08, 0.24}

\tikzset{
  mod/.style   ={rectangle,rounded corners=2pt,draw,very thick,align=center,
                 minimum width=4.2cm,minimum height=2.0cm,fill=#1!14},
  nav/.style   ={rectangle,rounded corners=2pt,draw,very thick,align=center,
                 minimum width=14.0cm,minimum height=2.0cm,fill=#1!14},
  io/.style    ={circle,draw,thick,inner sep=0pt,minimum size=2.8mm,fill=white},
  qarr/.style  ={thick,dashed,-{Stealth[length=2.2mm]}},
  rarr/.style  ={thick,-{Stealth[length=2.2mm]}},
  chip/.style  ={rectangle,rounded corners=1pt,draw,thin,inner sep=1pt,font=\tiny\bfseries,fill=white},
  dsicon/.style={rectangle,draw,thin,fill=white,minimum width=4.5mm,minimum height=4.5mm},
  badge/.style ={circle,draw,thin,fill=white,minimum size=5mm,font=\tiny\bfseries},
  tiny/.style  ={font=\scriptsize},
  xtiny/.style ={font=\tiny}
}

\newtheorem{theorem}{Theorem}[section]     
\newtheorem{lemma}[theorem]{Lemma}         

\newtheorem{proposition}[theorem]{Proposition}
\newtheorem{corollary}[theorem]{Corollary}

\newtheorem{assumption}{Assumption}[section]
\newtheorem*{remark}{Remark}

\title{GRL-SNAM: Geometric Reinforcement\\ Learning with Path Differential Hamiltonians for Simultaneous Navigation and Mapping\\ in Unknown Environments}

\author{
Aditya Sai Ellendula$^{1}$, Yi Wang$^{2}$, Minh Nguyen$^{3}$, Chandrajit Bajaj$^{1,2}$ \\
$^{1}$ Department of Computer Science, University of Texas at Austin, Austin, TX 78712, USA \\
$^{2}$ Oden Institute, University of Texas at Austin, Austin, TX 78712, USA \\
$^{3}$ Department of Mathematics, University of Texas at Austin, Austin, TX 78712, USA \\
\texttt{\{adityase,panzer.wy,minhpnguyen,bajaj\}@utexas.edu} \\
}

\iclrfinalcopy 
\begin{document}
\usetikzlibrary{matrix}

\maketitle

\begin{abstract}
    We present GRL-SNAM, a geometric reinforcement learning framework for Simultaneous Navigation and Mapping~(SNAM) in unknown environments. A SNAM problem is challenging as it needs to design hierarchical or joint policies of multiple agents that control the movement of a real-life robot towards the goal in mapless environment, i.e. an environment where the map of the environment is not available apriori, and needs to be acquired through sensors. The sensors are invoked from the path learner, i.e. navigator, through active query responses to sensory agents, and along the motion path. GRL-SNAM differs from preemptive navigation algorithms and other reinforcement learning methods by relying exclusively on local sensory observations without constructing a global map. Our approach formulates path navigation and mapping  as a dynamic shortest path search and discovery process using controlled Hamiltonian optimization: sensory inputs are translated into local energy landscapes that encode reachability, obstacle barriers, and deformation constraints, while policies for sensing, planning, and reconfiguration evolve stagewise via updating Hamiltonians. A reduced Hamiltonian serves as an adaptive score function, updating kinetic/potential terms, embedding barrier constraints, and continuously refining trajectories as new local information arrives. We evaluate GRL-SNAM on two different 2D navigation tasks. To show our geometric RL policies naturally decomposes and bring hierchacy, we build a hyperelastic robot that learns to squeeze through narrow gaps, detour around obstacles, and generalize to unseen environments; To show GRL--SNAM is generalizable to indoor scene layout, we build a point-nav system in an unseen indoor maze layouts. Comparing against \emph{local reactive} baselines (PF, CBF, staged DWA, staged PPO) and \emph{global} policy learning references (A$^{\star}$, PPO, SAC) under identical stagewise sensing constraints, GRL-SNAM maintains path quality while using the minimal map coverage. It preserves clearance, generalizes to unseen layouts, and demonstrates that Geometric RL learning via updating Hamiltonians enables high-quality navigation through \emph{minimal exploration} via local energy refinement rather than extensive global mapping. The code is publicly available on \href{https://github.com/CVC-Lab/GRL-SNAM}{Github}.
\end{abstract}

\section{Introduction}
\label{sec:intro}

Reinforcement learning (RL) has achieved impressive results in high-dimensional control, yet its application to real-world continuous navigation remains fundamentally limited. Modern deep RL methods such as SAC~\citep{haarnoja2018sac} and PPO~\citep{schulman2017ppo} typically require millions of interactions to learn a single robust policy, and their performance degrades sharply under even mild distribution shift. These challenges are amplified in simultaneous navigation and mapping (SNAM), where the agent must traverse unknown environments while building and updating spatial representations online. In this setting, long-horizon reasoning, multi-scale decision making, and the need for safe, sample-efficient adaptation often overwhelm existing methods.

\paragraph{Why many RL fails at SNAM.}
Three structural limitations are particularly acute in navigation:
(i) \emph{Sample efficiency}: continuous control demands rich exploration and fine-grained motor control; model-free RL compensates with massive data collection and unstable credit assignment.
(ii) \emph{Generalization}: black-box policies memorize training environments instead of discovering transferable navigation principles, leading to brittle behavior under new obstacle layouts or sensing conditions.
(iii) \emph{Temporal scale separation}: navigation couples fast obstacle avoidance, mid-range local planning, and slow global goal pursuit. Hierarchical RL~\citep{sutton1998between,vezhnevets2017feudal} introduces additional levels but still relies on monolithic value-based objectives and carefully engineered reward decompositions.

\begin{figure*}[htbp]
\centering
\setlength{\tabcolsep}{2pt}
\small\sffamily

\begin{tabular}{@{}c ccc@{}}
  & \textbf{Timestep A} & \textbf{Timestep B} & \textbf{Timestep C} \\[4pt]
  
  \rotatebox{90}{\parbox{20mm}{\centering\textcolor{red!65!black}{\textbf{Baseline}}}} &
  \includegraphics[width=0.29\linewidth]{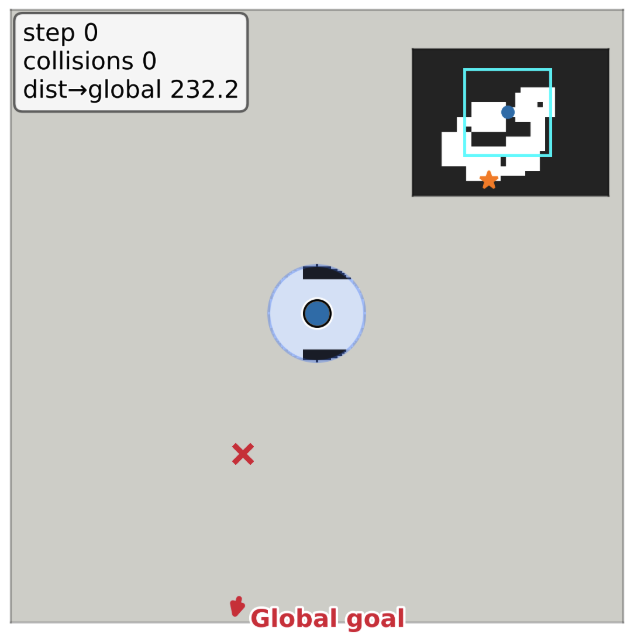} &
  \includegraphics[width=0.29\linewidth]{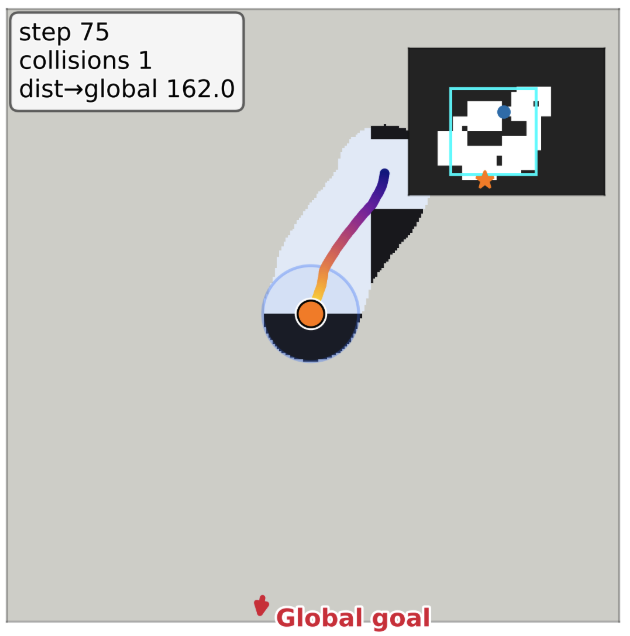} &
  \includegraphics[width=0.29\linewidth]{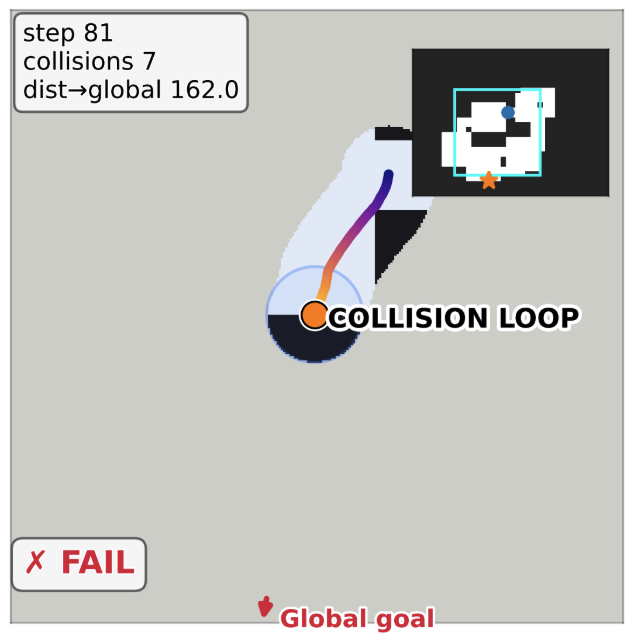} \\[4pt]
  
  \rotatebox{90}{\parbox{20mm}{\centering\textcolor{green!50!black}{\textbf{Ours}}}} &
  \includegraphics[width=0.29\linewidth]{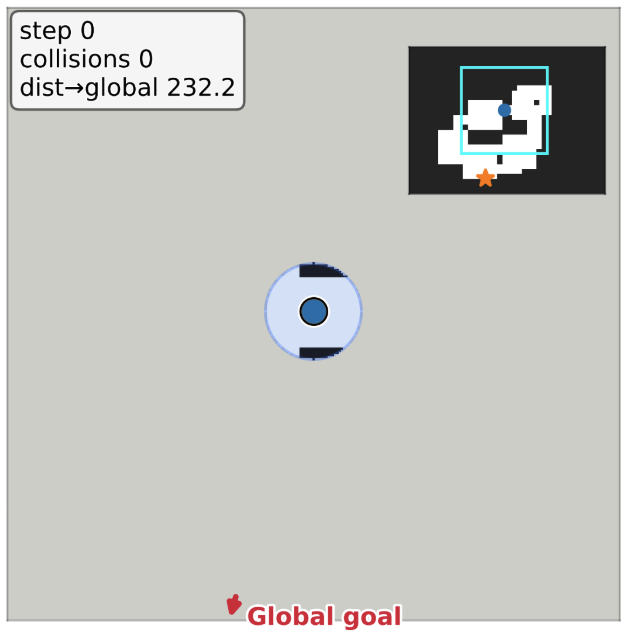} &
  \includegraphics[width=0.29\linewidth]{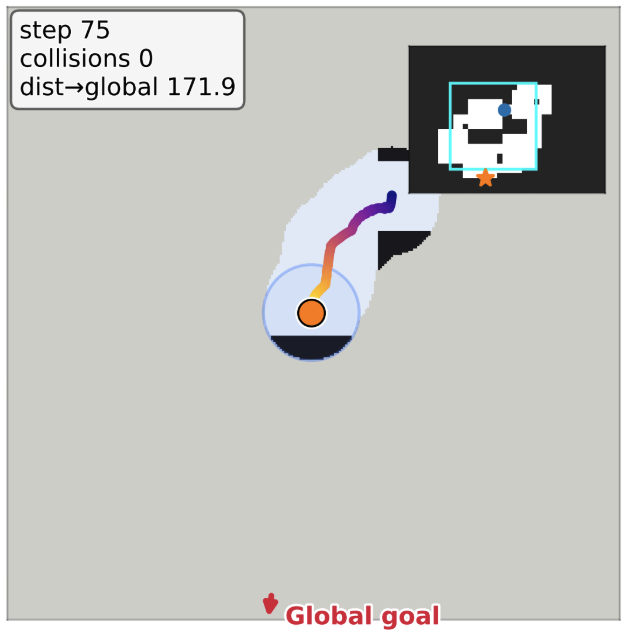} &
  \includegraphics[width=0.29\linewidth]{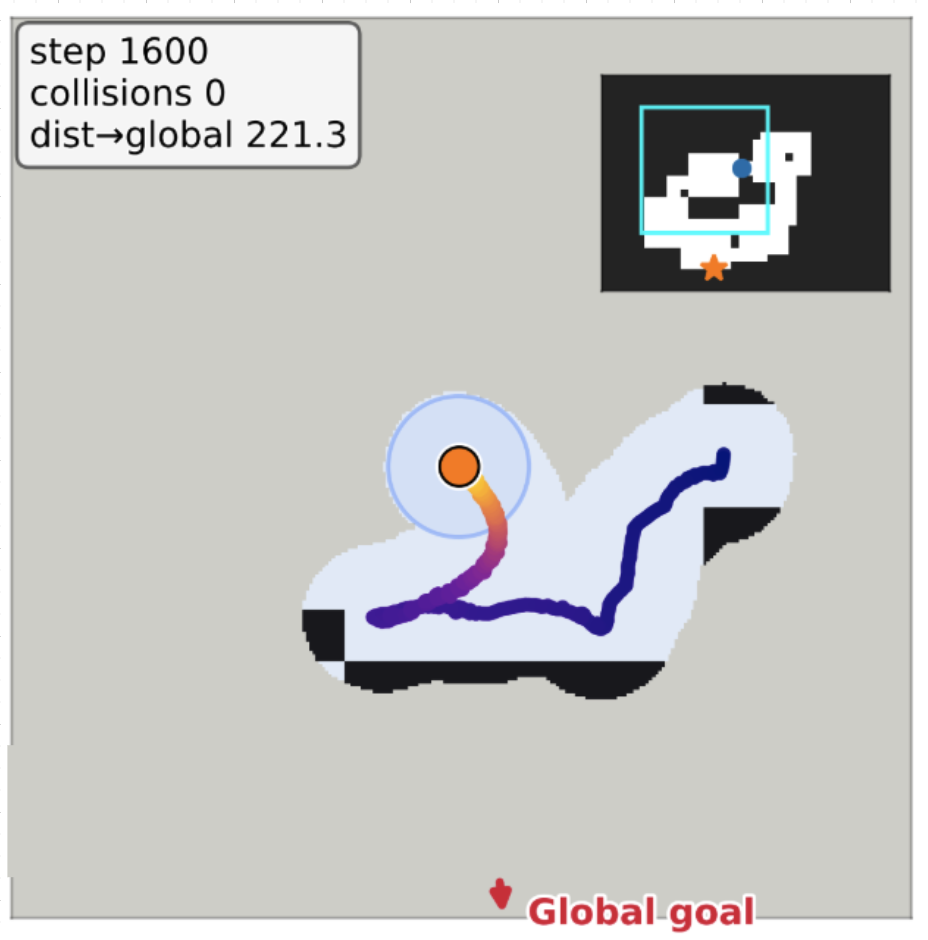} \\
\end{tabular}

\vspace{2mm}
{\small
\textcolor{red!65!black}{$\boldsymbol{\times}$~Frozen RL: stuck or collides}
\hfill
\textcolor{green!50!black}{$\boldsymbol{\checkmark}$~Online adaptation: reaches goal}}

\caption{\textbf{Fixed policy vs.\ online adaptation in dungeon navigation.} We display global time step, collision count, and remaining goal distance on top--left of a navigation process under a mapless environment; the inset shows the local sensed view used for decision-making. Top: a fixed RL controller executes a fixed behavior and fails by colliding / entering a collision loop as the local obstacle geometry changes.
Bottom: our method performs \emph{online correction} by adjusting its motion field using only short-horizon feedback (clearance, distance-to-goal, and speed) to avoid obstacles and keep making progress, ultimately reaching the global goal.
}

\label{fig:teaser_online_vs_rl}
\end{figure*}

These issues are magnified in SNAM. Recent systems such as Simultaneous
Goal Localization and Mapping (SGoLAM)~\citep{sgolam2021}, Cognitive Mapping and Planning (CMP)~\citep{gupta2017cognitive}, and Continual Learning of Simultaneous Localization and Mapping (CL-SLAM)~\citep{clslam2022} tightly couple mapping and policy learning, often by constructing increasingly detailed metric maps that are then consumed by learned or differentiable planners. While effective, they implicitly optimize for \emph{better maps}, not necessarily \emph{minimal, task-focused exploration}. In contrast, our objective is to \emph{reach the goal along high-quality, well-weighted paths while mapping as little of the unknown environment as possible}: as new local information is revealed, the navigator should differentially refine the path and its safety margins, rather than rebuilding global plans from scratch.

\paragraph{From unstructured policies to geometric structure.}
We view these limitations as symptoms of treating navigation as structureless optimization. Real robots are constrained by physics: conservation laws, kinematics, actuator limits, and contact constraints. Policies that ignore this structure are difficult to train, hard to interpret, and unstable over long rollouts. A geometric, physics-grounded formulation offers three advantages:
(i) \emph{Physical realizability}: dynamics remain compatible with real actuators and constraints;
(ii) \emph{Stability}: symplectic structure preserves invariants (e.g., energy, momentum) and mitigates long-horizon drift;
(iii) \emph{Compositionality}: complex behavior emerges from composing interpretable energy terms (goal attraction, collision barriers, deformation costs) that can be reweighted and adapted without relearning from scratch.

\begin{figure*}[htbp]
\centering
\begin{tikzpicture}[
  font=\small,
  panel/.style={
    draw=black!15,
    rounded corners=3pt,
    line width=0.6pt,
    inner sep=0pt,
  },
  lab/.style={
    font=\sffamily\scriptsize,
    text=white,
    fill=black!70,
    rounded corners=2pt,
    inner sep=3pt,
  },
  timestep/.style={
    font=\sffamily\scriptsize,
    text=teal!70!black,
    fill=teal!10,
    rounded corners=2pt,
    inner sep=2.5pt,
  },
  stagebadge/.style={
    font=\sffamily\small\bfseries,
    text=white,
    fill=teal!70!black,
    circle,
    minimum size=16pt,
    inner sep=0pt,
  },
  goalstyle/.style={fill=green!50!black},
  summarystyle/.style={fill=black!50},
]

\def\W{0.30\textwidth}  
\def\H{6pt}
\def\V{8pt}

\node[panel] (p1) {\includegraphics[width=\W]{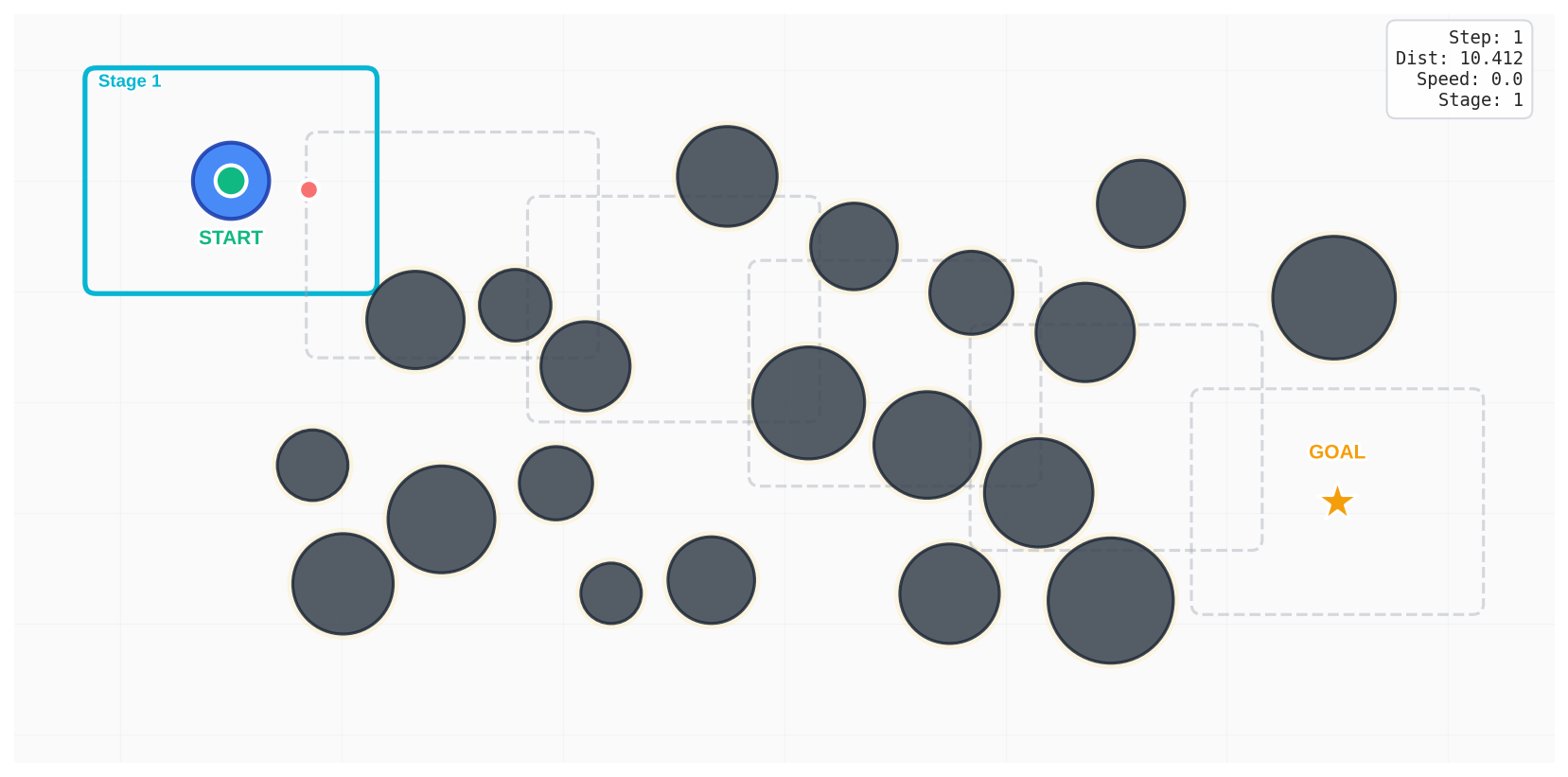}};
\node[panel, right=\H of p1] (p2) {\includegraphics[width=\W]{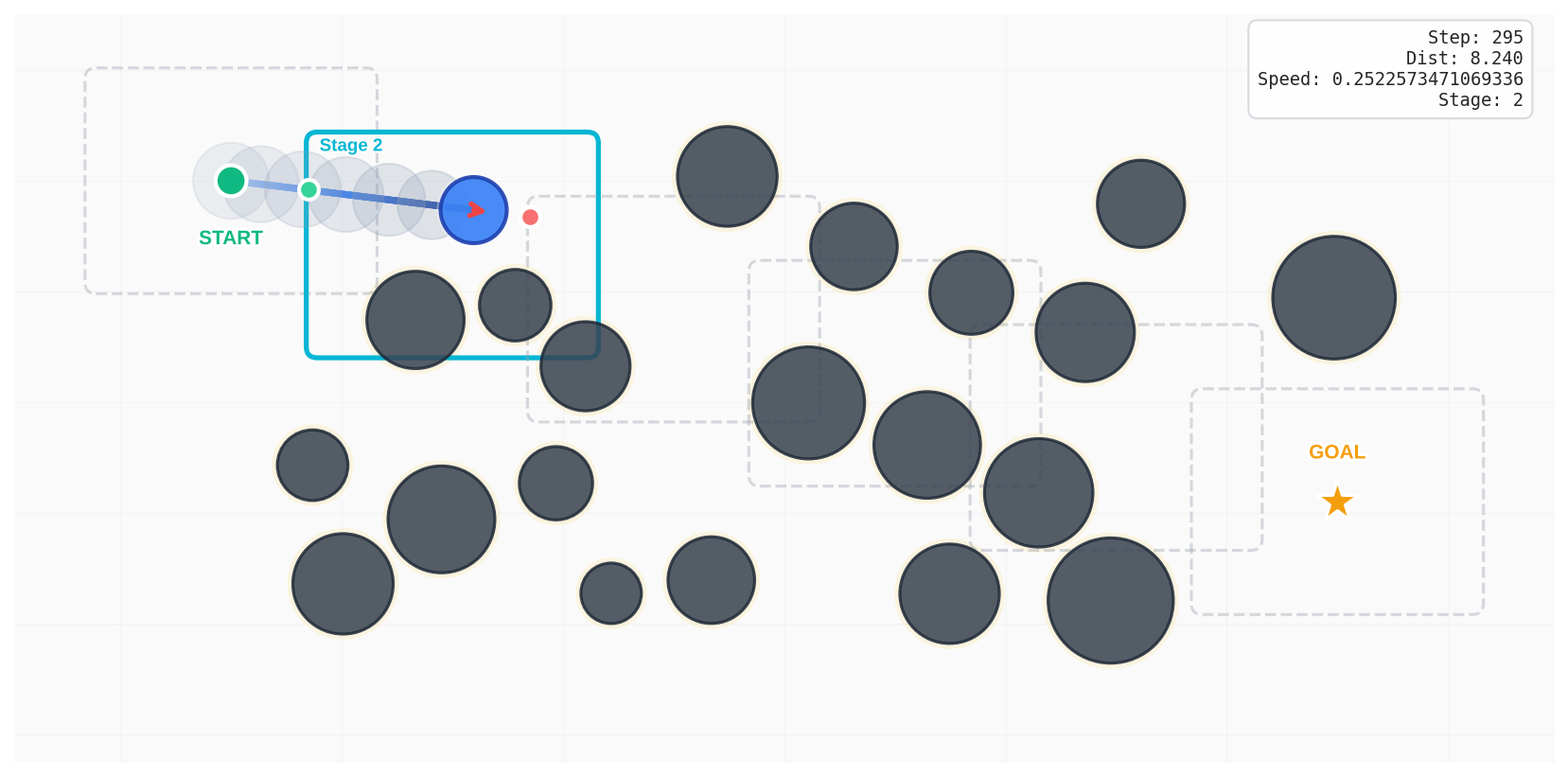}};
\node[panel, right=\H of p2] (p3) {\includegraphics[width=\W]{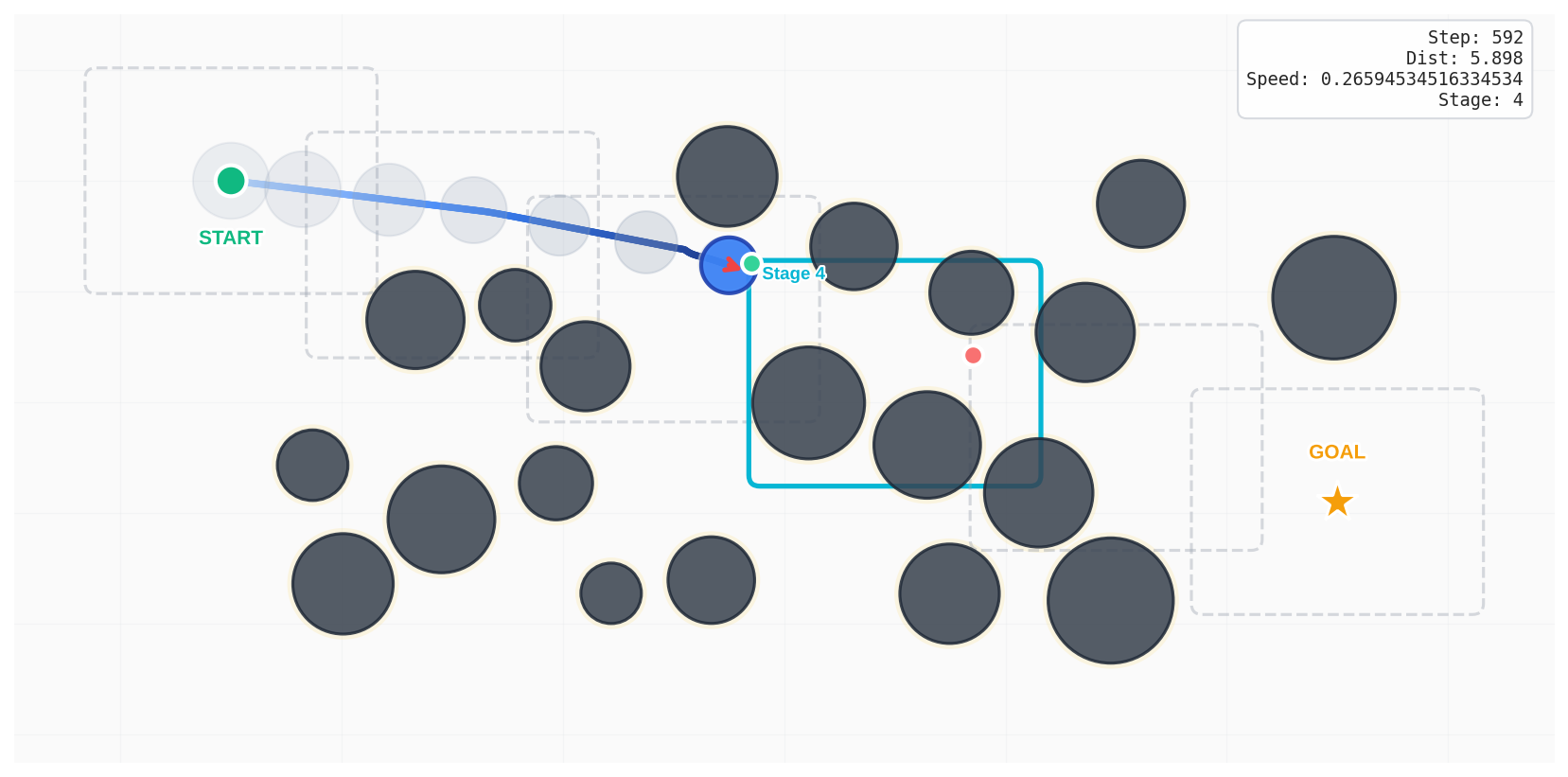}};

\node[panel, below=\V of p1] (p4) {\includegraphics[width=\W]{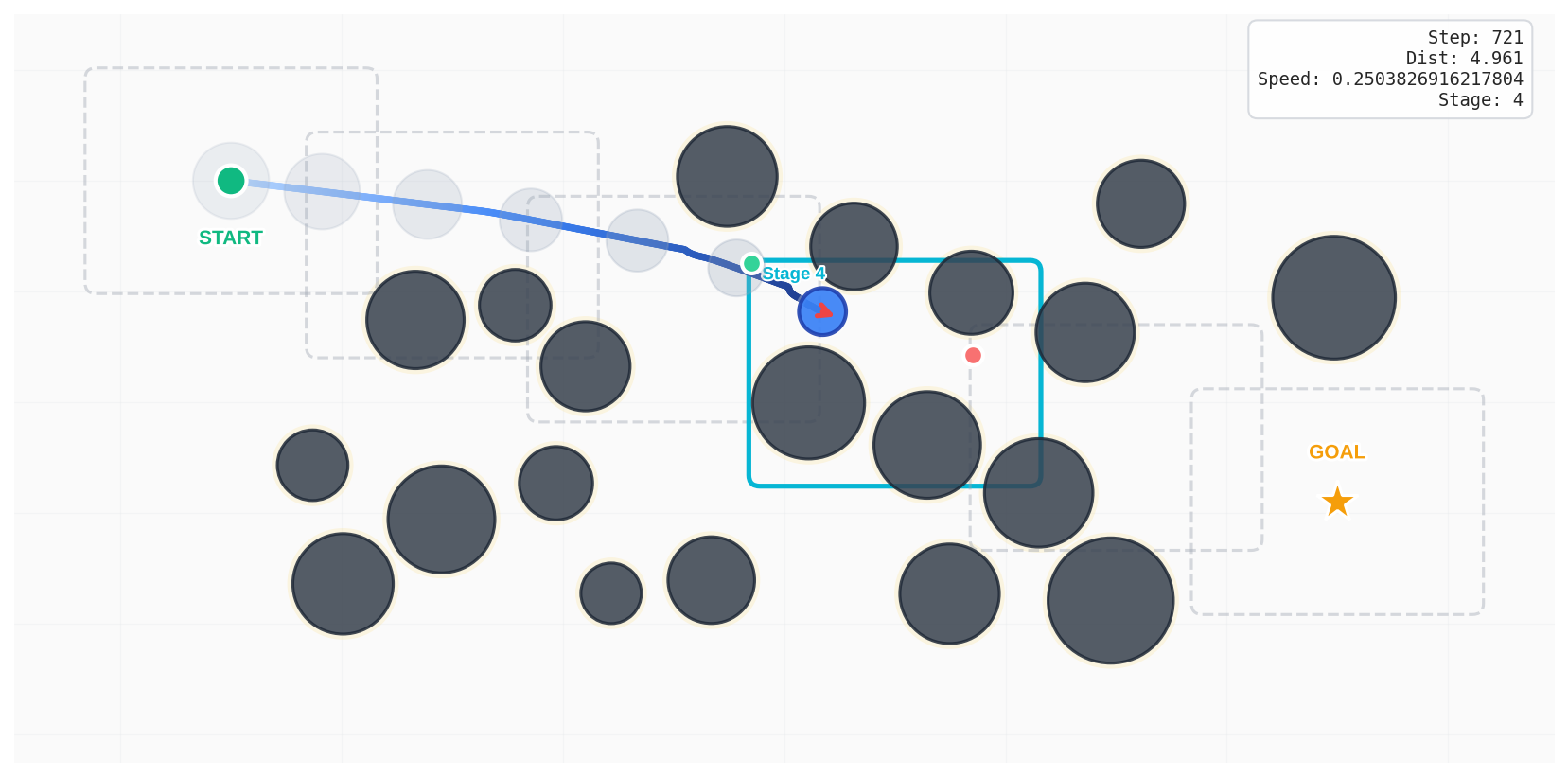}};
\node[panel, right=\H of p4] (p5) {\includegraphics[width=\W]{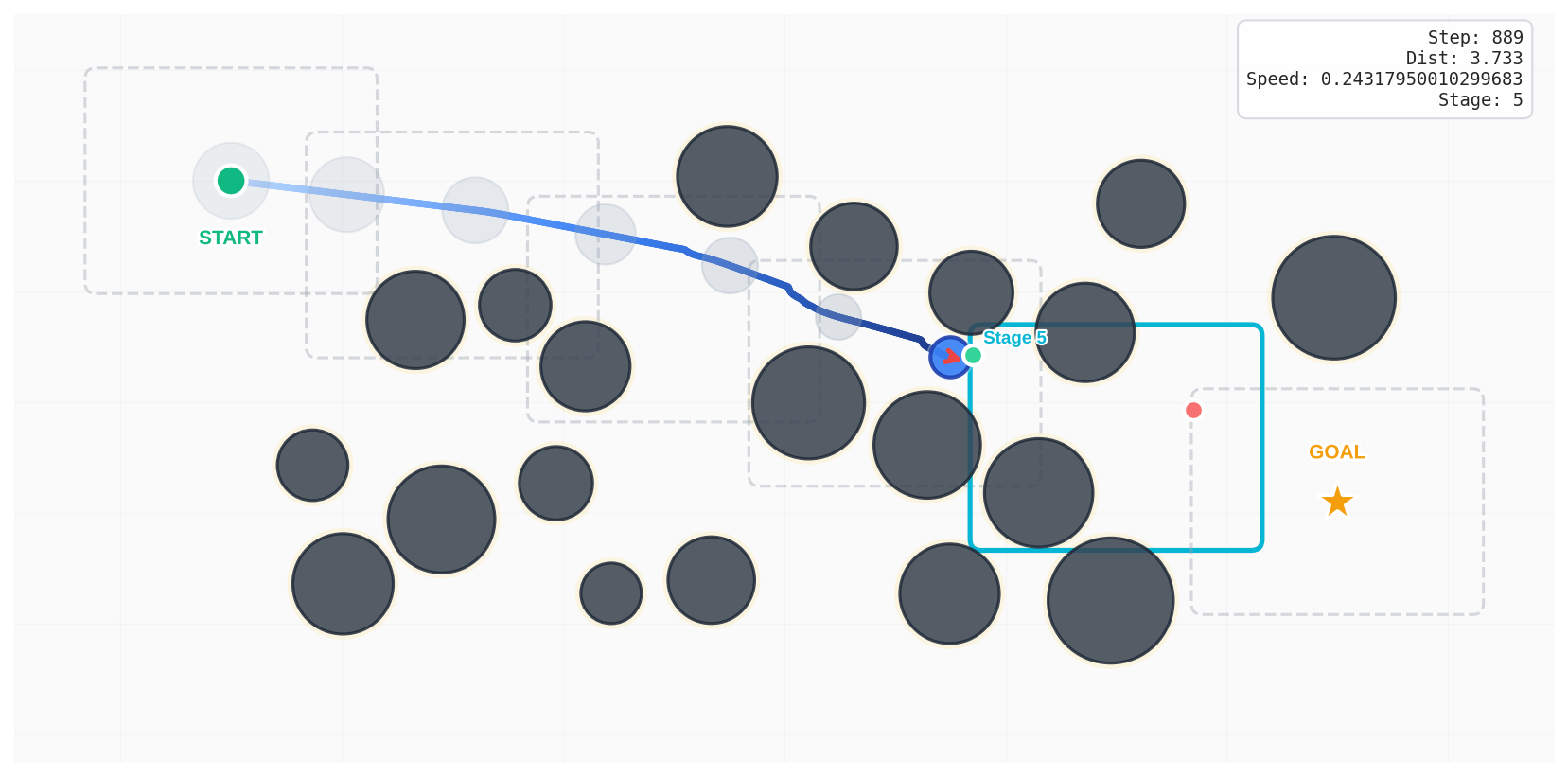}};
\node[panel, right=\H of p5] (p6) {\includegraphics[width=\W]{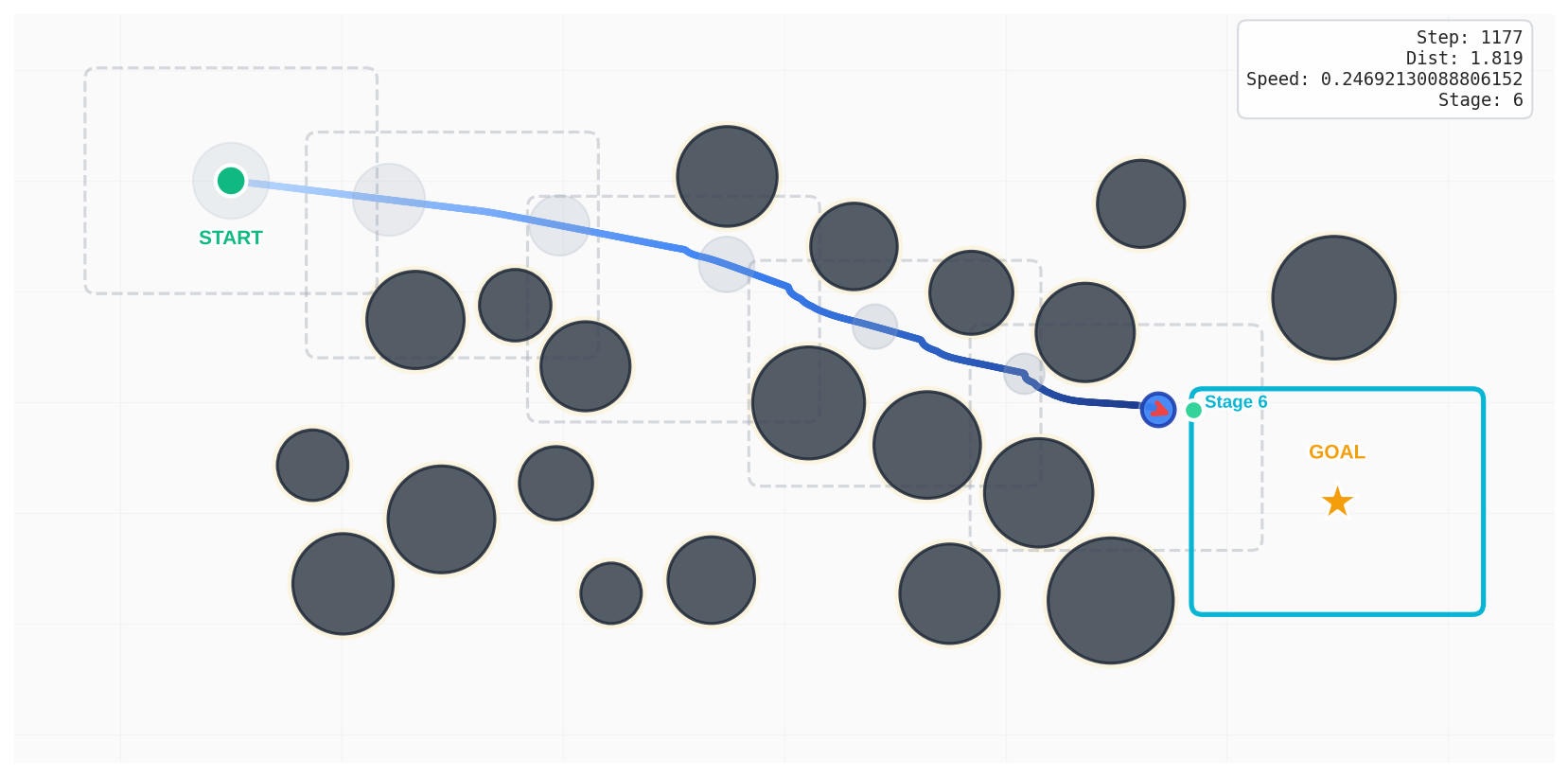}};

\node[panel, below=\V of p4] (p7) {\includegraphics[width=\W]{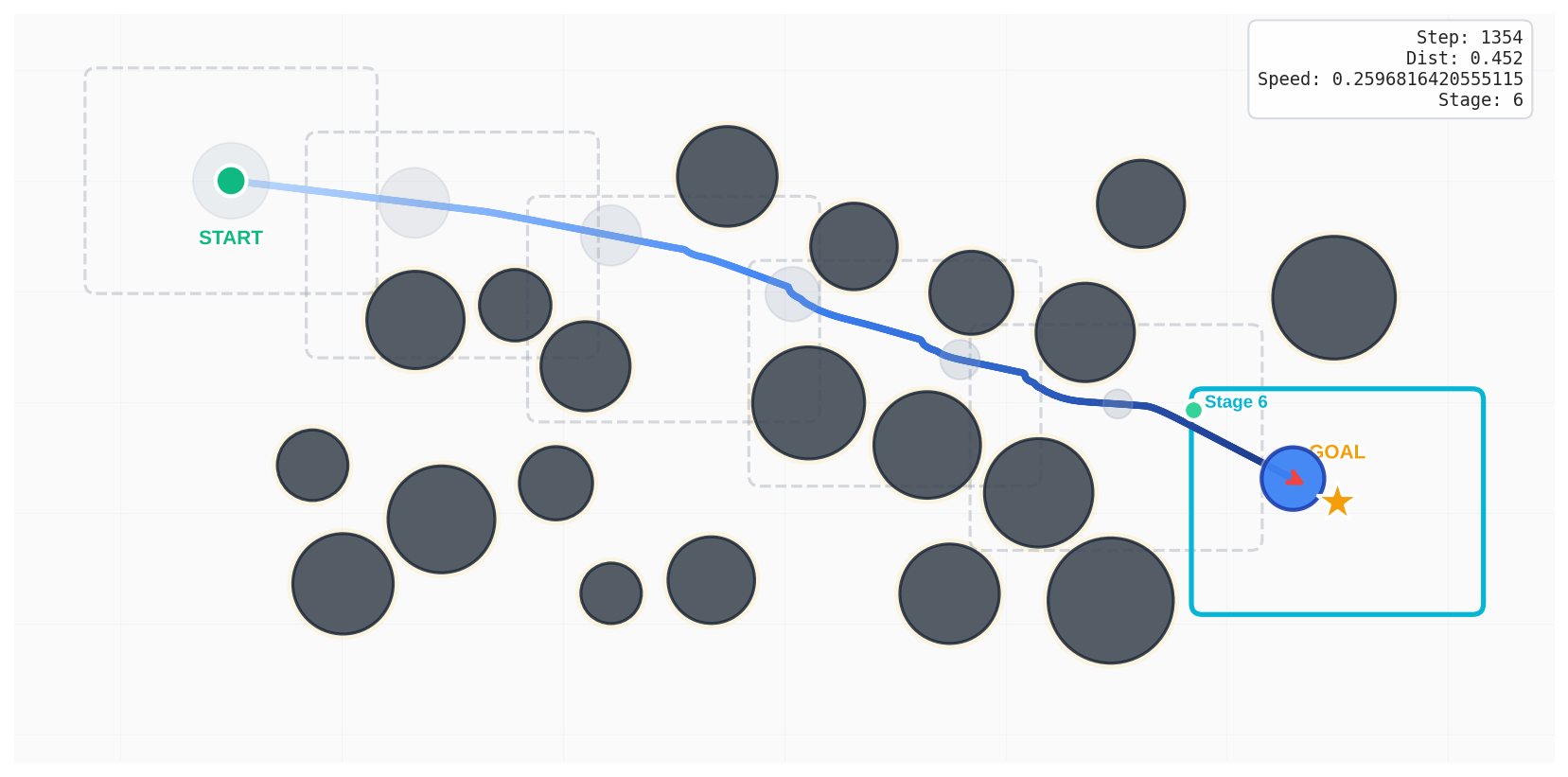}};
\node[panel, right=\H of p7] (p8) {\includegraphics[width=\W]{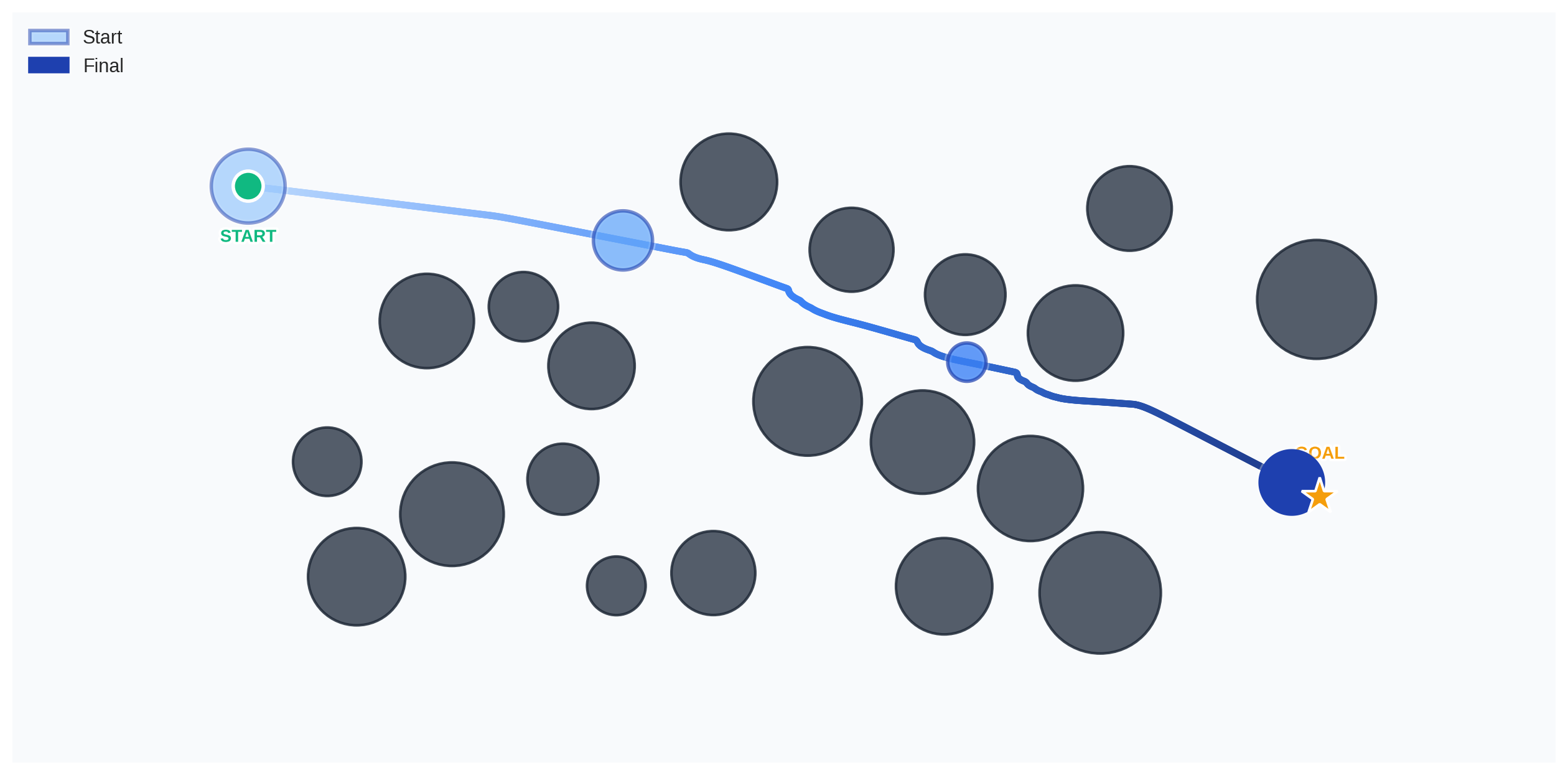}};

\node[stagebadge] at ($(p1.north east)+(-9pt,-9pt)$) {1};
\node[stagebadge] at ($(p2.north east)+(-9pt,-9pt)$) {2};
\node[stagebadge] at ($(p3.north east)+(-9pt,-9pt)$) {4};
\node[stagebadge] at ($(p4.north east)+(-9pt,-9pt)$) {4};
\node[stagebadge] at ($(p5.north east)+(-9pt,-9pt)$) {5};
\node[stagebadge] at ($(p6.north east)+(-9pt,-9pt)$) {6};
\node[stagebadge, goalstyle] at ($(p7.north east)+(-9pt,-9pt)$) {\checkmark};
\node[stagebadge, summarystyle] at ($(p8.north east)+(-9pt,-9pt)$) {$\Sigma$};

\node[timestep, anchor=north west] at ($(p1.north west)+(2pt,-2pt)$) {t\,=\,1};
\node[timestep, anchor=north west] at ($(p2.north west)+(2pt,-2pt)$) {t\,=\,295};
\node[timestep, anchor=north west] at ($(p3.north west)+(2pt,-2pt)$) {t\,=\,592};
\node[timestep, anchor=north west] at ($(p4.north west)+(2pt,-2pt)$) {t\,=\,721};
\node[timestep, anchor=north west] at ($(p5.north west)+(2pt,-2pt)$) {t\,=\,889};
\node[timestep, anchor=north west] at ($(p6.north west)+(2pt,-2pt)$) {t\,=\,1177};
\node[timestep, anchor=north west] at ($(p7.north west)+(2pt,-2pt)$) {t\,=\,1354};
\node[timestep, anchor=north west, fill=black!10, text=black!50] at ($(p8.north west)+(2pt,-2pt)$) {summary};

\node[lab, anchor=south west] at ($(p1.south west)+(2pt,2pt)$) {Initialization};
\node[lab, anchor=south west] at ($(p2.south west)+(2pt,2pt)$) {Corridor commit};
\node[lab, anchor=south west] at ($(p3.south west)+(2pt,2pt)$) {Bottleneck entry};
\node[lab, anchor=south west] at ($(p4.south west)+(2pt,2pt)$) {Squeeze-through};
\node[lab, anchor=south west] at ($(p5.south west)+(2pt,2pt)$) {Recovery};
\node[lab, anchor=south west] at ($(p6.south west)+(2pt,2pt)$) {Final approach};
\node[lab, goalstyle, anchor=south west] at ($(p7.south west)+(2pt,2pt)$) {Goal reached};
\node[lab, summarystyle, anchor=south west] at ($(p8.south west)+(2pt,2pt)$) {Full trajectory};

\draw[->, >=stealth, line width=0.8pt, teal!50!black]
  ($(p1.north west)+(0,10pt)$) -- ($(p3.north east)+(0,10pt)$)
  node[midway, above, font=\sffamily\scriptsize\itshape, text=teal!60!black] {time};
\draw[->, >=stealth, line width=0.8pt, teal!50!black]
  ($(p4.north west)+(0,5pt)$) -- ($(p6.north east)+(0,5pt)$);
\draw[->, >=stealth, line width=0.8pt, teal!50!black]
  ($(p7.north west)+(0,5pt)$) -- ($(p8.north east)+(0,5pt)$);

\end{tikzpicture}
\caption{%
\textbf{GRL-SNAM rollout on a cluttered layout (representative episode).}
Snapshots are ordered left-to-right within each row, with time increasing across rows.
\emph{Row 1:} initialization ($t{=}1$) and early \emph{corridor commitment} ($t{=}295$), followed by \emph{bottleneck entry} ($t{=}592$).
\emph{Row 2:} \emph{squeeze-through} of the narrow passage ($t{=}721$), \emph{recovery/re-centering} ($t{=}889$), and \emph{final approach} to the goal ($t{=}1177$).
\emph{Row 3:} successful arrival ($t{=}1354$) and the full-trajectory overlay.
Circular stage markers
(\protect\tikz[baseline=-0.5ex]{\protect\node[font=\tiny\sffamily\bfseries, text=white, fill=teal!70!black, circle, minimum size=10pt, inner sep=0pt] {n};})
denote the active hierarchical planning stage selected by the navigator at that snapshot.%
}
\label{fig:teaser_3col}
\end{figure*}
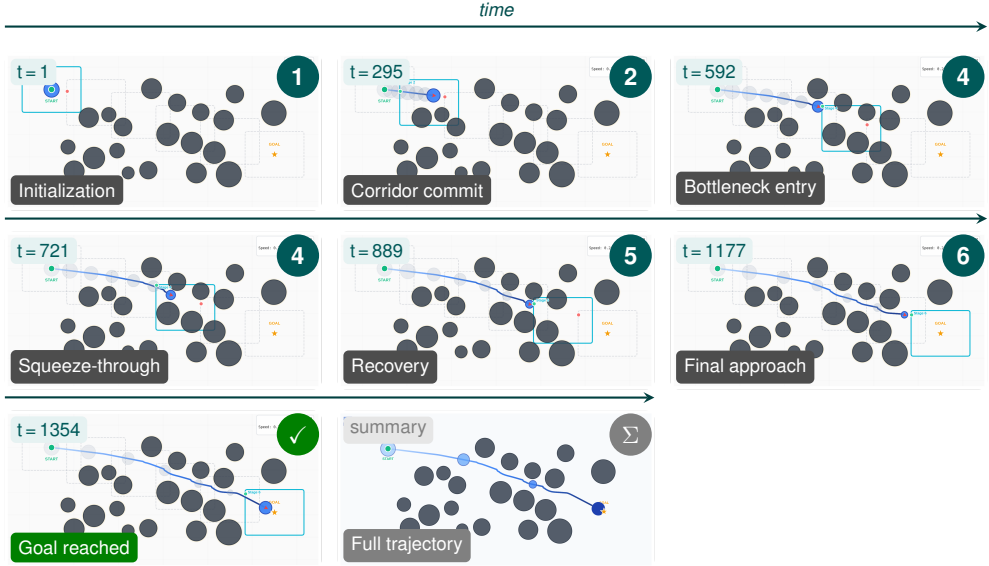

\paragraph{GRL--SNAM: a geometric multi-policy, path-learner navigator process.} We introduce \textbf{GRL--SNAM}, a \emph{Geometric RL framework for Simultaneous Path Navigation and Mapping}, that realize SNAM policy learning via learning to formulate Hamiltonian dynamics. Our main contributions are:

\begin{enumerate}
\item \textbf{Geometric RL framework for SNAM via Hamiltonians (\cref{subsec:problem_formulation}).}
We cast navigation and mapping as formulating  control optimization problem sequentially.
The optimal control problems(OCPs) are mapped to the problem of learning the blue geometry and topology of Hamiltonian landscape through Hamiltonian dynamics. This yields our geometric RL model: it produces differentiable updates of the Hamiltonian. Differentiation in phase space corresponds to a policy update, while variation across environments induces a meta-level adaptation rule {for the lowest amortized cost to move}.

\item \textbf{Multi-scale geometric coordination via multi-policy energies (\cref{subsec:modular_architecture}).}
We introduce a three-submodular-policy architecture whose local policy govern sensing, frame-level motion, and shape control. GRL-SNAM composes each policy via constructing induced Hamiltonian flow maps and realize a discrete, feedforward Hamiltonian dynamics rather than { a backward--forward Hamiltonian-Jacobi-Bellman  optimization}. This achieves temporal scale separation without hand-designed hierarchies as well.

\item \textbf{Physics-grounded offline--online adaptation (\cref{subsec:navigator_meta_learner}).}
Building on our Geometric RL framework, we learn { Hamiltonians via a hybrid offline--online learning scheme}. During offline training, from trajectory and velocity observations, we identify the family of admissible control objectives consistent with each context and learn an embedding of the resulting landscape of local OCPs. While generalizing to unseen, online environments, our method performs online geometric alignment by adapting input-source terms naturally in Hamiltonian dynamics. This { repeated--online} adaptation strategy yields stable, low-variance adaptation that preserves { geometric, topological and functional (i.e. motion physics)} structure.

\item \textbf{Empirical validation on deformable navigation (\cref{sec:experiments}).}
On a challenging hyperelastic ring robot that must squeeze through cluttered, narrow passages, GRL--SNAM achieves higher success rates, better safety margins, and superior sample efficiency compared to { many} standard deep RL baselines (SAC, {TRPO}, PPO), as well as task-specific navigation and mapping baselines such as potential-field (PF) planners, control barrier functions (CBF), and A$^{\star}$-based methods. 
\end{enumerate}

Together, these claims support our rudiments: \textbf{when policy update respects geometric and Hamiltonian structure, complex navigation and mapping behaviors emerge naturally from energy reshaping rather than from brittle reward engineering or value iteration}. We demonstrate this principle on deformable robot navigation in unknown environments, a testbed that captures the core difficulties of real-world SNAM while showcasing the benefits of physics-informed policy design.

\section{Related Works}

We focus on structure-preserving, deployable navigation with deformable bodies under strict local sensing. Our work intersects geometric and Hamiltonian learning, safety-critical control, deformable robot navigation, neural scene representations, and simultaneous navigation and mapping.

\paragraph{Geometric Reinforcement Learning in navigation.}
Most navigation RL methods treat the problem as control in flat Euclidean spaces using standard algorithms such as PPO and TD3~\citep{schulman2017ppo,fujimoto2018td3}. Variants with richer reward shaping or auxiliary losses improve collision avoidance~\citep{taheri2024enhanced}, but still ignore the underlying geometric structure and typically require millions of interactions for robust performance~\citep{dehghani2024survey}. Geometric approaches explicitly exploit configuration-space structure: SE(2)/SE(3)-equivariant policies for manipulation~\citep{hoang2025geometry}, Riemannian formulations that enforce constraints via tangent-space projections~\citep{martinez2023accelerated, riemannian2023safe}, and Hamiltonian / port-Hamiltonian neural networks that preserve symplectic structure and energy~\citep{aps2021port}. However, these works are largely restricted to simple control or manipulation tasks, do not address deformable-body navigation, and are not designed for SNAM. In contrast, GRL-SNAM uses a Hamiltonian formulation for \emph{navigation itself}: sensing, path planning, and deformation all evolve on a shared energy manifold with differential policy optimization rather than ad-hoc Euclidean policy updates. 

\begin{figure}[!htbp]
\centering
\begin{tikzpicture}[
    node distance=0.8cm,
    box/.style={rectangle,draw,thick,rounded corners,minimum width=3.2cm,minimum height=0.8cm,align=center,font=\scriptsize},
    approach/.style={font=\bfseries},
]

\node[approach] at (-4,3.5) {Standard RL};
\node[box,fill=blue!10] at (-4,2.5) {Offline: Learn Policy\\$\pi(a|s)$ from dataset};
\node[box,fill=red!10] at (-4,1.5) {Online: Fine-tune policy\\on new environment};
\node[box,fill=gray!10] at (-4,0.5) {Challenge: Policy transfer\\across domains};

\node[approach] at (1,3.5) {Our GRL-SNAM};
\node[box,fill=green!10] at (1,2.5) {Offline: Learn Hamiltonian\\$h^{\theta}(z,\mathcal{C},t)$ from trajectories};
\node[box,fill=orange!10] at (1,1.5) {Online: Contextual alignment\\$\Delta h^{\text{context}}$ to sensed $\mathcal{C}_t$};
\node[box,fill=purple!10] at (1,0.5) {Advantage: Physics structure\\ensures stable adaptation};

\end{tikzpicture}
\caption{Comparison between standard RL offline/online adaptation and our physics-grounded approach. Standard methods learn arbitrary policies and struggle with transfer, while our approach learns physically meaningful Hamiltonians that naturally adapt to environmental variations.}
\label{fig:rl_comparison}
\end{figure}
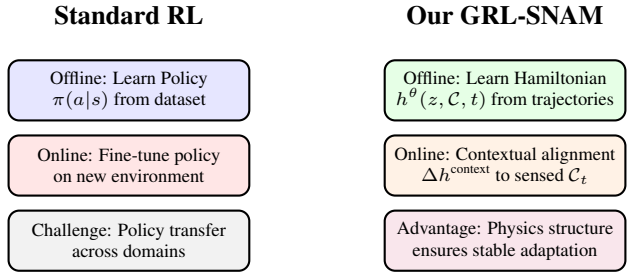

We consider the geometric property in the RL policy iteration itself as well. Most mainstream deep RL algorithms are developed within the dynamic programming principle (DPP): they optimize either value functions or policy surrogates whose correctness is ultimately justified by Bellman consistency. This includes value-based and actor–critic families, even when the update is implemented via policy gradients (e.g., PPO, SAC, TRPO), because the critic and advantage estimates are derived from Bellman-style recursions rather than from Hamilton–Jacobi–Bellman (HJB) analysis or dual optimal control constructions. In contrast, our approach follows a growing line of work that explicitly departs from DPP and instead leverages continuous-time optimal control structure—HJB and/or duality-based Hamiltonian formulations—to define learning objectives and updates without relying on Bellman backups. Representative examples include model-free HJB-based learning \citep{settai2025temporal}, martingale formulations that reinterpret or relax DPP conditions \citep{jia2023qlearning}, and duality-driven Hamiltonian RL \citep{dpo}, which we build upon and extend. The key distinction between GRL-SNAM and other RL navigation methods is two-folded: we do not reparameterize value-function gradients; we learn and adapt a reduced Hamiltonian whose energy decomposition directly induces the closed-loop dynamics. We refer readers Figure \ref{fig:rl_comparison} for a comparison statement.

\paragraph{Safety-critical and reward-exploring navigation.}

When training to plan obstacle-avoidance paths, the design of reward affects the quality of output trajectories via policy learning. Earlier work focuses on handcrafted or intrinsic curiosity reward \citep{tai2017virtual, Lei2018curiosity}. Control Barrier Functions (CBFs) have emerged as a principled way to enforce safety by projecting candidate actions into a safe set, including model-based CBF-RL combinations~\citep{li2023imagination}, neural zeroing barrier functions~\citep{feng2023safer}, adaptive barrier mechanisms for dynamic environments~\citep{sac2025barrier}, and social-navigation CBFs for human--robot interaction~\citep{social2024barrier}. While these methods provide formal safety guarantees, safety is typically treated as an \emph{external projection layer} on top of an otherwise standard controller, which often leads to conservative, sometimes myopic behavior and weak coupling between task optimality and safety. GRL-SNAM instead integrates safety \emph{within} the Hamiltonian energy via barrier potentials: obstacles, free-space preferences, and goal attraction jointly shape the local energy landscape, so that safety and optimality are co-optimized under a symplectic, structure-preserving flow.

\paragraph{Deformable and soft robot navigation.}
Soft and deformable robots have motivated sophisticated modeling and control techniques, including hyperelastic material models with pressure-stiffening control~\citep{roshanfar2023hyperelastic}, passivity-based control via differential geometry of curves~\citep{caasenbrood2022control}, and spectral submanifold reduction for real-time hyperelastic control~\citep{nature2025discovering}. Physics-informed learning further improves fidelity: PINN-Ray~\citep{wang2024pinn} and related models for non-conservative effects~\citep{liu2024physics} achieve accurate displacement prediction but primarily target offline modeling rather than adaptive navigation. For navigation specifically, ring-like or morphable bodies such as aerial LCE-based gap navigators~\citep{qi2024aerial} and HAVEN~\citep{haven2024haptic} use pre-programmed deformation sequences or offline-optimized actuation patterns with deterministic execution. These systems cannot adapt deformation strategies online when geometry of environment changes, and they lack a unifying energy-based view that couples shape, motion, and sensing. GRL-SNAM fills this gap by treating the deformable robot as an energy-carrying body: deformation penalties, contact forces, and path costs are encoded in a Hamiltonian whose gradients drive both motion and shape control under changing local observations.

\paragraph{Neural scene representations for navigation.}
Neural implicit maps provide dense geometric information for navigation. NeRF-based SLAM methods such as NICE-SLAM~\citep{zhu2022nice} and iMAP~\citep{sucar2021imap} reconstruct detailed 3D scenes, while Neural Topological SLAM~\citep{chaplot2020neural} and semantic extensions with large vision models~\citep{zheng2025semantic} combine learned mapping with classical planning. 3D Gaussian splatting approaches like GS-SLAM~\citep{yan2024gs} and SplaTAM~\citep{keetha2024splatam} offer real-time, high-fidelity reconstruction suitable for robot navigation. These methods aim to build accurate maps; the control policy is typically a separate planning layer, and the system often assumes the luxury of extensive mapping before or during navigation. GRL-SNAM is complementary: any of these scene representations can be converted into signed-distance and occupancy-derived potentials for our barrier and goal terms, but our objective is to \emph{minimize mapping effort}, using only local observations to progressively refine a Hamiltonian that encodes the task.

\paragraph{Simultaneous navigation and mapping (SNAM).}
Most SNAM methods prioritize constructing rich global maps and then planning on them. SGoLAM~\citep{sgolam2021} couples goal localization with occupancy mapping, CMP~\citep{gupta2017cognitive} integrates a differentiable planner into a learned map, and CL-SLAM~\citep{clslam2022} focuses on maintaining maps for long-term operation. These approaches do not explicitly reason about the cost of mapping itself and are not designed for deformable bodies or binded with optimal control problems. Some works encourage to learn a context-aware reward \citep{liang2023context} or an explicit terrain cost map \citep{sathyamoorthy2022terrapn} that can be used to compute dynamically feasible trajectories while simultaneously reading the partial map, yet they still rely on RL-driven policy network or model-based fixed rule (i.e. DWA, \citet{fox2002dynamic}) to apply the learned reward/cost. GRL-SNAM instead targets \emph{joint exploration}: the robot maintains only the local information needed to define a good energy landscape around its current belief, continually refining least-cost trajectories as new observations arrive rather than insisting on globally complete maps.

\paragraph{Hierarchical, multi-agent, and imitation-based navigation.}
Hierarchical RL decomposes complex navigation into sub-tasks with high-level coordinators, as in HRL4IN~\citep{li2020hrl4in} and adaptive policy-family methods~\citep{lee2023adaptive}. Multi-agent frameworks such as RoboBallet~\citep{lai2025roboballet} and MACRPO~\citep{macrpo2024} achieve impressive coordinated behavior using graph-based policies and improved information sharing. However, these methods typically rely on manually specified hierarchies or reward structures and seldom preserve underlying geometric or energy structure, making it difficult to guarantee stable coordination across time scales. Imitation learning and inverse RL further improve sample efficiency using expert trajectories: large-scale behavioral cloning (RT-1)~\citep{brohan2022rt1}, safe imitation via GAIL-style methods~\citep{gail2024safe}, DAgger-based continuous navigation~\citep{shi2025dagger}, and confidence-aware imitation for sub-optimal demonstrations~\citep{confident2024il}. These approaches are powerful when high-quality demonstrations exist but struggle when expert coverage of the full deformation and contact behavior space is limited. GRL-SNAM shares with hierarchical and imitation methods the aim of efficient learning but differs in \emph{how} structure is encoded: we do not hand-design a task graph; instead, sensing, path, and deformation policies are coupled through a shared Hamiltonian with differential updates, yielding principled multi-scale coordination and stability guarantees.

\paragraph{Foundation models for navigation reasoning.}
Recent work leverages large foundation models for high-level navigation reasoning, goal inference, and multi-agent coordination~\citep{zhu2024navi2gazeleveragingfoundationmodels,llm2025nav}. These systems excel at semantic understanding, language-grounded objectives, and flexible decision-making but generally outsource low-level control to conventional planners or RL policies, creating a gap between symbolic reasoning and physically consistent motion. GRL-SNAM is complementary: foundation models can supply high-level instructions or semantic goals that we encode as potential energy terms in our Hamiltonian, while our structure-preserving dynamics ensure that these goals translate into safe, physically plausible trajectories for a deformable body under local sensing.

\paragraph{Positioning of GRL-SNAM.}
Across these paradigms, existing methods typically provide \emph{some} of: geometric structure, safety, sample efficiency, deformable-body support, or SNAM capabilities, but not all simultaneously. Our GRL-SNAM framework is, to our knowledge, the first to (i) formulate deformable-body SNAM as coupled Hamiltonian dynamics on a shared energy manifold under strict local sensing, (ii) integrate safety, deformation costs, and navigation objectives directly into the energy rather than in external projection layers, and (iii) coordinate sensing, planning, and deformation via a differential multi-policy architecture with symplectic updates. This combination yields high-quality, well-weighted paths with minimal mapping effort, while preserving geometric structure and enabling principled analysis of stability and convergence.
\section{Methodology}
\label{sec:method}
\providecommand{\Ham}{\mathcal{H}}          
\providecommand{\Energy}{\mathcal{E}}       
\providecommand{\Potential}{\mathcal{R}}    
\providecommand{\Constraints}{\mathcal{C}}  
\providecommand{\Response}{\mathsf{R}}      
\providecommand{\Navigator}{\mathsf{N}}     
\providecommand{\PhaseState}{z}                  
\providecommand{\Pos}{q}                    
\providecommand{\Mom}{p}                    
\providecommand{\Mass}{\mathrm{M}}          
\providecommand{\J}{\mathbb{J}}             
\providecommand{\Lag}{\mathscr{L}}          
\providecommand{\res}{\mathbf{r}}           
\providecommand{\mult}{\boldsymbol{\lambda}}
\providecommand{\pen}{\boldsymbol{\rho}}    

\begin{figure}[htbp]
\centering
\begin{tikzpicture}[
  font=\small,
  >=Latex,
  box/.style={draw, rounded corners=2pt, very thick, inner sep=1pt, fill=black!1},
  rowtag/.style={draw, rounded corners=6pt, fill=black!6, inner xsep=6pt, inner ysep=2pt,
                 font=\scriptsize\bfseries, align=center},
  header/.style={draw, rounded corners=3pt, fill=black!10, very thick,
                 inner xsep=6pt, inner ysep=3pt, font=\scriptsize\bfseries, align=center},
  legendbox/.style={draw, rounded corners=2pt, thick, fill=white, inner sep=5pt},
  arrow/.style={->, very thick},
  accent/.style={draw=RoyalBlue, very thick},
  accent2/.style={draw=Crimson,  very thick},
  accent3/.style={draw=ForestGreen, very thick},
  alabel/.style={font=\scriptsize, fill=white, rounded corners=2pt, inner sep=2pt}
]

\matrix (G) [matrix,
             row sep=10mm,
             column sep=38mm,
             nodes={anchor=center}] at (0,0) {

  \node[box] (phys0) {\includegraphics[width=7.0cm,height=2.35cm,keepaspectratio]{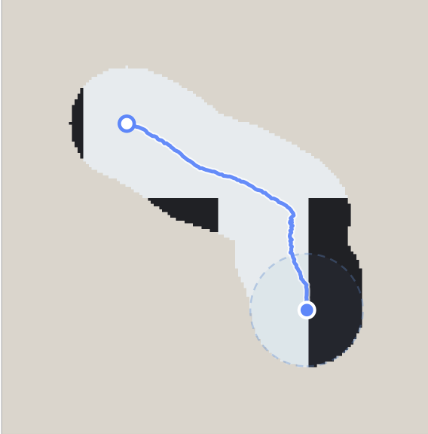}}; &
  \node[box] (eng0)  {\includegraphics[width=7.0cm,height=2.35cm,keepaspectratio]{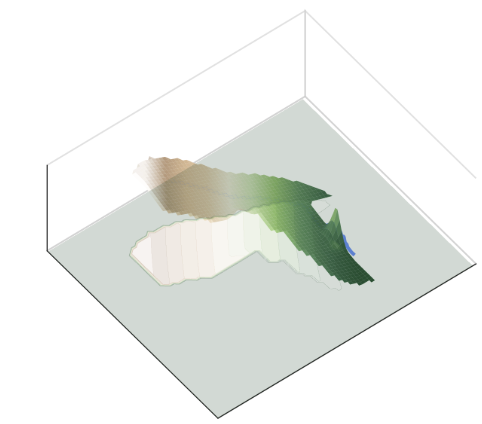}}; \\

  \node[box] (phys1) {\includegraphics[width=7.0cm,height=2.35cm,keepaspectratio]{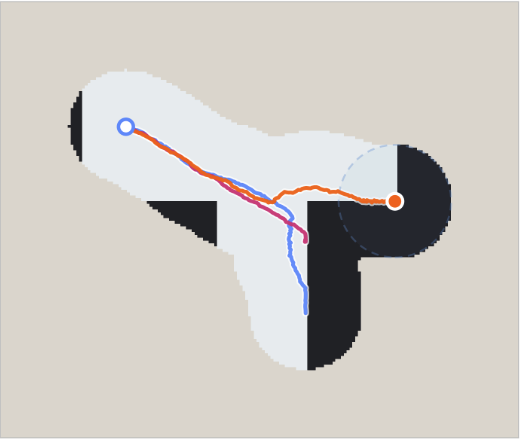}}; &
  \node[box] (eng1)  {\includegraphics[width=7.0cm,height=2.35cm,keepaspectratio]{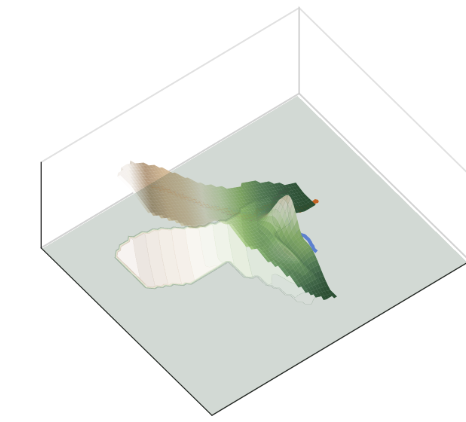}}; \\

  \node[box] (phys2) {\includegraphics[width=7.0cm,height=2.35cm,keepaspectratio]{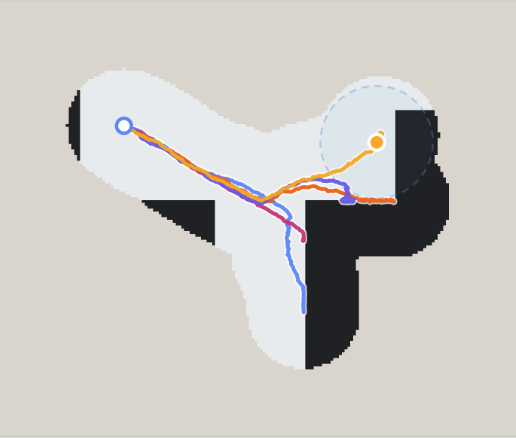}}; &
  \node[box] (eng2)  {\includegraphics[width=7.0cm,height=2.35cm,keepaspectratio]{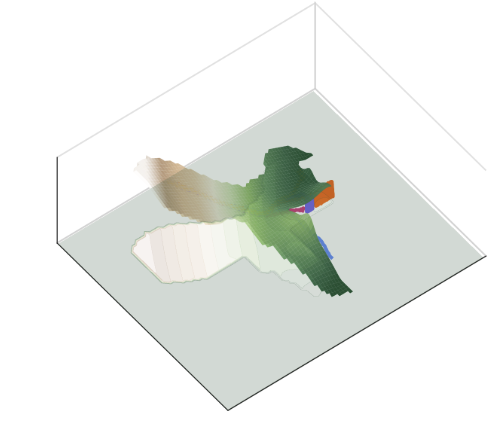}}; \\
};

\node[header, text width=4.5cm, anchor=south] (colL) at ($(phys0.north)+(0,7mm)$)
{Physical space\\(unknown $\rightarrow$ revealed by local sensing)};

\node[header, text width=4.0cm, anchor=south] (colR) at ($(eng0.north)+(0,7mm)$)
{Hamiltonian energy terrain\\(learned progressively)};

\node[font=\small\bfseries, anchor=south]
  at ($(colL.north)!0.5!(colR.north)+(0,2.5mm)$)
{Progressive Hamiltonian Landscape Learning from Local Sensing {along Hamiltonian geodesic pathways}};

\draw[arrow] (phys0.south) -- (phys1.north);
\draw[arrow] (phys1.south) -- (phys2.north);
\draw[arrow] (eng0.south) -- (eng1.north);
\draw[arrow] (eng1.south) -- (eng2.north);

\draw[arrow, accent] (phys0.east) -- (eng0.west)
  node[midway, above=2mm, alabel] {scan local env $\Rightarrow$ build initial $H_{\tau}$};

\draw[arrow, accent2] (phys1.east) -- (eng1.west)
  node[midway, above=2mm, alabel] {Try 1 feedback $\Rightarrow$ update $H_{\tau}$};

\draw[arrow, accent3] (phys2.east) -- (eng2.west)
  node[midway, above=2mm, alabel] {Try 2 feedback $\Rightarrow$ refine $H_{\tau}$};

\node[draw, rounded corners=2pt, thick, fill=black!1,
      inner sep=6pt, text width=13cm, align=center, anchor=north]
  at ($(G.south)+(0,-11mm)$) {
  \textbf{Idea:} In an unknown environment, the agent repeatedly senses a \emph{local} neighborhood,
  attempts to reach the stage goal, and incrementally learns a Hamiltonian energy landscape whose valleys guide future attempts under partial observability.
};

\end{tikzpicture}
\caption{\textbf{Conceptual overview: progressive energy-landscape shaping from local sensing.}
\textit{Left}: the physical workspace is initially unknown and is revealed only through local sensing around the agent, producing a partial view of nearby obstacles and a stage goal.
\textit{Right}: from this partial context, the agent builds an initial internal “energy terrain” and then repeatedly attempts a short plan toward the stage goal; each attempt produces simple feedback (e.g., collisions/clearance and progress), which is used to refine the terrain. Over successive refinements, the terrain becomes smoother and its low-energy valleys align with traversable corridors, guiding future attempts even under partial observability.}

\label{fig:teaser_progressive_landscape}
\end{figure}

\paragraph{Overview.} 
We address how GRL--SNAM converts an online mapless navigation task into a stagewise {\emph{energy landscape exploration and mapping}} process under dynamical and geometric constraints. The SNAM task is to move a (deformable) robot in 2d space from $\vec x_0\in\mathbb R^2$ to a goal $\vec x_g\in\mathbb R^2$ in an unknown workspace whose obstacles are encoded by an (unknown) occupancy field $I:\mathbb R^2\to\{0,1\}$. The robot is bound with state $\vec q=(\vec y, \vec c, \vec\psi)\in \mathcal{Q}$ where $\vec c\in\mathbb R^2$ parameterize the robot pose in world coordinate, and $\vec y\in \mathbb{R}^{\text{sensor}},\vec\psi\in \mathbb{R}^{\text{robot}}$ collect sensory and deformation/object configuration variables. The configuration space $\mathcal{Q}$ is assumed to be decomposable into three individual configuration space $\mathcal{Q}=\mathcal{Q}_y\times \mathcal{Q}_f\times \mathcal{Q}_o$, with $\vec{y}\in\mathcal{Q}_y$, $\vec{c}\in \mathcal{Q}_f$ and $\vec\psi\in\mathcal{Q}_o$.

{ We assume the navigation system can query an independent sensor and the sensor would return an environment context $\mathcal{E}$ sensed via current configuration} $\vec{q}$\footnote{How to design the best sensory pipeline is out of this paper's discussion. We assume that sensor can always return the context of environment upon giving {its current} configuration $\vec{q}$.}. The context $\mathcal{E}$ at least include a local target goal $\vec{c}_{target}(\mathcal{E})\in \mathbb{R}^{2}$,  an obstacle cardiality $C(\mathcal{E})\in \mathbb{N}$ and $C(\mathcal{E})$ non-collided constraints $\{d_i(\vec{q};\mathcal{E})\geq 0\}_{i=1}^{C(\mathcal{E})}$. Given current stage's context $\mathcal{E}$, GRL--SNAM learns to ``construct" optimal control problems, whose solution offers an optimally controlled path, by learning to map each environment context $\mathcal E_\tau$ at stage $\tau=0,1,2,\ldots,$ into a \textit{reduced Hamiltonian} via a parametrized network $\eta_{\xi}$:
\[
\eta_{\xi}:\mathcal E_\tau \longmapsto H(\cdot;\mathcal E_\tau)\in \mathscr H,
\quad 
\mathscr H\cong C^{\infty}(T^{*}\mathcal{Q},\mathbb{R}),
\]
where $T^{*}\mathcal{Q}$ refers cotangent bundle of $\mathcal{Q}$. An element of $T^{*}\mathcal{Q}$ is a pair $(\vec{q},p)$ with $\vec{q}\in \mathcal{Q}$ and $p\in T^{*}_{\vec{q}}\mathcal{Q}$, where $p$ is a covector (a linear functional)
acting on tangent vectors $\vec{v}\in T_{\vec{q}}\mathcal{Q}$. The \emph{policy} $\pi$, different from conventional RL definition, is denoted as the time-$\Delta t$ flow map induced by Hamiltonian vector field:
\begin{equation}
\pi(\vec{q}|p):=\ \Phi^{\Delta t}_{H}(\vec{q},p),\quad \text{ where } \Phi^{\Delta t}_{H}(\cdot):T^*\mathcal{Q}\to T^*
\mathcal{Q},
\quad (\vec{q}_{t+\Delta t},p_{t+\Delta t})=\Phi^{\Delta t}_{H}(\vec{q}_{t},p_{t}).
\end{equation}

It then treats a SNAM problem as finding three policies operating at different effective timescales of each subsystem by decomposing $H$ into three Hamiltonians $H_y:T^{*}\mathcal{Q}_y\rightarrow \mathbb{R}$, $H_f:T^{*}\mathcal{Q}_f\rightarrow \mathbb{R}$ and $H_o:T^{*}\mathcal{Q}_o\rightarrow \mathbb{R}$. 
 \begin{enumerate}[i.)]
     \item a {\emph{sensing policy}} $\pi_y: T^*\mathcal{Q}_y\to T^*
\mathcal{Q}_y$ that maintains how to update sensor configuration to acquire environment representation and active collision constraints from partial observations;
     \item a \emph{frame policy} $\pi_f:  T^*\mathcal{Q}_f\to T^*
\mathcal{Q}_f$ that plans short-horizon motion in local frames toward intermediate “stage” goal $\vec{c}_{target}$;
     \item a \emph{shape policy} $\pi_o: T^*\mathcal{Q}_o\to T^*\mathcal{Q}_o$ that modulates the robot’s morphology (e.g., deformation or scale) to exploit narrow passages.
 \end{enumerate}
 Rather than stacking these as a manually tuned hierarchy, we treat each policy as a flow map with its own energy terms under a shared latent codex of sensored environments. We denote $g_{\xi}$, which contains both $\eta_{\xi}$ and policy specific parameter (discussed in \cref{subsec:modular_architecture}), as the overall navigator function of GRL--SNAM. GRL--SNAM further separates learning $g_{\xi}$ into complementary regimes: \emph{offline}, we train reference Hamiltonians on trajectory data to capture fundamental couplings between sensing, planning, and deformation in local frames;
\emph{online}, we adapt open-loop control terms that reweight fixed energy components and add minimal non-conservative corrections injected into policies. Formally, offline learning yields a initial reference reduced Hamiltonian $H$ per local scenario $\mathcal{E}$, and online adaptation performs geometric alignment by updating $\pi_k,k\in\{s,f,o\}$. This realizes a conservative adaptation principle: systems default to learned physics-based behavior while adding just enough correction to handle new obstacles or map updates, rather than catastrophically relearning policies. \cref{algo:online_grl_snam_main:short} shows the overall pipeline. \cref{alg:meta_training} discuss how to obtain $g_{\xi}$ offline. The full pseudo-code, including initialization, query protocols, energy assembly, integration details, and adaptation rules, is deferred to Algorithm~\ref{algo:online_grl_snam} in Appendix~\ref{app:algorithm_details}.

\begin{algorithm}[h!]
\caption{Online GRL-SNAM pipeline}
\label{algo:online_grl_snam_main:short}
\begin{algorithmic}[1]
\State \textbf{Input:} goal $\mathbf{x}_g$, initial $z_0=(q_0,p_0)$ (typically $p_0{=}0$), step $\tau$, horizons $(T_y,T_f,T_o)$, max steps $N_{\max}$
\State \textbf{Init:} $n\!\leftarrow\!0$, environment $\mathcal{E}_0$, environment counter $\tau\gets 0$ active set $\mathcal{C}_0\!\leftarrow\!\emptyset$
\State \textbf{Init:} meta state $(\eta_0,\mu_0,u_{f,0})$ from network $g_\xi$,
\State Assemble $H$ by computing $g_{\xi}(\mathcal{E}_{0})$
\While{$\neg$\textsc{ReachedGoal}$(\vec{c}_n,\mathbf{x}_g)$ \textbf{and} $n<N_{\max}$}
    
    \If{$n \equiv 0 \pmod{T_y}$}   
        \State Assemble $H_y$, provide $z_{y,0}$.
        \State Query Sensor Process via $\pi_y$ and get response $\mathsf{R}_y$
    \EndIf
    \If{$n \equiv 0 \pmod{T_f}$}
        \State Assemble $H_f$, $u_{f}^{\xi,n}, \mu_{f}^{\xi,n}$, provide $z_{f,0}$.
        \State Query Free Path Extractor via $\pi_f$ and get response $\mathsf{R}_f$
    \EndIf
    \If{$n \equiv 0 \pmod{T_o}$}
        \State Assemble $H_o$, provide $z_{o,0}$.
        \State Query Shape Reconfig via $\pi_o$ and get response $\mathsf{R}_o$
    \EndIf
    \State Collect QoIs $y_n$ from responses $\mathsf{R}_y$, $\mathsf{R}_f$ and $\mathsf{R}_o$.
    \State Compute correction term $u_f^{\xi,n}$ via $y_n$.
    \If{$n \equiv 0 \pmod{T_y}$ \textbf{or} reach stage goal}
    \State Update $\mathcal{E}_{\tau+1}$ via response $\mathsf{R}_y$.
    \State Reassemble $H$ by computing $g_{\xi}(\mathcal{E}_{\tau+1})$
    \State $\tau\gets \tau+1$
    \EndIf
    \State $n\leftarrow n+1$
\EndWhile
\State \textbf{Return:} trajectory $\{z_n\}_{n=0}^{N}$ and parameter history $\{(\eta_n,\mu_{f}^{\xi,n},u_{f}^{\xi,n})\}_{n=0}^{N}$
\end{algorithmic}
\end{algorithm}

\subsection{Navigation as Hamiltonian Optimization}
\label{subsec:problem_formulation}
\paragraph{From a stagewise constrained optimal control problem to a reduced Hamiltonian system.}
At a fixed stage $\tau$ and local scenario $\mathcal E_\tau$, we model motion planning for the deformable robot state
$\vec q\in Q$ as a \emph{constrained optimal control problem}:
\begin{equation}
\label{eq:motion:planning:continuous:original:form}
\min_{\vec u(\cdot)\in\mathcal U}\ \int_0^{T} L(\vec q(t),\vec u(t),t;\mathcal E_\tau)\,dt
\quad
\text{s.t.}\;
\begin{cases}
\dot{\vec q}(t)= v(\vec q(t),\vec u(t),t;\mathcal E_\tau), \\
d_i(\vec q(t);\mathcal E_\tau)\ge 0,\quad i=1,2,\ldots, C(\mathcal E_\tau), \\
\vec q(0)=\vec q_0,\quad \vec q(T)=\vec q_{\text{target}}(\mathcal E_\tau).
\end{cases}
\end{equation}
Here $\vec q_0$ is inherited from the previous stage $\mathcal E_{\tau-1}$ (or initialized as $(\vec x_0,\vec 0,\vec 0,\vec 0)$ for $\tau=1$),
and $\vec q_{\text{target}}(\mathcal E_\tau)$ is a local goal specified by the stagewise planner.
The inequality constraints $d_i(\vec q;\mathcal E_\tau)\ge 0$ encode collision avoidance with the locally sensed geometry;
their number $C(\mathcal E_\tau)$ and active subset may vary across scenarios.
Under standard regularity assumptions, a necessary condition for optimality in \cref{eq:motion:planning:continuous:original:form}
is given by the \emph{state-constrained Pontryagin maximum principle(PMP)} \citep{FERREIRA1994whenpmp, hartl1995survey, pearson1966discrete}.
We summarize the resulting reduction to a Hamiltonian system below (a formal statement and proof are provided in
\cref{thm:forward_state_constrained_pmp_reduced_ham}).

\begin{lemma}
\label{thm:fwd_pmp_to_reduced_ham:short}
Consider \cref{eq:motion:planning:continuous:original:form} and assume:
(i) control-affine dynamics $v(\vec q,\vec u,t;\mathcal E)=f(\vec q;\mathcal E)+A(\vec q;\mathcal E)\,u$,
(ii) separable running cost $L(\vec q,\vec u,t;\mathcal E)=\ell(\vec q;\mathcal E)+\varphi(\vec u)$ with $\varphi$ proper, closed, and strictly convex,
and (iii) the regularity/qualification conditions required for the state-constrained maximum principle.
Then there exist a costate $p:[0,T]\to\mathbb R^{\dim Q}$ of bounded variation and nonnegative Radon measures
$\{\mu_i\}_{i=1}^{C(\mathcal E_\tau)}$ such that, defining the reduced Hamiltonian
\begin{equation}
\label{eq:reduced_ham_short}
H_{PMP}(\vec{q},p;\mathcal E):=\ell(\vec{q};\mathcal E)+p^\top f(\vec{q};\mathcal E)+\varphi^\ast\!\big(A(\vec{q};\mathcal E)^\top p\big),
\end{equation}
the optimality conditions can be written (in measure form) as
\begin{equation}
\label{eq:reduced_ham_flow_short}
\begin{aligned}
\dot{\vec{q}}^{\star}(t)&=\nabla_p H_{PMP}\big(\vec{q}^{\star}(t),p(t);\mathcal E\big),\\
dp(t)&= -\nabla_q H_{PMP}\big(\vec{q}^{\star}(t),p(t);\mathcal E\big)\,dt
-\sum_{i=1}^{C(\mathcal E)} \nabla d_i\!\big(\vec{q}^\star(t);\mathcal E\big)\,\mu_i(dt).    
\end{aligned}
\end{equation}
In particular, on interior intervals where all constraints are inactive, $\mu_i\equiv 0$ and \cref{eq:reduced_ham_flow_short}
reduces to the classical Hamiltonian system $\dot{\vec{q}}^\star=\nabla_p H$, $\dot p=-\nabla_{\vec{q}} H$.
\end{lemma}

\paragraph{Barrier relaxation and the learnable {sub-modular} Hamiltonian.}
For learning and differentiable rollout, we employ a smooth barrier relaxation of the state constraints \citep{NocedalWright2006NumericalOptimization}.
Let $b_i:\mathbb R\to\mathbb R$ be a $C^2$ barrier potential applied to the clearance $d_i(q;\mathcal E)$, and let
$\alpha_i(\mathcal E)\ge 0$ be environment-conditioned weights.
Define the \emph{barrier-shaped} running cost
\begin{equation}
\label{eq:barrier_shaped_cost}
L_{\alpha}(q,u;\mathcal E)
:=\ell(q;\mathcal E)+\varphi(u)+\sum_{i=1}^{C(\mathcal E)} \alpha_i(\mathcal E)\, b_i\!\big(d_i(q;\mathcal E)\big),
\end{equation}
which yields an \emph{unconstrained} OCP that approximates \cref{eq:motion:planning:continuous:original:form} as the barriers stiffen.
Applying the reduction in \cref{cor:mechanical_reduced_hamiltonian} (quadratic $\varphi$ and $f\equiv 0$)
leads to the PMP Hamiltonian
\begin{equation}
\label{eq:def:pmp:hamiltonian:grl-snam}
H_{PMP}(q,p;\mathcal E)
=\tfrac12\,p^\top M^{-1}p- V(q;\mathcal E),
\quad
V(q;\mathcal E)
:= \ell(q;\mathcal E)+\sum_{i=1}^{C(\mathcal E)} \alpha_i(\mathcal E)\, b_i\!\big(d_i(q;\mathcal E)\big),
\end{equation}
where $M\in\mathbb S_{++}^2$ is the (learned or fixed) mass/metric induced by the control penalty. We commented on \cref{prop:barrier_convergence} that under certain assumptions log-barrier relaxed OCPs can recover the constrained OCP. The statement of the proposition is inspired from \citet{malisani2023interior}.

\begin{proposition}[Barrier relaxation converges to the state-constrained OCP (log-barrier)]
\label{prop:barrier_convergence}
Fix a scenario $\mathcal E$ and consider the state-constrained OCP
\begin{equation}
\begin{aligned}
\min_{u(\cdot)}&\ J(u):=\int_0^T L(q(t),u(t),t;\mathcal E)\,dt
\\ \text{s.t.}&\quad
\dot q=f(q;\mathcal E)+A(q;\mathcal E)u,\ \ d_i(q(t);\mathcal E)\ge 0\ \forall t,\ \ q(0)=q_0,\ q(T)=q_T.    
\end{aligned}
\end{equation}

Assume:
(A1) (\emph{Slater}) there exists a strictly feasible trajectory, i.e., some admissible $(\bar q,\bar u)$ with
$d_i(\bar q(t);\mathcal E)\ge \delta$ for all $t\in[0,T]$ and all $i$, for some $\delta>0$;
(A2) $f,A,L,d_i$ are $C^2$ in $q$ (and $L$ is convex in $u$), and the constrained OCP admits a (locally) optimal solution
$(q^\star,u^\star)$ satisfying a suitable constraint qualification and second-order sufficient conditions.

For $t>0$, define the \emph{log-barrier relaxed} (unconstrained) OCP
\[
\min_{u(\cdot)}\ J_t(u):=\int_0^T \Big(
L(q(t),u(t),t;\mathcal E)\;+\;\frac{1}{t}\sum_{i=1}^{C(\mathcal E)} \alpha_i(\mathcal E)\, \phi(d_i(q(t);\mathcal E))
\Big)\,dt,
\quad
\phi(s):=-\log s,
\]
subject only to the dynamics and boundary conditions (and implicitly $d_i(q(t);\mathcal E)>0$ due to $\phi$).
Let $(q^t,u^t)$ be a (locally) optimal solution of the barrier subproblem.

Then:
\begin{enumerate}
\item (\emph{Strict feasibility}) For each $t$, $d_i(q^t(t);\mathcal E)>0$ for all $t\in[0,T]$ and all $i$.
\item (\emph{Convergence of primal variables}) Any sequence $t_k\to\infty$ has a subsequence (not relabeled) such that
$(q^{t_k},u^{t_k})$ converges (in a topology appropriate for the OCP, e.g.\ $q^{t_k}\to q^\star$ strongly and $u^{t_k}\rightharpoonup u^\star$ weakly)
to a (locally) optimal solution $(q^\star,u^\star)$ of the original state-constrained OCP.
\item (\emph{Recovery of multipliers}) Define the barrier-induced (approximate) multipliers
\[
\lambda_i^t(\tau)\ :=\ \frac{\alpha_i(\mathcal E)}{t}\,\frac{1}{d_i(q^t(\tau);\mathcal E)}\ \ge 0.
\]
Along the same subsequence, $\lambda_i^{t_k}$ converges (weakly$^\ast$) to a nonnegative multiplier $\lambda_i^\star$
appearing in the state-constrained PMP/KKT system of the original OCP; moreover complementarity holds in the limit:
$\lambda_i^\star(\tau)\,d_i(q^\star(\tau);\mathcal E)=0$ for a.e.\ $\tau$.
\end{enumerate}
\end{proposition}

\begin{proof}
Defer to \cref{prop:barrier_to_state_constrained_complete}.
\end{proof}

\paragraph{Our geometric reinforcement learning framework.}

GRL-SNAM {learns} a scenario-to-energy map $\mathcal E\mapsto H(\cdot;\mathcal E)$. With a costate initialized, the resulting phase-space integrator
$x_{t+1}=\Phi_{\eta}^{\Delta t}(x_t;\mathcal E)$ with $x_t=(q_t,p_t)$ produces a feasible start-to-goal rollout using only forward passes. We view our proposed method as an Energy-Based Model \citep{hopfield1982neural,ackley1985learning,lecun2006tutorial} yet with augmented search space considering geometric structure of energy landscape \citep{nguyen2025stochastic, ryan2025bayesian, mclennan2025learning}. For deterministic dynamics $\dot q=f(q)+A(q)u$ and cost minimization, the value function satisfies the HJB equation
\[
-\partial_t \mathcal V^{*}(t,q;\mathcal E)
=\inf_{u\in U}\Big\{L(q,u;\mathcal E)+\nabla_q\mathcal V^{*}(t,q;\mathcal E)^\top\big(f(q)+A(q)u\big)\Big\},
\qquad
\mathcal V^{*}(T,q;\mathcal E)=\delta_{q_T}(q).
\]
When $\mathcal V^{*}$ is smooth, the PMP costate admits the feedback representation
$p(t)=\nabla_q\mathcal V^{*}(t,q(t);\mathcal E)$ but one needs to solve the space-time PDE backward in time.
Conventional RL methods approximate $\mathcal V^{*}$ via Bellman recursion/backups, and explore its stochastic counterpart. Our design principle is that the optimal motion could be generated by a forward dynamics. To this end, we propose to learn a generalized Hamiltonian dynamical system whose reduced Hamiltonian is defined as:
\begin{equation}
\label{eq:def:reduced:hamiltonian:grl-snam}
H(q,p;\mathcal E)
=\tfrac12\,p^\top M^{-1}p + \mathcal{R}(q;\mathcal E),
\quad
\mathcal{R}(q;\mathcal E)
:= \ell(q;\mathcal E)+\sum_{i=1}^{C(\mathcal E)} \alpha_i(\mathcal E)\, b_i\!\big(d_i(q;\mathcal E)\big).
\end{equation}
The sign of potential term is reversed as we are approximating the feedforward dynamics using $\nabla_q H$, and the flow reversibility is addressed in \cref{lem:mech_time_reversal}. It lies in the least action principle energy space we want to search, and we address the search space immediately below.

\paragraph{Search space for the Hamiltonian.}

\begin{figure}[htbp]
\centering
\begin{tikzpicture}[scale=0.7]

\begin{scope}[shift={(-7.5,2)}]
    \draw[thick] (-1.5,-1.5) rectangle (1.5,1.5);
    \node[above] at (0,1.7) {\textbf{Goal Attraction (Frame)}};
    \fill[red] (0,0) circle (3pt) node[below] {\tiny robot};
    \fill[blue] (1,1) circle (3pt) node[above] {\tiny goal};
    \draw[->,thick,red] (0.1,0.1) -- (0.8,0.8);
    \node at (0,-2) {\scriptsize $E_{\mathrm{goal}}(\vec{c},\mathbf{x}_g) = \|\vec{c}-\mathbf{x}_g\|^2$};
\end{scope}

\begin{scope}[shift={(-2.5,2)}]
    \draw[thick] (-1.5,-1.5) rectangle (1.5,1.5);
    \node[above] at (0,1.7) {\textbf{Sensor Cost (Sensor)}};
    \fill[red] (0,0) circle (3pt);
    \draw[blue,dashed] (0,0) ellipse (0.8 and 0.6);
    \node[blue] at (0,-0.9) {\tiny sensor region};
    \node at (0,-2) {\scriptsize $E_{\mathrm{sensor}}(y_t)=\|y_t\|_{S}^2$};
\end{scope}

\begin{scope}[shift={(2.5,2)}]
    \draw[thick] (-1.5,-1.5) rectangle (1.5,1.5);
    \node[above] at (0,1.7) {\textbf{Deformation (Shape)}};
    \draw[red,thick] (0,0) circle (0.9);   
    \draw[red,dashed] (0,0) circle (0.5); 
    \node at (0,-2) {\scriptsize $E_{\mathrm{obj}}(q)$};
\end{scope}

\begin{scope}[shift={(7.5,2)}]
    \draw[thick] (-1.5,-1.5) rectangle (1.5,1.5);
    \node[above] at (0,1.7) {\textbf{Collision Barriers}};
    \fill[red] (0,0) circle (3pt);
    \fill[gray] (0.8,0.3) circle (0.3);
    \fill[gray] (-0.6,-0.4) circle (0.25);
    \draw[->,purple] (0.05,0.02) -- (0.3,0.15);
    \node at (0,-2) {\scriptsize $\sum_{i\in\mathcal{C}_t} b(d_i(q;\mathcal{E}))$};
\end{scope}

\end{tikzpicture}
\caption{
Policy-aligned energy decomposition in GRL--SNAM.
The sensor policy exposes a sensor configuration cost $E_{\mathrm{sensor}}$, 
the frame policy induces goal attraction $E_{\mathrm{goal}}$, 
and the shape policy contributes deformation energy $E_{\mathrm{obj}}$, 
while all modules share collision barrier terms $b(d_i)$ around active constraints $\mathcal{C}_t$.
A meta-policy $g_\xi$ assigns weights $(\beta,\lambda,\{\alpha_i\})$ to these components to form the surrogate potential
$\mathcal{R}(q;\eta_\xi^t,\mathcal{E}) = E_{\mathrm{sensor}} + \beta^t E_{\mathrm{goal}} + \lambda^t E_{\mathrm{obj}} + \sum_{i\in\mathcal{C}_t} \alpha_i^t b(d_i)$
used in the Hamiltonian dynamics.
}
\label{fig:energy_decomposition}
\end{figure}
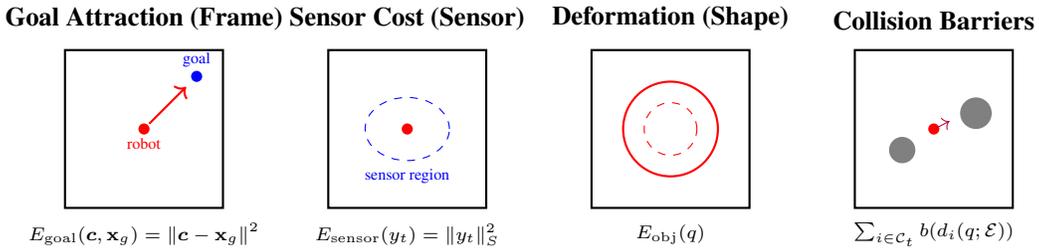

    The Hamiltonian defined in \cref{eq:def:reduced:hamiltonian:grl-snam} is a function on the cotangent bundle $T^{*}Q$:
\[
H \in \mathscr{H}:=\big\{\,H(q,p;\mathcal{E})=\tfrac12\,p^\top M^{-1}p\;+\;\mathcal{R}(q;\mathcal{E}) 
\ \big|\ M\in \mathbb{S}_{++}^2,\ \mathcal{R}\in \mathscr{R} \,\big\}.
\]
We regard $\mathscr{R}$ as a Hilbert space containing admissible potentials. For 2d navigation we \emph{restrict} the potential space $\mathscr{R}\cong \mathbb{R}^{\vert \xi\vert}$ by defining a parametrized map $\mathcal E\!\to\!\eta_{\xi}(\mathcal E)$ producing
nonnegative dual weights that shape the primal potential. This particular choice is compatible with soft barrier potential addition addressed in \cref{eq:def:reduced:hamiltonian:grl-snam}. In general, each energy
term may itself be parametrized:
\[
H(q,p;\omega,\xi,\mathcal E)\;=\;\tfrac12\,p^\top M(\omega_M)^{-1}p
\;+\;\mathcal{R}(q;\omega,\eta_{\xi}(\mathcal E)),
\]
\vspace{-0.75ex}
\begin{equation}
\begin{aligned}
\mathcal{R}(q;\omega, \eta_{\xi}(\mathcal E))\;&=\;
E_{\mathrm{sensor}}(q;\mathcal{E},\omega_y)+ 
\beta(\mathcal E)\,E_{\mathrm{goal}}(q;\mathcal E,\omega_g)
+\lambda(\mathcal E)\,E_{\mathrm{obj}}(q;\omega_d)
\\
&+\sum_{i\in\mathcal C(\mathcal E,q)} \alpha_i(\mathcal E)\,
b\big(d_i(q;\mathcal E);\omega_b\big),
\end{aligned}
\end{equation}
with $\eta_{\xi}(\mathcal E)\!=\!(\beta(\mathcal E),\lambda(\mathcal E),\{\alpha_i(\mathcal E)\})\in\mathbb R_+^{\,m(\mathcal E)}$.
Here $\omega\!=\!(\omega_y,\omega_M,\omega_g,\omega_d,\omega_b)$ are \emph{intra-term} parameters
(e.g.\ metric, goal shape, deformation model, barrier template), while $\eta_\xi$
learns the \emph{inter-term} tradeoffs by mapping the environment $\mathcal E$
to dual weights. The cardinality $m(\mathcal E)=2+|\mathcal C(\mathcal E,q)|$ is
environment/active-set dependent, so $\eta_\xi$ is implemented with a permutation-invariant
set encoder that outputs per-constraint scores
$\alpha_i(\mathcal E,t)\!\ge\!0$, together
with scalars $\beta(\mathcal E),\lambda(\mathcal E)\!\ge\!0$. The active set $\mathcal{C}(\mathcal{E},q):=\{i~|~d_i(q,\mathcal{E})\leq \hat{d}\}$ is discovered online by sensing. We explain role of each potential term in \cref{fig:energy_decomposition}.

\subsection{Navigator score fields and context-conditioned policies.}
\label{subsec:modular_architecture}

Rather than learning a single monolithic mapless navigation policy, we decompose each module of robots by a small collection of
\emph{module score fields} defined on phase space. Each module is a context-conditioned Hamiltonian system: its energy
induces a vector field, and the time-$\Delta t$ flow map of that vector field serves as the module's policy operator.

\paragraph{Sub-Modular score fields and induced policies.}
Let $\mathcal K=\{y,f,o\}$ index the sensor, frame, and shape modules. For each $k\in\mathcal K$, let $Q_k$ denote its
configuration manifold and let $\mathcal Z_k:=T^*\mathcal{Q}_k$ be its phase space with state $z_k=(q_k,p_k)$, where
$q_k\in \mathcal{Q}_k$ and $p_k\in T^*_{q_k}\mathcal{Q}_k$ is the conjugate costate.
Let $\theta_k\in\Theta_k$ denote the module-specific parameters, and assume the parameter sets are disjoint:
$\Theta_i\cap\Theta_j=\emptyset$ for all $i\neq j$.
Let $\mathcal E\in\mathscr E$ denote a local environment context (e.g., a local observation patch together with a goal descriptor).
We model the module energy by a differentiable Hamiltonian surrogate
\[
 h_k^{\theta_k}:\ \mathcal Z_k\times\mathscr E\times\mathbb R_{\ge 0}\to\mathbb R,\qquad (z_k,\mathcal E,t)\mapsto h_k^{\theta_k}(z_k;\mathcal E,t),
\]
where $t$ denotes the (local) integration time within a short rollout horizon.
Given a context-conditioned damping matrix $\Gamma_k^{\xi}(q_k;\mathcal E)\succeq 0$ and a nonconservative input
$G_k^{\xi}(q_k;\mathcal E)\,u_k^{\xi}(z_k,t;\mathcal E)$, { we define the \emph{modular score field} (deterministic drift) as the differential Hamiltonian 
{\small
\begin{equation}
\label{eq:module_score_field}
 s_k^{\theta_k,\xi}(z_k,t;\mathcal E)
 \!:=\!
 \begin{bmatrix}
 \nabla_{p_k} h_k^{\theta_k}(z_k;\mathcal E,t)\\
 -\nabla_{q_k} h_k^{\theta_k}(z_k;\mathcal E,t)
 \end{bmatrix}
 \!+\!
 \begin{bmatrix}
 0\\
 -\Gamma_k^{\xi}(q_k;\mathcal E)\,\nabla_{p_k} h_k^{\theta_k}(z_k;\mathcal E,t)+G_k^{\xi}(q_k;\mathcal E)\,u_k^{\xi}(z_k,t;\mathcal E)
 \end{bmatrix}.
\end{equation}
}
The modular stagewise} dynamics are the ODE $\dot z_k=s_k^{\theta_k,\xi}(z_k,t;\mathcal E)$.
Fix an integration rule with step size $\Delta t>0$ and denote by $\Phi_{k}^{\Delta t}(\cdot;\mathcal E):\mathcal Z_k\to\mathcal Z_k$
its time-$\Delta t$ flow map.
We define the (deterministic) \emph{context-conditioned modular policy} as the rollout operator
\begin{equation}
\label{eq:module_policy_operator}
 \pi_k^{\theta_k,\xi}(\mathcal E):=\Phi_{k}^{\Delta t}(\cdot;\mathcal E),\qquad z_{k,t+\Delta t}=\pi_k^{\theta_k,\xi}(\mathcal E)(z_{k,t}).
\end{equation}
If $\Gamma_k^{\xi}\equiv 0$ and $u_k^{\xi}\equiv 0$, then $s_k^{\theta_k,\xi}$ reduces to the Hamiltonian vector field and
$\pi_k^{\theta_k,\xi}(\mathcal E)$ is symplectic; otherwise it remains a (local) diffeomorphism under standard Lipschitz
regularity but is generally non-symplectic.

The disjointness of parameter sets implies $\partial s_k^{\theta_k,\xi}/\partial\theta_j\equiv 0$ for all $j\neq k$,
so the intra-modular energies can be trained in parallel while coordination is enforced through shared context-dependent
constraints and shared dual weights.

\paragraph{Template {Sub-modular} Hamiltonians and shared dual weights.}
We couple modules through a shared mechanical energy family and context-conditioned potential { reshaping}.
Let $\eta(\mathcal E)=\eta_{\xi}(\mathcal E)\in\mathbb R_+^{m(\mathcal E,t)}$ denote the nonnegative dual weights produced by
the navigator's context encoder, with components
$\eta(\mathcal E)=(\beta(\mathcal E),\lambda(\mathcal E),\{\alpha_i(\mathcal E,t)\}_{i\in\mathcal C_t})$.
The active constraint set $\mathcal C_t(\mathcal E,q):=\{i\mid d_i(q;\mathcal E)\ge 0\}$ is discovered online by sensing.
For each $k\in\mathcal K$ we use a mechanical template
\begin{equation}
\label{eq:module_template_hamiltonian}
 H_k(q_k,p_k;\mathcal E)
 =\tfrac12\,p_k^\top M^{-1}p_k+\mathcal R_k(q_k;\eta(\mathcal E),\mathcal E), \quad h_k^{\theta_k}(z_k;\mathcal{E},0)=H_k(z_k;\mathcal{E}).
\end{equation}
and instantiate the context-shaped potentials
\begin{align}
\label{eq:module_potentials}
 \mathcal R_y(q_y;\eta,\mathcal E)
 &=E_{\mathrm{sensor}}(q_y;\mathcal E,\omega_y)
 +\sum_{i\in\mathcal C_t(\mathcal E,q_y)}\alpha_i(\mathcal E,t)\,b\!\big(d_i(q;\mathcal E);\omega_b\big),\\
 \mathcal R_f(q_f;\eta,\mathcal E)
 &=\beta(\mathcal E)\,E_{\mathrm{goal}}(q_f;\mathcal E,\omega_g)
 +\sum_{i\in\mathcal C_t(\mathcal E,q_f)}\alpha_i(\mathcal E,t)\,b\!\big(d_i(q;\mathcal E);\omega_b\big),\\
 \mathcal R_o(q_o;\eta,\mathcal E)
 &=\lambda(\mathcal E)\,E_{\mathrm{obj}}(q_o;\omega_d)
 +\sum_{i\in\mathcal C_t(\mathcal E,q_o)}\alpha_i(\mathcal E,t)\,b\!\big(d_i(q;\mathcal E);\omega_b\big).
 \label{eq:module_template_potential:o}
\end{align}
Here $\omega=(\omega_y,\omega_g,\omega_d,\omega_b)$ collects intra-term parameters, while $\eta(\mathcal E)$ controls
inter-term tradeoffs via nonnegative weights.

\paragraph{{Sub-modular} responses and aggregation.}
Each module integrates its phase dynamics over a short horizon and returns a standardized response map $\mathsf R_k$  to the navigator. The navigator aggregates these responses to update $\eta_{\xi}(\mathcal E)$ (and, when enabled,
damping/correction parameters) and to generate stable trajectories across stages.

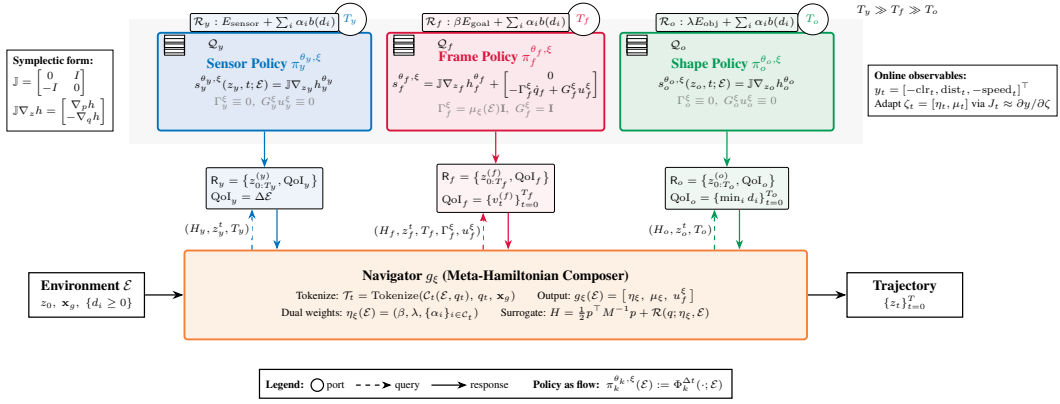
\begin{figure}[h]
\centering
\resizebox{\linewidth}{!}{%
\begin{tikzpicture}[
  node distance=0.5cm,
  font=\small,
  sensorcolor/.style={draw=RoyalBlue, fill=RoyalBlue!10},
  framecolor/.style ={draw=Crimson,   fill=Crimson!10},
  shapecolor/.style ={draw=ForestGreen,fill=ForestGreen!10},
  navcolor/.style   ={draw=Orange,    fill=Orange!10},
  band/.style       ={fill=black!3, draw=none},
  mod/.style   ={rectangle,rounded corners=2pt,draw,very thick,align=center,
                 minimum width=4.8cm,minimum height=2.2cm,#1},
  nav/.style   ={rectangle,rounded corners=2pt,draw,very thick,align=center,
                 minimum width=14.0cm,minimum height=2.0cm,#1},
  io/.style    ={circle,draw,thick,inner sep=0pt,minimum size=2.8mm,fill=white},
  qarr/.style  ={thick,dashed,-{Stealth[length=2.2mm]}},
  rarr/.style  ={thick,-{Stealth[length=2.2mm]}},
  chip/.style  ={rectangle,rounded corners=1pt,draw,thin,inner sep=2pt,
                 font=\scriptsize,fill=white},
  dsicon/.style={rectangle,draw,thin,fill=white,minimum width=4.5mm,minimum height=4.5mm},
  badge/.style ={circle,draw,thin,fill=white,minimum size=5mm,font=\scriptsize\bfseries},
  xtiny/.style ={font=\scriptsize},
  response/.style={rectangle,draw,thin,fill=white,rounded corners=1pt,font=\scriptsize,align=left,inner sep=3pt}
]

\coordinate (ColY) at (-5.2,0);   
\coordinate (ColF) at ( 0.0,0);   
\coordinate (ColO) at ( 5.2,0);   

\node[rectangle,band,minimum width=16.5cm,minimum height=2.8cm,anchor=center] at (0,6.0) {};

\node[mod=sensorcolor,anchor=center] (sensor) at ($(ColY)+(0,6.0)$) {
    {\bfseries\color{RoyalBlue}Sensor Policy $\pi_y^{\theta_y,\xi}$}\\[2pt]
    {\scriptsize $s_y^{\theta_y,\xi}(z_y,t;\mathcal{E})=\mathbb{J}\nabla_{z_y}h_y^{\theta_y}$}\\[1pt]
    {\scriptsize\color{gray} $\Gamma_y^\xi\equiv 0,\; G_y^\xi u_y^\xi\equiv 0$}
};

\node[mod=framecolor,anchor=center] (frame) at ($(ColF)+(0,6.0)$) {
    {\bfseries\color{Crimson}Frame Policy $\pi_f^{\theta_f,\xi}$}\\[1pt]
    {\scriptsize $s_f^{\theta_f,\xi}=\mathbb{J}\nabla_{z_f}h_f^{\theta_f}+\begin{bmatrix}0\\-\Gamma_f^\xi\dot{q}_f+G_f^\xi u_f^\xi\end{bmatrix}$}\\[1pt]
    {\scriptsize\color{gray} $\Gamma_f^\xi=\mu_\xi(\mathcal{E})\mathbf{I},\; G_f^\xi=\mathbf{I}$}
};

\node[mod=shapecolor,anchor=center] (shape) at ($(ColO)+(0,6.0)$) {
    {\bfseries\color{ForestGreen}Shape Policy $\pi_o^{\theta_o,\xi}$}\\[2pt]
    {\scriptsize $s_o^{\theta_o,\xi}(z_o,t;\mathcal{E})=\mathbb{J}\nabla_{z_o}h_o^{\theta_o}$}\\[1pt]
    {\scriptsize\color{gray} $\Gamma_o^\xi\equiv 0,\; G_o^\xi u_o^\xi\equiv 0$}
};

\foreach \M/\L in {sensor/$\mathcal{Q}_y$,frame/$\mathcal{Q}_f$,shape/$\mathcal{Q}_o$}{
  \node[dsicon] (ds-\M) at ([xshift=4mm,yshift=-3mm]\M.north west) {};
  \draw[thin] (ds-\M.south west)++(0,0.6mm) rectangle ++(4.5mm,0.8mm);
  \draw[thin] (ds-\M.south west)++(0,1.8mm) rectangle ++(4.5mm,0.8mm);
  \draw[thin] (ds-\M.south west)++(0,3.0mm) rectangle ++(4.5mm,0.8mm);
  \node[xtiny] at ([xshift=7mm]ds-\M.east) {\L};
}

\node[badge,text=RoyalBlue]   at ([xshift=-14pt,yshift=8pt]sensor.north east) {$T_y$};
\node[badge,text=Crimson]     at ([xshift=-14pt,yshift=8pt]frame.north east)  {$T_f$};
\node[badge,text=ForestGreen] at ([xshift=-14pt,yshift=8pt]shape.north east)  {$T_o$};

\node[xtiny,anchor=west] at (8.0,7.6) {$T_y \gg T_f \gg T_o$};

\node[chip,fill=RoyalBlue!5] at ([yshift=6pt]sensor.north) {$\mathcal{R}_y: E_{\mathrm{sensor}} + \sum_i\alpha_i b(d_i)$};
\node[chip,fill=Crimson!5] at ([yshift=6pt]frame.north) {$\mathcal{R}_f: \beta E_{\mathrm{goal}} + \sum_i\alpha_i b(d_i)$};
\node[chip,fill=ForestGreen!5] at ([yshift=6pt]shape.north) {$\mathcal{R}_o: \lambda E_{\mathrm{obj}} + \sum_i\alpha_i b(d_i)$};

\node[response,fill=RoyalBlue!5] (Ry) at ($(ColY)+(0,3.6)$) {
    $\mathsf{R}_y=\big\{z_{0:T_y}^{(y)},\mathrm{QoI}_y\big\}$\\
    $\mathrm{QoI}_y=\Delta\mathcal{E}$
};
\node[response,fill=Crimson!5] (Rf) at ($(ColF)+(0,3.6)$) {
    $\mathsf{R}_f=\big\{z_{0:T_f}^{(f)},\mathrm{QoI}_f\big\}$\\
    $\mathrm{QoI}_f=\{v_t^{(f)}\}_{t=0}^{T_f}$
};
\node[response,fill=ForestGreen!5] (Ro) at ($(ColO)+(0,3.6)$) {
    $\mathsf{R}_o=\big\{z_{0:T_o}^{(o)},\mathrm{QoI}_o\big\}$\\
    $\mathrm{QoI}_o=\{\min_i d_i\}_{t=0}^{T_o}$
};

\node[nav=navcolor,anchor=center] (nav) at ($(0,1.2)$) {
  {\bfseries Navigator $g_{\xi}$ (Meta-Hamiltonian Composer)}\\[2pt]
  {\scriptsize Tokenize: $\mathcal{T}_t=\mathrm{Tokenize}(\mathcal{C}_t(\mathcal{E},q_t),\,q_t,\,\mathbf{x}_g)$\quad\quad Output: $g_\xi(\mathcal{E})=\big[\,\eta_\xi,\;\mu_\xi,\;u_f^\xi\,\big]$}\\[1pt]
  {\scriptsize Dual weights: $\eta_\xi(\mathcal{E})=(\beta,\lambda,\{\alpha_i\}_{i\in\mathcal{C}_t})$\quad\quad Surrogate: $H=\tfrac{1}{2}p^{\top}M^{-1}p+\mathcal{R}(q;\eta_\xi,\mathcal{E})$}
};

\draw[qarr,draw=RoyalBlue]
  ([xshift=-8pt]nav.north-|ColY) -- node[xtiny,fill=white,inner sep=1pt,pos=0.5,left]
  {$(H_y,z_y^t,T_y)$} ([xshift=-8pt]Ry.south-|ColY);

\draw[qarr,draw=Crimson]
  ([xshift=-8pt]nav.north-|ColF) -- node[xtiny,fill=white,inner sep=1pt,pos=0.5,left]
  {$(H_f,z_f^t,T_f,\Gamma_f^\xi,u_f^\xi)$} ([xshift=-8pt]Rf.south-|ColF);

\draw[qarr,draw=ForestGreen]
  ([xshift=-8pt]nav.north-|ColO) -- node[xtiny,fill=white,inner sep=1pt,pos=0.5,left]
  {$(H_o,z_o^t,T_o)$} ([xshift=-8pt]Ro.south-|ColO);

\draw[rarr,draw=RoyalBlue]
  ([xshift=8pt]Ry.south-|ColY) -- ([xshift=8pt]nav.north-|ColY);

\draw[rarr,draw=Crimson]
  ([xshift=8pt]Rf.south-|ColF) -- ([xshift=8pt]nav.north-|ColF);

\draw[rarr,draw=ForestGreen]
  ([xshift=8pt]Ro.south-|ColO) -- ([xshift=8pt]nav.north-|ColO);

\draw[rarr,draw=RoyalBlue] (sensor.south) -- (Ry.north);
\draw[rarr,draw=Crimson] (frame.south) -- (Rf.north);
\draw[rarr,draw=ForestGreen] (shape.south) -- (Ro.north);

\node[rectangle,draw,thick,fill=white,minimum width=2.6cm,minimum height=1.0cm,align=center]
      (env) at (-9.2,1.2) {{\bfseries Environment $\mathcal{E}$}\\ {\scriptsize $z_0,\;\mathbf{x}_g,\;\{d_i\geq 0\}$}};
\node[rectangle,draw,thick,fill=white,minimum width=2.6cm,minimum height=1.0cm,align=center]
      (traj) at ( 9.2,1.2) {{\bfseries Trajectory}\\ {\scriptsize $\{z_t\}_{t=0}^{T}$}};

\draw[thick,-{Stealth[length=2.5mm]}] (env.east) -- (nav.west);
\draw[thick,-{Stealth[length=2.5mm]}] (nav.east) -- (traj.west);

\node[rectangle,draw,thin,fill=white,align=left,font=\scriptsize,inner sep=4pt]
      at (-9.8,5.8) {
       \textbf{Symplectic form:}\\[1pt]
       $\mathbb{J}=\begin{bmatrix}0 & I\\-I & 0\end{bmatrix}$\\[1pt]
       $\mathbb{J}\nabla_z h = \begin{bmatrix}\nabla_p h\\-\nabla_q h\end{bmatrix}$
      };

\node[rectangle,draw,thin,fill=white,align=left,font=\scriptsize,inner sep=4pt]
      at (10.5,5.8) {
       \textbf{Online observables:}\\[1pt]
       $y_t=[-\mathrm{clr}_t,\mathrm{dist}_t,-\mathrm{speed}_t]^\top$\\[1pt]
       Adapt $\zeta_t=[\eta_t,\mu_t]$ via $J_t\approx\partial y/\partial\zeta$
      };

\node[rectangle,draw,thick,fill=white,align=left,font=\scriptsize,inner sep=4pt]
      at (0,-0.8) {
       \textbf{Legend:}\;
       \tikz[baseline=-0.5ex]{\node[io]{};} port\quad
       \tikz[baseline=-0.5ex]{\draw[qarr] (0,0)--(8mm,0);} query\quad
       \tikz[baseline=-0.5ex]{\draw[rarr] (0,0)--(8mm,0);} response
       \qquad
       \textbf{Policy as flow:}\; $\pi_k^{\theta_k,\xi}(\mathcal{E}):=\Phi_k^{\Delta t}(\cdot;\mathcal{E})$
      };

\end{tikzpicture}}
\caption{%
Modular score-field architecture for GRL-SNAM.
The Navigator $g_\xi$ tokenizes active constraints $\mathcal{C}_t$, state $q_t$, and goal $\mathbf{x}_g$, then outputs dual weights $\eta_\xi=(\beta,\lambda,\{\alpha_i\})$, friction $\mu_\xi$, and port correction $u_f^\xi$.
Each module $k\in\{y,f,o\}$ integrates its score field $s_k^{\theta_k,\xi}=\mathbb{J}\nabla_{z_k}h_k^{\theta_k}$ (plus non-conservative terms for the frame module) over horizon $T_k$ and returns response $\mathsf{R}_k$ with phase-space rollouts and QoIs.
The frame module uniquely receives dissipation $\Gamma_f^\xi$ and port input $u_f^\xi$; sensor and shape modules have purely symplectic dynamics.
}
\label{fig:independent-score-architecture}
\end{figure}

\subsection{GRL-SNAM offline training and online adaptation}
\label{subsec:navigator_meta_learner}

The GRL-SNAM operates as a meta-learning system that coordinates multi-scale policies by learning how to update their Hamiltonian energy functions rather than directly manipulating phase space states.

\paragraph{Choice of Hamiltonian and Hamiltonian dynamics}
Throughout the paper we further simplify our navigator's Hamiltonian surrogate by assuming fixed $\omega$ throughout all environment when we primarily focus on learning to optimize the
environment-indexed linear cone generated by task energies. The potential energy is denoted as follows and is shown in Figure \ref{fig:energy_decomposition}:  
\begin{equation}
\begin{aligned}
\mathcal{R}(q; \eta(\mathcal E)) &= \underbrace{\|\vec{y}\|_{S}^2}_{\text{Sensor Cost ($E_{sensor}$)}}
\;+\;
\underbrace{\beta\|\vec{c}-\mathbf{x}_g\|_2^2}_{\text{Goal Attraction ($E_{goal}$)}} \\ &\;+\;
\underbrace{\lambda E_{obj}(q(t))}_{\text{Deformation Energy ($E_{obj}$)}} \;+\;
\underbrace{\sum_{i\in\mathcal{C}_t(\mathcal{E},q)} \alpha_i b(d_i,\hat{d})}_{\text{Collision Barriers (FPE)}} .
\end{aligned}
\end{equation}

In addition, we assume friction and additional forces (e.g. derived from safety constraints) are happened in FPE submodule only, and we pick a particular parametrization as:
\begin{equation}
\begin{aligned}
    \Gamma_y^{\xi} &\equiv 0 ,\quad G_y^{\xi}\equiv 0,\quad u_y^{\xi}\equiv 0, \\
    \Gamma_f^{\xi} &=\mu_\xi(\mathcal{E})\mathbf{I},\quad G_f^{\xi}=\mathbf{I}, \quad u_f^{\xi}\ne 0,\\
    \Gamma_o^{\xi} &\equiv 0 ,\quad G_o^{\xi}\equiv 0,\quad u_o^{\xi}\equiv 0, 
\end{aligned}
\end{equation}

Thus the meta policy $g_\xi$ \emph{directly} produces the cone coordinates of potentials and non-conservative forces correction in FPE module as a friction and a port correction term:
\begin{equation}
g_{\xi}(\mathcal{E})=\big[\;\eta_{\xi}(\mathcal{E}), \mu_\xi(\mathcal{E}) ,u_f^{\xi}\;],    
\label{eq:def:score:function:g}
\end{equation}

which defines the potential—and therefore the generalized force —for
the stagewise motion-planning Hamiltonian rollouts. Learning $g_\xi$ gives a \emph{meta-policy} that maps environments to energy
weights, i.e.\ a stagewise bilevel scheme where the inner layer optimizes motion
under $H_k$ and the outer layer trains $\xi$ so that scenario-level QoIs/constraints
are satisfied across environments.

\begin{figure}[htbp]
    \centering
    \includegraphics[width=0.72\linewidth]{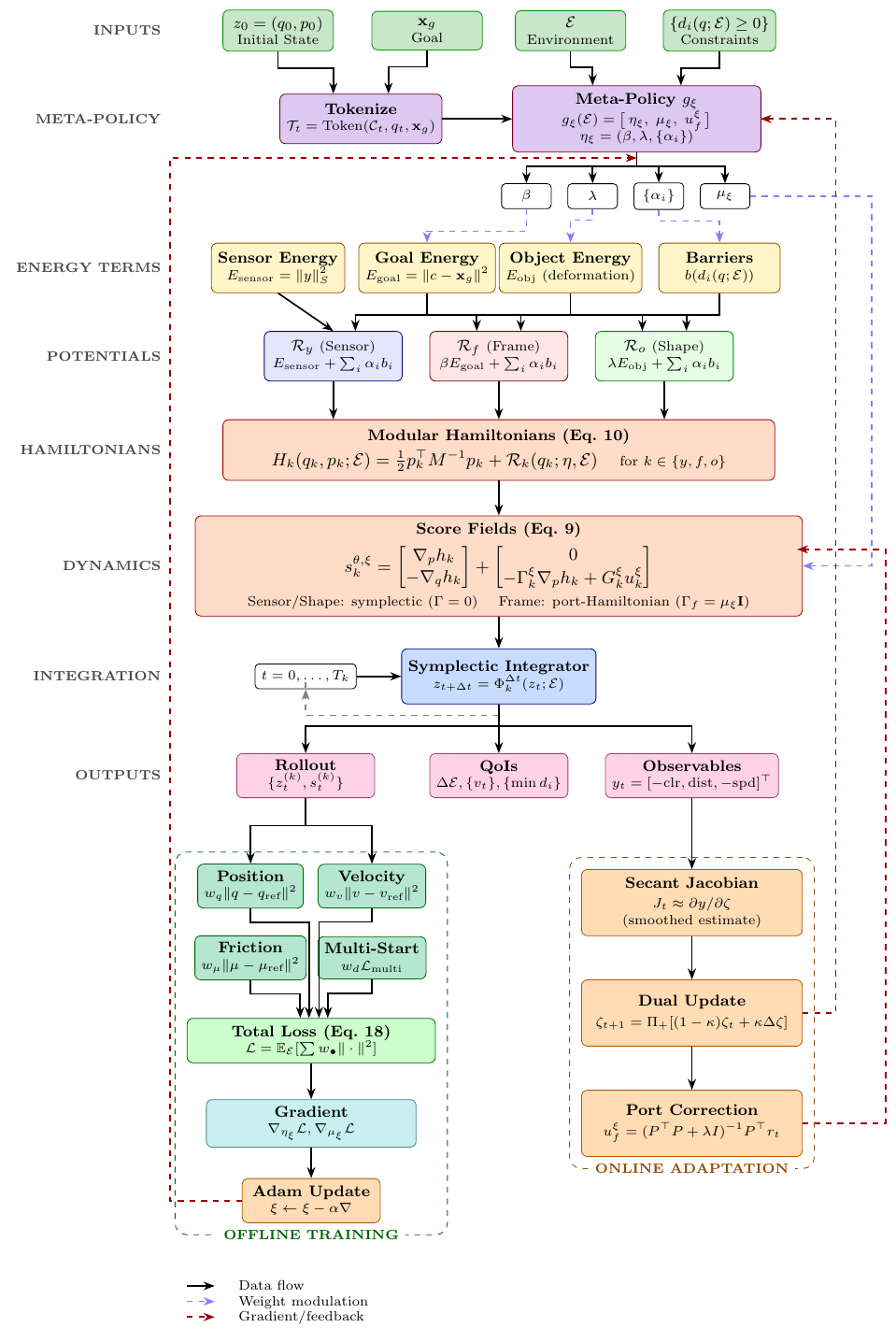}
\caption{\textbf{GRL--SNAM computational flow: sub-modular Hamiltonian structure with offline training and online adaptation.}
Given the current state \(z_0=(\vec{q}_0,p_0)\), goal \(\mathbf{x}_g\), and locally sensed environment context \(\mathcal{E}\) (obstacle constraints \(\{d_i(q;\mathcal{E})\ge 0\}\)), the navigator \(g_\xi\) tokenizes the active local information and outputs a small set of \emph{meta-parameters}: the nonnegative energy weights \(\eta_\xi=(\beta,\lambda,\{\alpha_i\})\), the frame dissipation \(\mu_\xi\), and an optional frame port input \(u_f^\xi\).
These parameters modulate a fixed library of interpretable energy terms (sensor cost, goal attraction, deformation, and barrier penalties) to form module-specific potentials \(\mathcal{R}_y,\mathcal{R}_f,\mathcal{R}_o\), which define the corresponding modular Hamiltonians
\(H_k(\vec{q}_k,p_k;\mathcal{E})=\tfrac12 p_k^\top M^{-1}p_k + \mathcal{R}_k(\vec{q}_k;\eta_\xi,\mathcal{E})\) (c.f.\ \cref{eq:module_template_hamiltonian}).
Each module induces a score (drift) field; sensor and shape follow conservative (symplectic) dynamics, while the frame module additionally includes dissipation and a nonconservative port term.
A symplectic integrator rolls out the dynamics for a short horizon, producing trajectories and quantities of interest (environment updates, velocities, and clearances) as well as low-dimensional observables \(y_t=[-\mathrm{clr}_t,\ \mathrm{dist}_t,\ -\mathrm{spd}_t]^\top\).
\emph{Offline training} (left branch) learns the mapping from context to meta-parameters by minimizing a supervised loss on position/velocity/friction, augmented with a multi-start robustness term near obstacles.
\emph{Online adaptation} (right branch) updates only the current meta-parameters using a smoothed secant Jacobian estimate to track the desired observables, and uses the residual to compute a corrective port input when energy reweighting alone is insufficient.
Arrows indicate (i) forward data flow, (ii) weight modulation from \(g_\xi\) into energies, and (iii) gradient/feedback pathways connecting rollouts to offline updates and online corrections.}
    \label{fig:full:loop:grl:snam}
\end{figure}

\paragraph{Learning $\eta_{\xi}(\mathcal{E}), \mu_\xi(\mathcal{E})$ through observation of Hamiltonian dynamics.}
Each submodule exposes a response map that provides integrated dynamics rollout and additional quantity of interests (QoI) as the feedback:
\[
\mathsf{R}_k:\ \big(H_k;\theta_k,\mathcal E, \xi)\ \longmapsto\ 
\{z_t^{(k)}, \mathrm{QoI}_k\}_{t=0}^{T_k},\qquad k\in\mathcal{K},
\]
where $T_k$ refers time scale for each submodule. To be more specific:
\begin{equation}
    \mathrm{QoI}_{y}=\Delta\mathcal{E}, \quad \mathrm{QoI}_{f}=\{v_t^{(f)}:=M_f^{-1}p_t^{(f)}\}_{t=0}^{T_f} ,\quad \mathrm{QoI}_{o}= \{\min_{i\in\mathcal{C}_t(\mathcal{E},q)} d_i(q_t^{(o)};\mathcal{E})\}_{t=0}^{T_o}.
\end{equation}

Namely, feedback QoIs are: environmental update, velocity observation, and min distance clearance (and thus collision violation). Additional QoIs, that can be deduced from $z_t^{(k)},s_t^{(k)}$, are not stated here explicitly. To train a policy that output $\eta_{\xi}(\mathcal{E})$ and $\mu_{\xi}(\mathcal{E})$ with different environments, we propose to minimize
\begin{equation}
\label{eq:meta_loss}
\begin{aligned}
\mathcal L(\eta_\xi,\mu_\xi)
= \mathbb{E}_{\mathcal{E}}\left[
w_{q}\big\|\vec{q}-\vec{q}_{ref}\big\|_2^2
\;+\;
w_{v}\big\|\vec{v}-\vec{v}_{ref}\big\|_2^2
\;+\;
w_{\mu}\big\|\mu-\mu_{ref}\big\|_2^2
\;+\;
w_{d}\ \mathcal L_{\text{multi}}\right].
\end{aligned}
\end{equation}
where $\mathcal L_{\text{multi}}$ is a short multi-start robustness penalty that re-rolls from perturbed $q_t^{(k)}$ seeds near obstacles to discourage brittle $\eta_{\xi}(\mathcal{E})$ and $\mu_{\xi}(\mathcal{E})$ (details are addressed in Algorithm \ref{alg:meta_training}), and $w_{\bullet}\!\ge\!0$ are user-input hyperparameters. The training via \cref{eq:meta_loss} can be conducted offline, component-wise, or even fine-tuned online, but we state that it is important to fully utilize the instantaneous response from a real navigation scenario which provides the scheme of per-scneario online correction even when $\eta_\xi(\mathcal{E})$ and $\mu_\xi(\mathcal{E})$ are properly trained under large-scale simulated dataset with reference potentials.

\begin{algorithm}[htbp]
\caption{Offline GRL-SNAM training of $\eta(\mathcal{E})$ and $\mu_{\xi}(\mathcal{E})$}
\label{alg:meta_training}
\begin{algorithmic}[1]
\Require dataset of different environments $\{\mathcal{E}, \alpha_{ref},\beta_{ref},\lambda_{ref},\mu_{ref}\}\in \mathcal D$, step range $T\!\in\!\{2,3,4,5,6\}$, short rollout trials $M$, time step $\Delta t$, weights $w_{\bullet}$
\State Initialize $\xi$
\For{epoch $=1,2,\dots$}
  \For{batch $\mathcal B\subset\mathcal D$}
    \State Sample $(q_0,p_0)$ per scene $\mathcal{E}$ in $\mathcal{B}$
    \State Build \textbf{tokens} from active constraints $\mathcal{C}_t(\mathcal{E},q_0)$, goal $\vec{x}_g$, and current $(q_0,p_0)$
     \State $(\{\alpha_j\},\beta)\gets \eta_\xi(\textbf{tokens}),\quad \mu\gets \mu_{\xi}(\textbf{tokens})$
     \For{$m=1,2,\dots,M$}
    \State Sample a near obstacle point $\tilde{q}_0$, initialize $\tilde{p}_0$ towards nearest obstacle at $\tilde{q}_0$.
    \State Integrate \cref{eq:module_score_field} for $T$ steps from $(\tilde{q}_0,\tilde{p}_0)$. \State \Comment{e.g. {Leapfrog}}
    \State Compute $clr\gets \min_{i,t} d_i(\tilde{q}_t;\mathcal{E})$
    \State Compute $\mathcal L_{\text{multi}}\gets\mathcal{L}_{\text{multi}}+\frac{1}{M}b\!\left(r_{min}-\mathrm{clr},\hat{d}\right).$ \State \Comment{$r_{min}$ refers minimal radius of robot object.}
    \EndFor
    \State Integrate \cref{eq:module_score_field} for $T_k$ steps from $(q_0,p_0)$
    \State Evaluate loss $\mathcal L(\eta_\xi,\mu_\xi)$ in \cref{eq:meta_loss}\State Update $\eta_\xi,\mu_\xi$ using  $\nabla_{\eta_\xi}\mathcal L$ and $\nabla_{\mu_\xi}\mathcal L$   \Comment{ Uncertainty hyperpaprameter weighted}
  \EndFor
\EndFor
\end{algorithmic}
\end{algorithm}

\begin{figure}[htbp]
  \centering
  \includegraphics[width=\linewidth]{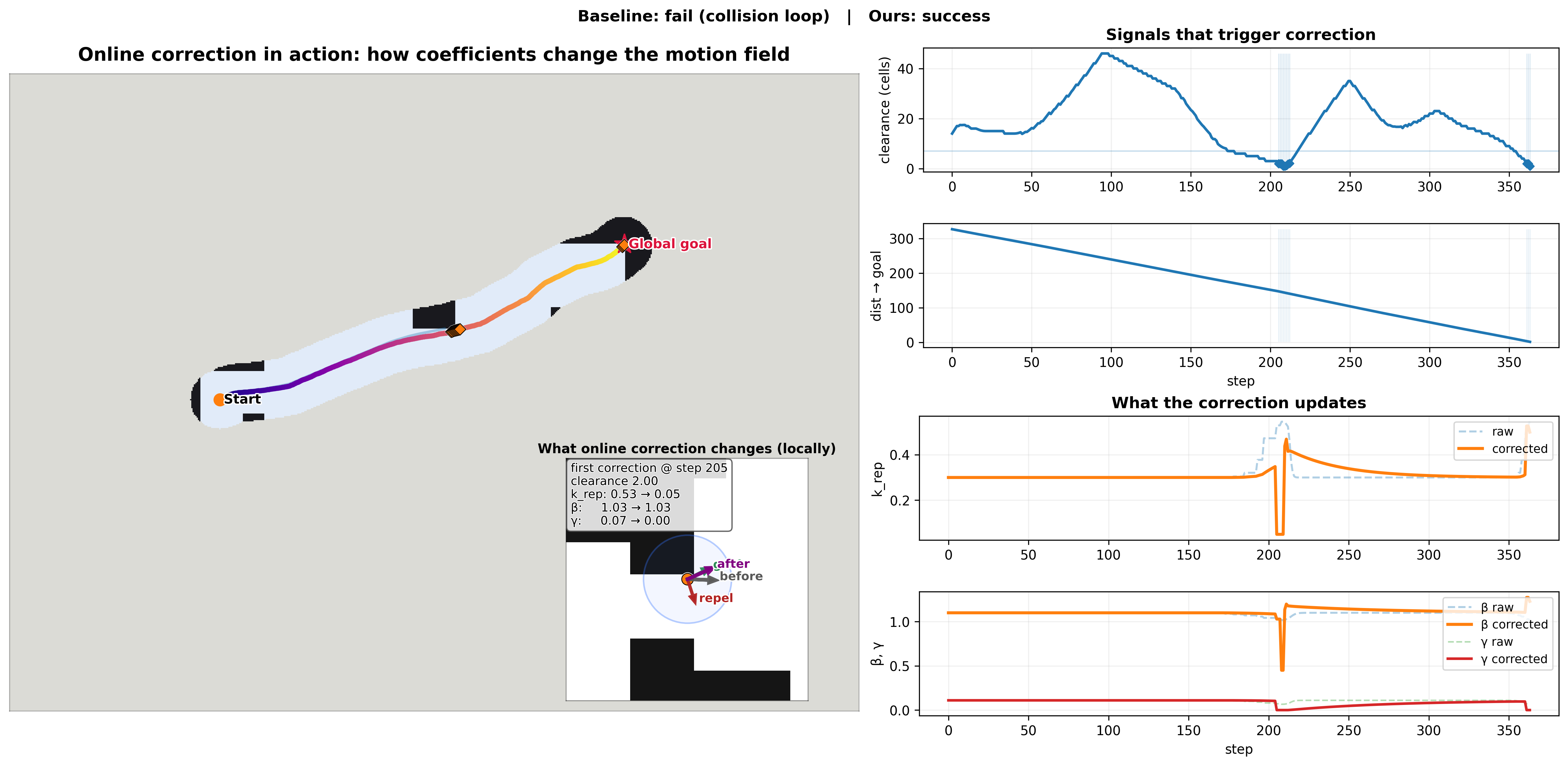}
\caption{\textbf{Online correction mechanism (Sec.~\ref{subsec:navigator_meta_learner}).}
{Left: a representative rollout where the frozen baseline enters a collision loop, while GRL--SNAM reaches the global goal by applying local corrections at detected failure points; the inset visualizes the first correction as a change in the local motion field (goal-seeking plus obstacle repulsion) around the active obstacles.
Right: the observable vector \(y_t=[-\mathrm{clr}_t,\ \mathrm{dist}_t,\ -\mathrm{speed}_t]^\top\) used for triggering/adaptation, with shaded bands marking correction events.
Bottom-right: the corresponding parameter updates i.e.; the dual weights \(\tilde{\eta}_t=[\beta_t,\ \lambda_t,\ \alpha_{t,\mathcal I_t}]^\top\) (active-set barrier weights) and dissipation/port terms \((\mu_t,\ u_{f,t}^\xi)\) show how the navigator performs lightweight, stage-local updates to the Hamiltonian-shaped policy to restore safety (clearance) while maintaining goal progress.}}
  \label{fig:online-correction-mechanism}
\end{figure}

\paragraph{Online Adaptation of $g_{\xi}(\mathcal{E})$ via QoIs}
We state how response map for each environment can yield a correction term under online navigation scenario. Given response $\mathsf{R}_k$ at time $t$ we construct an \emph{observable} measurement vector and its reference goal:
\[
y_t \;=\; \begin{bmatrix}
-\,\mathrm{clr}_t \\[2pt]
\mathrm{dist}_t \\[2pt]
-\,\mathrm{speed}_t
\end{bmatrix}\in\mathbb R^3,
\qquad
y_t^\star \;=\; \begin{bmatrix}
-\,m_{\mathrm{safe}}\\[2pt]
\mathrm{dist}_t-\varepsilon_{\mathrm{prog}}\\[2pt]
-\,\max\big(\mathrm{speed}_t,\ \mathbf 1_{\{\mathrm{clr}_t\ge m_{\mathrm{safe}}\}}\,v_{\min}\big)
\end{bmatrix},
\]
where $\mathrm{clr}$ is the minimum clearance to inflated obstacles, $\mathrm{dist}$ is goal distance (to the global goal point), and $\mathrm{speed}=\|v\|$.  
We update only the \emph{active} barrier weights by selecting an index set $\mathcal I_t$ that represents nearby obstacles. Define the parameter vector
\[
\tilde{\eta}_t\;=\;\begin{bmatrix}\beta_t\\ \lambda_t\\\alpha_{t,\mathcal I_t} \end{bmatrix}\in\mathbb R^{2+|\mathcal I_t|}_{\,+}, \quad \zeta_t=[\tilde{\eta}_t,\mu_t].
\]

We denote an estimator of Jabocian $J_t=\frac{\partial y}{\partial \zeta_t}$ as  
\[
\widetilde J_t\;=\;\frac{(y_t-y_{t-1})(\zeta_t-\zeta_{t-1})^\top}{\|\zeta_t-\zeta_{t-1}\|_2^2+\varepsilon},
\qquad
J_t\;=\;\rho\,J_{t-1}+(1-\rho)\,\widetilde J_t,
\]
with smoothing $\rho\in[0,1)$ and $\varepsilon>0$. Then, given the desired observable change $\Delta y^{\mathrm{des}}_t:=y_t^\star-y_t$, one can upate meta-policy parameter via a Tiknohov-regularized least square steps:
\[
\Delta\zeta_t \;=\; \arg\min_{\Delta\zeta}\ \big\|J_t\Delta\zeta-\Delta y^{\mathrm{des}}_t\big\|_2^2
\;+\;\lambda\|\Delta\zeta\|_2^2,
\quad\Rightarrow\quad
\Delta\zeta_t=(J_t^\top J_t+\lambda I)^{-1}J_t^\top\Delta y^{\mathrm{des}}_t.
\]
This yields updated $\beta_{t+1}$, $\gamma_{t+1}$, and $\alpha_{t+1,i}$ for $i\in\mathcal I_t$ (inactive weights keep their previous values).
\[
\zeta_{t+1}\;=\;\Pi_{\mathbb R_+}\big((1-\vec{\kappa})\,\zeta_t+ \vec{\kappa}\,\Delta\zeta_{t}\big),\quad \vec{\kappa}_i\in[0,1).
\]

Since the update is a result of least square step, there exists residual:
\[
r_t \;=\; \Delta y^{\mathrm{des}}_t - J_t\,\Delta\zeta_t.
\]
The online port correction term can amend the energy change by solving another least square problem given port–observable sensitivity $P_t\approx\partial y/\partial u_f$:
\[
u_{f,t}^{\xi}
\;=\;\arg\min_{u\in\mathcal U}\ \big\|P_t u - r_t\big\|_2^2+\lambda_u\|u\|_2^2
\;=\;(P_t^\top P_t+\lambda_u I)^{-1}P_t^\top r_t,
\]
followed by componentwise clipping to a feasible box $\mathcal U$.
A simple choice is to use $P_t=\operatorname{diag}(0,0,\kappa_v)$ so that the port primarily regulates speed while the energy weights steer clearance and goal progress. We comment that richer $P_t$ can be learned online by the same secant recipe as $J_t$.

\paragraph{Multi-Scale Temporal Coordination}
\label{subsec:temporal_coordination}

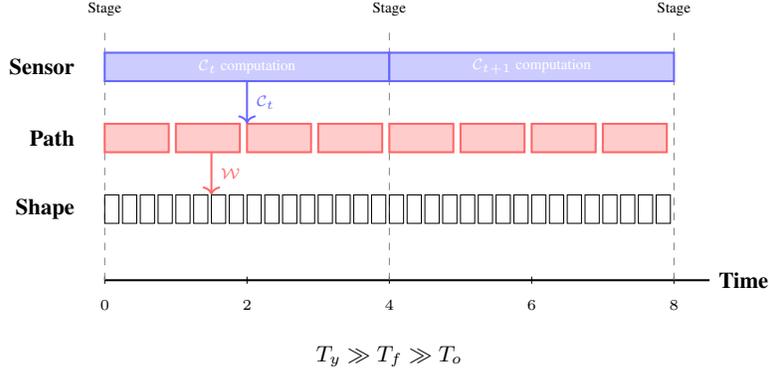
\begin{figure}[ht]
\centering
\resizebox{0.75\textwidth}{!}{
\begin{tikzpicture}[
  every node/.style={font=\footnotesize},
  sensor/.style={fill=blue!20, draw=blue!60, thick},
  path/.style={fill=red!20, draw=red!60, thick},
  shape/.style={fill=green!20, draw=green!60, thick},
  arrow/.style={->, thick, draw=gray!70},
]

\draw[thick] (0,0) -- (8.5,0) node[right] {\textbf{Time}};
\foreach \x in {0,2,4,6,8} {
  \draw (\x,-0.05) -- (\x,0.05);
  \node[below] at (\x,-0.15) {\tiny $\x$};
}

\foreach \x in {0,4,8} {
  \draw[dashed, gray] (\x,0) -- (\x,3.5);
  \node[above] at (\x,3.6) {\tiny Stage};
}

\node[left] at (-0.3,3) {\textbf{Sensor}};
\draw[sensor] (0,2.8) rectangle (4,3.2);
\node[white] at (2,3) {\tiny $\mathcal{C}_t$ computation};
\draw[sensor] (4,2.8) rectangle (8,3.2);
\node[white] at (6,3) {\tiny $\mathcal{C}_{t+1}$ computation};

\node[left] at (-0.3,2) {\textbf{Path}};
\foreach \x in {0,1,2,3} {
  \draw[path] (\x,1.8) rectangle (\x+0.9,2.2);
}
\foreach \x in {4,5,6,7} {
  \draw[path] (\x,1.8) rectangle (\x+0.9,2.2);
}

\node[left] at (-0.3,1) {\textbf{Shape}};
\foreach \x in {0,0.25,0.5,0.75,1,1.25,1.5,1.75,2,2.25,2.5,2.75,3,3.25,3.5,3.75} {
  \draw(\x,0.8) rectangle (\x+0.2,1.2);
}
\foreach \x in {4,4.25,4.5,4.75,5,5.25,5.5,5.75,6,6.25,6.5,6.75,7,7.25,7.5,7.75} {
  \draw(\x,0.8) rectangle (\x+0.2,1.2);
}

\draw[arrow, blue!60] (2,2.8) -- (2,2.2) node[midway,right] {\tiny $\mathcal{C}_t$};
\draw[arrow, red!60] (1.5,1.8) -- (1.5,1.2) node[midway,right] {\tiny $\mathcal{W}$};

\node[below] at (4,-0.8) {$T_{y} \gg T_{f} \gg T_{o}$};
\end{tikzpicture}
}
\caption{Temporal hierarchy. Sensor policy operates at low frequency (once per stage), establishing environmental constraints $\mathcal{C}_t$. Path policy operates at medium frequency within each stage, computing waypoints $\mathcal{W}$. Shape policy operates at high frequency, continuously adapting at each integration step. This creates a natural hierarchy where slow sensor updates provide stable constraints for faster path and shape adaptations.}
\label{fig:temporal_scales}
\end{figure}

The policies operate at natural temporal hierarchies, creating stable multi-scale coordination, as shown in \cref{fig:temporal_scales}. This temporal separation enables a \textbf{nested quasi-static approximation}: the fastest dynamics (reconfiguration) equilibrate within each frame update, and frame dynamics settle before the slower sensor policy evolves.
This hierarchy prevents destabilizing interactions across timescales 
while preserving the necessary coupling for coherent, coordinated behavior.

\section{Hyperelastic Ring Robot Model}
\label{app:hyperelastic_model}

Our framework in Sec.~\ref{sec:method} is agnostic to the underlying robot body: the score-based policies $\{\pi_y,\pi_f,\pi_o\}$ only assume access to a local context $\mathcal{C}_t$ , a reduced state $z_t$ and a modular hamiltonian proposal. For experiments, we instantiate this framework on a planar \emph{hyperelastic ring} that must squeeze through clutter and narrow passages. In this domain, the shape policy $\pi_o$ acts on a low-dimensional shape state
\[
z_o(t) = \big(s(t), p_s(t)\big),
\]
where $s(t)\!>\!0$ is a uniform scale, and one denote $\mathbf{o}(t)\in\mathbb{R}^2$ as the ring center coincide with robot frame center $\vec{c}$.

\paragraph{Design rationale.}
We choose a hyperelastic ring with a spline boundary for three reasons that match the Hamiltonian formulation in Sec.~\ref{sec:method}:
(i) it is the simplest soft body that \emph{provably requires} shape change to solve our benchmarks (a rigid disc of any fixed radius cannot pass through the same bottlenecks), so deformation and collision–clearance trade-offs cannot be “faked” by rigid planning alone;
(ii) the reduced state $s$ provides a low-dimensional, analytically tractable testbed for the energy decomposition in \cref{eq:module_template_hamiltonian}, letting us study Hamiltonian energy shaping and meta-weights $(\beta,\lambda,\{\alpha_i\})$ without confounding internal DOF; and

Thus, the hyperring model is not a restriction of GRL--SNAM, but a controlled, low-DOF instantiation that stresses exactly the squeezing and clearance behaviors we target.

\begin{figure}[htbp]
\centering
\begin{tikzpicture}[scale=0.8]
\draw[thick,blue] (0,0) ellipse (1.5 and 1.2);
\fill[blue] (0,0) circle (2pt);
\node[below] at (0,-0.3) {$\mathbf{o}(t)$};
\draw[<->,red] (-1.5,0) -- (1.5,0);
\node[above,red] at (0,0.3) {$s(t)$};

\foreach \angle in {0,45,90,135,180,225,270,315} {
  \fill[orange] ({1.5*cos(\angle)},{1.2*sin(\angle)}) circle (1.5pt);
}
\node[orange] at (2,0.5) {$\mathbf{X}_j$};
\draw[->,purple,thick] (1.5,0) -- (2.2,0.3);
\draw[->,purple,thick] (0,1.2) -- (0.3,1.8);
\node[purple] at (2.5,0.8) {$\mathbf{F}_{\text{ext}}$};
\fill[gray] (2.5,-1) circle (0.3);
\fill[gray] (-2,-0.5) circle (0.4);
\node[gray] at (2.5,-1.5) {obstacles};
\draw[->] (-3,-2) -- (-2,-2) node[right] {$x$};
\draw[->] (-3,-2) -- (-3,-1) node[above] {$y$};
\end{tikzpicture}
\caption{Hyperelastic ring instantiation of the generic shape policy $\pi_o$.
The reduced configuration $s$ is mapped to spline samples $\mathbf{X}_j$ where policy-induced forces act.}
\label{fig:hyperelastic_ring}
\end{figure}
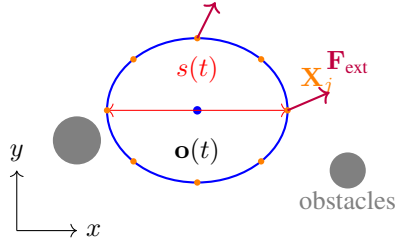

\paragraph{Geometric Representation of Hyperelastic Ring}

We represent the boundary of the hyperelastic ring object by a periodic cubic B-spline with $n_{\text{ctrl}}$ control points
\begin{equation}
\mathbf{S}(u) = \sum_{i=1}^{n_{\text{ctrl}}} N_{i,3}(u)\,\mathbf{P}_i,
\quad u \in [0,1],
\end{equation}
where $N_{i,3}$ are $C^2$-continuous B-spline basis functions.
This choice provides a smooth closed curve with well-defined normals and curvature, and allows fast evaluation through precomputed matrices. A reference unit ring is initialized as
\begin{equation}
\mathbf{P}_{0,i} 
= r_{\text{base}}
\begin{bmatrix}
\cos(2\pi i/n_{\text{ctrl}}) \\
\sin(2\pi i/n_{\text{ctrl}})
\end{bmatrix},
\end{equation}
and transformed to world coordinates by a similarity transform
\(\mathbf{P}_i(t) = \mathbf{o}(t) + s(t)\,\mathbf{P}_{0,i}.\)
For numerical evaluation we precompute spline matrices $B,D\in\mathbb{R}^{K\times n_{\text{ctrl}}}$ and sample $K$ points on the boundary
\begin{equation}
\mathbf{X}_j(t) = \sum_{i} B_{ji}\,\mathbf{P}_i(t), 
\quad
\mathbf{X}'_j(t) = \sum_{i} D_{ji}\,\mathbf{P}_i(t),
\end{equation}
with arc lengths $\ell_j = \|\mathbf{X}'_j\|$ and unit tangents $\hat{\mathbf{T}}_j = \mathbf{X}'_j / \ell_j$.
We use uniform quadrature weights $\omega_j= 1/K$.

\paragraph{{Shape Sub-Modular Hamiltonian.}}

In this domain, the generic Hamiltonian from \cref{eq:module_template_hamiltonian} specializes to
\begin{equation}
\mathcal{H}_{\text{o}}(z_o;\mathcal{E}_\tau)
=\frac{1}{2}M_s^{-1} p_s^2 
+ \mathcal{R}_o(q_o;\mathcal{E}_t),
\end{equation}
where $M_s$ refers a fixed ``mass" term of scale and $p_s=M_s^{-1}\dot{s}$ refers conjugate momenta.
We match the navigation specialization in \cref{eq:module_template_potential:o} by defining the barrier potentials $b(d_i(q,\mathcal{E}))$ and object intrinsic energy $E_{\mathrm{obj}}$ for deformation.

\paragraph{{Barrier potentials.}}
The sensor policy $\pi_y$ exposes { weighted point obstacles} $(\mathbf{c}_i,r_i,w_i)$ in $\mathcal{C}_t$ from $\mathcal{E}$. Here $\vec{c}_i$ denotes the center of objects and $r_i$ denotes the distance. $w_i$ denotes the scale of obstacles. For each sample point $\mathbf{X}_j$ we compute distances \(d_{ji} = \|\mathbf{X}_j - \mathbf{c}_i\| - r_i,\)
and accumulate an Incremental Potential Contact (IPC, \citet{li2020incremental}) barrier function:
\begin{equation}
d_i(q,\mathcal{E}):=
\sum_{j=1}^K \omega_j \ell_j  w_k b_{\text{IPC}}(d_{jk};\hat{d}),
\end{equation}
with $b_{\text{IPC}}(\cdot)$ in the standard piecewise form
\begin{equation}
b_{\text{IPC}}(d;\hat{d}) =
\begin{cases}
-(d-\hat{d})^2(\log d - \log \hat{d}) & 0 < d < \hat{d},\\[2pt]
0 & d \geq \hat{d},\\[2pt]
V_{\text{penalty}} & d \leq 0.
\end{cases}
\end{equation}
 
The term $\hat{d}$ refers the contact distance thresholding while $V_{\text{penalty}}$ is a user-specified constant that penalize penetration. This is a concrete instantiation of the barrier term in \cref{eq:barrier_shaped_cost}, with IPC chosen because its differential yields near obstacle contact forces and time integrations resembles physical-world behaviors of an elastic object. { We provide \cref{cor:ipc_from_logbarrier} to justify IPC potentials are also a valid barrier function that maintains the conclusions addressed in \cref{prop:barrier_convergence}.}
\paragraph{Object potential.}

Let $A_{\text{ref}}$ denote the reference rest area of the ring. We denote the scaled area $A(s)= s^2 A_{\text{ref}}$.
We compute the minimal clearance \(d_{\min}:=\min_{j,i}~\{d_{ji}\},\)
and use it to define a clearance-dependent scale target 
\begin{equation}
s_{\text{target}}
:=
s_0 + (1-s_0)\tanh\big(\delta\,\max(d_{\min},0)\big),
\label{eq:def:radius:target}
\end{equation}
which encourages compression in tight passages while keeping the ring expanded in free space:
for large $d_{\min}$, $s_{\text{target}}$ shall be close to $1$, while for lower clearance the target shrinks smoothly.
The bulk potential is thus defined as
\begin{equation}
{E}_{\text{obj}}(q):=\mathcal{U}_{bulk}(s;d_{\min})
=
\frac{k_{\text{bulk}}}{2}\big(A(s) - A(s_{\text{target}})\big)^2.
\end{equation}
This realizes $E_{\mathrm{obj}}$ with a \emph{single} deformation DOF (scale $s$), striking a balance between expressivity (squeezing is possible) and the low-dimensional Hamiltonian search space assumed in Sec.~\ref{subsec:problem_formulation}. The three policies $\{\pi_y,\pi_f,\pi_o\}$ interact with the ring exactly at the level of boundary samples, in a way that mirrors the abstract modular decomposition of $\mathcal{R}$.

\section{Experimental Evaluation}
\label{sec:experiments}

We evaluate GRL-SNAM on two navigation benchmarks that share the same \emph{stagewise, local-sensing} interface but differ in dynamics: (i) deformable hyperelastic ring navigation in cluttered workspaces, and (ii) point-agent dungeon navigation derived from~\citet{liang2023context}. In both tasks, GRL-SNAM receives the state $q_t$, a stage-exit goal $g_t$, and obstacles within a local window; task-specific dynamics and energy terms instantiate the reduced Hamiltonian $\mathcal{H}$.

\subsection{Tasks and Environments}
\label{subsec:tasks}

\paragraph{Stagewise local sensor and mapping effort quantification.}
At time $t$, the agent observes obstacles intersecting a local window $\mathcal{W}(q_t)$ of size $2\hat d \times 2\hat d$ centered at $q_t$, and a stage signal (macro-cell index) with its current exit goal $g_t$ from a stage manager over a workspace tiling. We quantify sensing effort by the mapping ratio
\begin{equation}
\rho_{\mathrm{map}} \;:=\; \frac{\mathrm{area}\!\left(\bigcup_t \mathcal{W}(q_t)\right)}{L^2}.
\label{eq:mapping_ratio}
\end{equation}

\subsubsection{ Narrow-Gap Passing Navigation Task with a Hyperelastic Ring Robot}
\label{subsubsec:ring_task}

\paragraph{Environment.}
Each instance is a $[0,L]\times[0,L]$ workspace with procedurally sampled circular obstacles (positions/radii/density), spanning bottlenecks, zig-zag corridors, and cul-de-sacs. Start--goal pairs are sampled so that a deformable ring admits a collision-free solution, while a rigid disc of comparable size may not. We construct Train, {In distribution Test(\textbf{Test-ID}, matched statistics), and Out-of-distribution Test (\textbf{Test-OOD}, shifted gap/density) splits}.

\paragraph{Dynamics and interface.}
The robot is a reduced-order hyperelastic ring with Hamiltonian $\mathcal{H}(q,p)$. GRL-SNAM interacts only through $(q_t, g_t, \mathcal{W}(q_t))$ and the resulting generalized forces/coefficients; implementation details of the hyperelastic model are provided in Section.~\ref{app:hyperelastic_model}.

\paragraph{Reduced Hamiltonian used in this task.}
We use a reduced Hamiltonian whose potential combines a goal term toward $g_t$ and barrier terms from signed distances to locally observed obstacles, with dissipation/regularization. Context encoders predict coefficients $(\beta,\gamma,\alpha)$ that modulate these terms.

\subsubsection{ Dungeon Navigation via a Point-Nav system }
\label{subsubsec:dungeon_task}

\paragraph{Environment and dynamics.}
We use dungeon layouts from~\citet{liang2023context}. The agent is a point mass in continuous 2D with state $q_t$ (position; optionally velocity) and action $\mathbf{a}_t=[v_x,v_y]\in[-3,3]^2$. Collisions with walls are penalized/terminating; reaching the goal yields reward.

\paragraph{Observations (matched to the stagewise interface).}
A \texttt{StagewiseSensor} selects the active macro-cell and exposes its exit as the local goal $g_t$. An \texttt{ObstacleExtractor} fits circular obstacles to wall segments inside $\mathcal{W}(q_t)$, returning centers, radii, and a visibility mask. The observation concatenates stage-relative position, $g_t$, and the local obstacle set.

\paragraph{Use in evaluation.}
We instantiate GRL-SNAM with a simple quadratic goal energy and radial barrier energies from the extracted obstacles, and compare to PPO/TRPO/SAC under (i) full-episode long-horizon training and (ii) short-rollout stagewise training with the same observation/action interface as GRL-SNAM.

\subsubsection{Experimental Setups}

We justify our choice of predefined hyperparemeters for each task in \cref{tab:hyperparams} and further explain them in \ref{subsec:hyperparam}. Most parameters are learnable or sharable across tasks. Only when maps' scale and robot configurations vary shall one choose additional hyperparameters. Experiments are performed on one NVIDIA A100 GPU with 40 GB VRAM, using Linux operating system, and all networks are trained via PyTorch \citep{paszke2019pytorch} library. For RL baselines, we adapt PPO, SAC and TRPO algorithms provided in Stable-Baselines3 \citep{stable-baselines3} for comparison.

\begin{table}[t]
\centering
\caption{Hyperparameter nomenclature and selection across experiments. \textbf{Default: fixed.}
Parameters marked \learn are learned/predicted by the meta-policy (softplus-enforced nonnegativity where applicable); parameters marked \pre are precomputed.}
\label{tab:hyperparams}
\setlength{\tabcolsep}{5pt}
\renewcommand{\arraystretch}{1.12}
\small
\begin{tabularx}{\linewidth}{@{}l l X l@{}}
\toprule
\textbf{Block} & \textbf{Symbol} & \textbf{Description} & \textbf{Value / Range} \\
\midrule

\rowcolor{gray!15}\multicolumn{4}{@{}l@{}}{\textit{Learned coefficients (variable-dimension $\zeta_t$)}} \\
Barrier & $\bm{\alpha}_j$\learn & Per-obstacle barrier weight & $\alpha_j \ge 0$ (softplus) \\
Goal & $\bm{\beta}$\learn & Goal attraction strength & $\beta \ge 0$ (softplus) \\
Damping & $\bm{\gamma}$\learn & Linear damping scale & $\gamma \ge 0$ (softplus) \\
\midrule

\rowcolor{gray!15}\multicolumn{4}{@{}l@{}}{\textit{Shared pipeline: loss \& optimization}} \\
Opt. & $\eta$ & Learning rate & $1\times 10^{-4}$--$3\times 10^{-4}$ \\
Loss & $w_{\text{traj}},\, w_{\text{vel}}$ & Position / velocity loss weights & $1.0,\;1.0$ \\
Loss & $w_{\text{fric}},\, w_{\text{multi}}$ & Damping / multi-start penalty & $0.1,\;0.5$ \\
Rollout & $H$ & Surrogate rollout horizon & $2$--$6$ steps \\
Test-time & $\lambda_{\text{prox}}$ & Proximal regularization & $10^{-3}$ \\
\midrule

\rowcolor{gray!15}\multicolumn{4}{@{}l@{}}{\textit{Shared pipeline: IPC barrier \& staging}} \\
Barrier & $\hat d$ & Barrier activation distance & $0.28$--$1.0$ \\
Barrier & $b(d)$ & $-(d-\hat d)^2\log(d/\hat d)$ & --- \\
Staging & $(W,H)$ & Stage window dimensions & $(2.6,\,2.0)$ \\
Staging & $\rho_{\text{overlap}}$ & Stage overlap ratio & $0.3$ \\
Geometry & $r_{\text{inflate}}$ & Obstacle inflation (tube radius) & $0.05$--$0.4$ \\
\midrule

\rowcolor{blue!8}\multicolumn{4}{@{}l@{}}{\textit{Hyperelastic ring: dynamics}} \\
Trans. & $M_{\vec{c}},\,\gamma_{\vec{c}}$ & Robot pose mass / damping & $1.5,\;4.0$ \\

Scale & $M_s,\,\gamma_s$ & Scale mass / damping & $1.0,\;2.0$ \\
Material & $k_{\text{bulk}}$ & Bulk modulus (area stiffness) & $1.5$ \\
\midrule

\rowcolor{blue!8}\multicolumn{4}{@{}l@{}}{\textit{Hyperelastic ring: geometry}} \\
Shape & $r_{\text{robot}}$ & Nominal ring radius & $0.30$--$0.50$ \\
Spline & $n_{\text{ctrl}},\,K$ & B-spline control / sample points & $20,\;240$ \\
Time & $\Delta t$ & Integration timestep & $0.03$ \\
\midrule

\rowcolor{green!8}\multicolumn{4}{@{}l@{}}{\textit{Dungeon: FPE navigation}} \\
Speed & $v_{\max}$ & Maximum velocity & $2.5$--$3.0$ \\
Goal & $k_{\text{goal}}$ & Goal attraction gain & $20.0$--$25.0$ \\
Barrier & $k_{\text{barrier}}$ & Barrier repulsion gain & $0.0$ \\
Time & $\Delta t$ & Integration timestep & $0.3$ \\
\midrule

\rowcolor{green!8}\multicolumn{4}{@{}l@{}}{\textit{Dungeon: environment}} \\
Map & $\mathrm{SDF}(x,y)$\pre & Signed distance field (from grid) & --- \\
Staging & $s_{\text{stage}}$ & Stage window size & $60$ px \\
Waypoints & $d_{\text{wp}}$ & Minimum waypoint wall clearance & $1.5$--$4.0$ px \\
\bottomrule
\end{tabularx}
\end{table}

\subsection{Baseline Methods}
\label{subsec:baselines}

We compare GRL-SNAM against a spectrum of planning and learning methods that differ in (i) map access (full vs.\ local), (ii) planning scope (global vs.\ local), and (iii) supervision signal (hand-crafted surrogates vs.\ learned policies). Below we summarize the baselines and their roles.

\subsubsection{Baselines for Hyperelastic Ring Navigation}
\label{subsubsec:baselines_ring}

For the deformable hyperring task we consider two non-learning families plus our GRL-SNAM controller, all instantiated on the same hyperelastic model.

\paragraph{Global planners (full-map references).}
Assuming full occupancy grids, these methods provide reference paths and define $L_{\mathrm{ref}}$ for SPL and detour:
\begin{itemize}
    \item \textbf{Rigid A$^{\star}$:} A$^{\star}$ on an obstacle-inflated grid for a rigid disc of radius $r_{\text{rest}}$, yielding conservative collision-free paths that may be overly cautious for a deformable ring.
    \item \textbf{Deformable A$^{\star}$:} Clearance-aware A$^{\star}$ that augments edge costs with a deformation penalty as clearance approaches a minimum radius $r_{\min}$, allowing squeezing through tight passages at an energetic cost.
\end{itemize}

\paragraph{Local reactive planners (stagewise, same information as GRL-SNAM).}
To provide fair local baselines, we instantiate classical controllers that share the \emph{same} stage manager and local window as GRL-SNAM:
\begin{itemize}
    \item \textbf{Potential Fields (PF, staged):} Attractive potential toward the current stage exit plus repulsive potentials from locally sensed obstacles and stage boundaries.
    \item \textbf{Control Barrier Functions (CBF, staged):} A nominal control toward the stage exit filtered through a CBF quadratic program using clearance-based barrier functions to enforce local safety.
    \item \textbf{Dynamic Window Approach (DWA, staged):} Sampling of $(v,\omega)$ pairs within dynamics limits, short-horizon rolling, and scoring based on goal alignment, velocity, and clearance with penalties for leaving the active stage.
\end{itemize}
All three operate under identical sensing and stagewise constraints as GRL-SNAM, but rely on hand-crafted surrogates instead of learned Hamiltonians.

\paragraph{Ours: GRL-SNAM (hyperring instantiation).}
On the hyperelastic ring, GRL-SNAM uses the reduced Hamiltonian described in Section~\ref{subsec:navigator_meta_learner} and ~\ref{app:hyperelastic_model}, combining goal, barrier, and dissipation terms with adaptive coefficients $(\beta,\gamma,\alpha)$ predicted by context encoders. It shares the same stage manager and local window as the reactive baselines but learns the local force field from short rollouts instead of fixing it a priori. All hyperring results (minimal sensing vs.\ navigation quality, planner comparisons, Hamiltonian field analysis, ablations, robustness) are reported against this set.

\subsubsection{Baselines for Dungeon Point-Agent Navigation}
\label{subsubsec:baselines_dungeon}

\begin{figure}[h!]
    \centering
    \includegraphics[width=\linewidth]{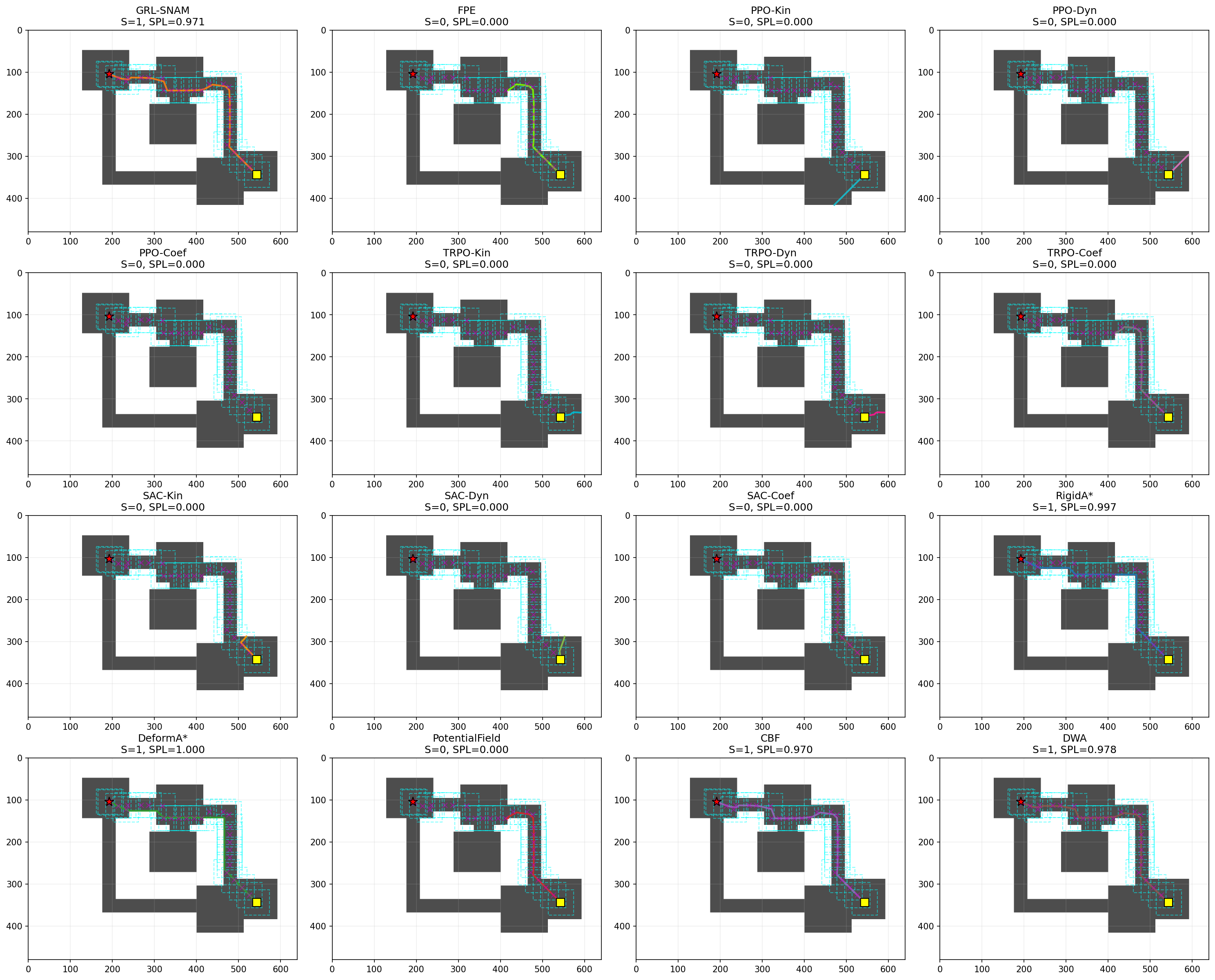}
    \caption{\textbf{Representative dungeon rollouts under stagewise local sensing.} Each panel shows the same layout with the goal (red $\times$), start (yellow square), executed trajectory (colored by time), and the union of visited sensing windows $\mathcal{W}(q_t)$ (cyan boxes). Panel titles report episode success $S$ and SPL (with Grid A$^{\star}$ as the reference path length $L_{\mathrm{ref}}$). GRL-SNAM reaches the goal with near-planner efficiency, whereas stagewise deep-RL baselines (PPO/TRPO/SAC with Kin/Dyn/Coef parameterizations) and the greedy potential-field baseline fail on this instance; classical planners and safety-filtered local methods (CBF/DWA) succeed but follow more conservative routes.}

    \label{fig:dungeon_map}
\end{figure}

All dungeon methods share the same point-agent dynamics and action space, so differences reflect the planning/learning rule and map access (Fig.~\ref{fig:dungeon_map}).

\paragraph{Classical baselines.}
\begin{itemize}
    \item \textbf{Grid A$^{\star}$ (full map):} 8-connected A$^{\star}$ on the true occupancy grid; used as the path-length reference $L_{\mathrm{ref}}$.
    \item \textbf{Greedy PF (local):} moves toward the goal with repulsion from walls in the local occupancy patch.
    \item \textbf{Local DWA (local):} samples velocity commands $(v_x,v_y)$, rolls out short trajectories against locally sensed walls, and scores by goal alignment, clearance, and length.
\end{itemize}

\paragraph{Deep RL baselines.}
We train PPO~\citep{schulman2017ppo}, TRPO~\citep{schulman2015trust}, and SAC~\citep{haarnoja2018sac} with the same encoder and observation encoding under two regimes:
\begin{itemize}
    \item \textbf{Full-episode RL (long horizon):} standard end-to-end episodes with a global goal and a fixed step budget (e.g., 2000), without stage supervision.
    \item \textbf{Stagewise RL (short rollouts):} the same stagewise observation/action interface as GRL-SNAM: stage-relative state, stage-exit goal, and local extracted obstacles; episodes are extended to $K_{\mathrm{ext}}=8K$ to support exploration.
\end{itemize}

\paragraph{Within-framework variants (same data, different learning rule).}
Holding the stagewise interface and short-rollout dataset fixed, we replace Hamiltonian-structured supervision with:
\begin{itemize}
    \item \textbf{Velocity regression:} supervised prediction of $\mathbf{a}_t$ from observations.
    \item \textbf{Coefficient RL:} policy-gradient learning over Hamiltonian coefficients $(\alpha,\beta,\gamma)$ instead of the DfPO-style objective.
\end{itemize}

\subsection{Metrics and Evaluation Protocol}
\label{subsec:metrics_protocol}

Across both tasks we evaluate methods along four axes:
(i) task success and path efficiency,
(ii) safety and physical plausibility,
(iii) mapping / sensing budget, and
(iv) learning efficiency and robustness.
Unless stated otherwise, we report averages over multiple random seeds and environment instances.

\paragraph{Navigation performance.}
We report:
(i) \textbf{success rate} (fraction of episodes reaching the goal without hard collisions),
(ii) \textbf{Success-weighted Path Length (SPL)}, measuring path efficiency relative to a reference planner (Rigid A$^{\star}$ for hyperring, Grid A$^{\star}$ for dungeon),
and (iii) \textbf{detour ratio} $L / L_{\mathrm{ref}}$ over successful runs, indicating excess path length relative to the reference. We also track \textbf{smoothness} via average heading change per step to assess physical plausibility.

\paragraph{Safety and clearance.}
Safety is quantified by the \textbf{minimum} and \textbf{mean clearance} to the nearest obstacle along a trajectory (positive values indicate safety margins, negative values indicate penetrations), the number of \textbf{hard collisions} per episode, and, for methods with explicit barriers (CBF, GRL-SNAM, Deformable A$^{\star}$), the fraction of timesteps that violate a prescribed safety threshold (barrier violations).

\paragraph{Mapping and sensing budget.}
To measure how much of the environment each method must observe, we define the \textbf{mapping ratio} as the fraction of workspace area that ever falls inside a local sensing window during an episode. All stagewise methods (PF, CBF, DWA, GRL-SNAM, dungeon local baselines) use the same window size $2\hat d \times 2\hat d$; global planners (Rigid A$^{\star}$, Grid A$^{\star}$) have mapping ratio $100\%$.

\paragraph{Learning efficiency.}
For learning-based methods (GRL-SNAM, deep RL, within-framework variants) we track \textbf{sample efficiency} via success and SPL as functions of training interaction steps, summarize performance with area under the learning curve, and report the steps required to reach target thresholds (e.g., $80\%$ success or SPL $\geq 0.7$). For GRL-SNAM, “steps’’ are short-rollout samples used in supervised Hamiltonian training; for PPO/TRPO/SAC they are environment transitions. We also report total environment steps, gradient updates, and update-to-sample ratios where relevant.

\paragraph{Evaluation protocols.}
For hyperelastic ring experiments, each environment is evaluated from multiple random start–goal pairs (typically $50$), and we report aggregated statistics on Test-ID and Test-OOD splits. Unless noted, navigation-quality tables (e.g., Table~\ref{tab:success_only}) are computed on \emph{successful runs only} to isolate path quality from outright failure. For dungeon experiments, full-episode baselines are trained and evaluated on long-horizon episodes with fixed step budgets, while short-rollout baselines and GRL-SNAM are trained on a stagewise dataset with horizon $H \in [2,6]$ and evaluated on a held-out set of stagewise test samples with identical observation and action spaces. Robustness experiments vary sensing and dynamics fidelity over a grid of perturbation levels (e.g., position jitter, radius errors, missed/false obstacles, damping perturbations) and report success, SPL, clearance, and collisions; ablations vary only the training objective while keeping environments fixed, reporting the same core metrics plus qualitative behavior summaries.


\subsection{Results at a Glance}
\label{subsec:results_overview}

Across the hyperelastic ring and dungeon point-agent tasks, results support four takeaways:

\begin{enumerate}
    \item { Effective Path Planning under minimal sensing}
    On the hyperelastic ring, GRL-SNAM attains CBF-level efficiency while observing only sparse stagewise windows: Table~\ref{tab:success_only} shows SPL $\approx 0.95$ and detour $\approx 1.09$ at a mapping ratio of $\sim$10--11\%, comparable to PF's sensing budget but with substantially higher path efficiency.

    \item \textbf{Stronger navigation than {conventional, reactive, and deep RL baselines.}}
    On ring Test-ID/OOD, GRL-SNAM achieves near-perfect success with high SPL, smooth trajectories, and positive clearance, while A$^{\star}$ planners, PF/CBF/DWA, and RL baselines trade off safety vs.\ efficiency or fail in clutter (Fig.~\ref{fig:exp1_paths}, Fig.~\ref{fig:exp1_comparison}). In the dungeon benchmark, GRL-SNAM substantially outperforms PPO/TRPO/SAC trained end-to-end over millions of steps, and also outperforms their stagewise short-rollout versions under the same observation/action interface (Table~\ref{tab:full_episode_rl}, Table~\ref{tab:short_rollout_rl}).

    \item \textbf{The gain comes from the { adaptive reward and geometric Hamiltonian learning.}}
    Under matched stagewise sensing and short-rollout data, PPO/TRPO/SAC plateau below $30\%$ success, whereas GRL-SNAM reaches $\sim$90\% success with $\sim$500k supervised updates on the same interface (Table~\ref{tab:short_rollout_rl}), indicating the advantage is driven by Hamiltonian-structured supervision rather than algorithm choice.

    \item \textbf{Interpretable fields, learned constraints, and robust adaptation.}
    GRL-SNAM learns a coherent force field that balances goal attraction and barrier repulsion via adaptive coefficients $(\beta,\gamma,\alpha)$ (Fig.~\ref{fig:vector_fields}, Fig.~\ref{fig:time_series}) and recovers barrier-like energy profiles from interaction (Fig.~\ref{fig:barrier_profile}). Ablations isolate the role of dissipation and near-contact training (Table~\ref{tab:ablation_fric_multi}), and robustness tests show graceful degradation under sensing/dynamics noise (Table~\ref{tab:robustness}). Replay analysis indicates the system stores compact local transitions and constraint summaries rather than full maps (Fig.~\ref{fig:replay_buffer}).
\end{enumerate}

We next quantify sensing--performance trade-offs, then present the main ring and dungeon comparisons, followed by field analyses, ablations, and robustness.

\subsection{Navigation Quality vs.\ Mapping Effort under Minimal Sensing}
\label{subsec:minimal_sensing}

We test whether GRL-SNAM can achieve near-planner navigation while sensing only a small fraction of the workspace. {The dataset contains all the noise levels mentioned in \cref{tab:robustness}}
All \emph{stagewise} methods share the same stage manager and local sensing window of size $2\hat d \times 2\hat d$; they differ only in how local observations are converted into forces/actions. We compare (i) classical local controllers (PF, CBF), (ii) GRL-SNAM, and (iii) deep RL trained under the \emph{same} stagewise sensing and short-rollout distribution as GRL-SNAM.

\paragraph{RL under matched interfaces.}
For PPO, TRPO, and SAC we evaluate three control parameterizations:
\textbf{Kin} (output velocities), \textbf{Dyn} (output forces integrated by a damped point-mass model), and \textbf{Coef} (output Hamiltonian coefficients $(\alpha,\beta,\gamma)$). 
Thus, RL baselines share observations, local dynamics, and Hamiltonian building blocks with GRL-SNAM; only the learning rule and control parameterization change.

\paragraph{Metrics.}
On \emph{successful runs} we report SPL, detour $L/L_{\mathrm{ref}}$ (Rigid A$^{\star}$ reference), minimum clearance (negative indicates penetration), and mapping ratio (fraction of workspace covered by any sensing window).

\begin{table}[h!]
\centering
\caption{Navigation quality vs.\ mapping effort on the hyperelastic ring task (success-only runs). 
All stagewise methods use the same sensing window and stage manager. RL variants share the same stagewise sensing and short-rollout distribution as GRL-SNAM, differing only in control parameterization (Kin/Dyn/Coef) and learning rule.}
\vspace{0.4em}
\begin{tabular}{lcccc}
\toprule
Method       & SPL $\uparrow$ & Detour $\downarrow$ & MinClear (m) $\uparrow$ & Mapping (\%) $\downarrow$ \\
\midrule
PF           & 0.77     & 1.42 & 0.18           & 10.3 \\
CBF          & \textbf{0.96}     & \textbf{1.04} & \textbf{0.32}  & 11.2 \\
GRL-SNAM     & \textit{0.95} & \textit{1.09} & \textit{0.26} & \textbf{10.7} \\
\midrule
PPO-Kin      & 0.143    & 1.87 & $-0.358$       & 15.4 \\
PPO-Dyn      & 0.000    & 2.31 & 0.771          & 18.1 \\
PPO-Coef     & 0.071    & 1.65 & $-0.092$       & 14.7 \\
TRPO-Kin     & 0.000    & 2.12 & $-0.327$       & 16.5 \\
TRPO-Dyn     & 0.000    & 1.98 & 0.019          & 15.9 \\
TRPO-Coef    & 0.571    & 1.44 & 0.004          & 15.3 \\
SAC-Kin      & 0.000    & 2.27 & $-0.229$       & 16.8 \\
SAC-Dyn      & 0.000    & 2.41 & $-0.229$       & 16.9 \\
SAC-Coef     & 0.571    & 1.53 & 0.004          & 14.6 \\
\bottomrule
\end{tabular}
\label{tab:success_only}
\end{table}

\paragraph{Near-planner quality at minimal mapping.}
Table~\ref{tab:success_only} shows that GRL-SNAM matches CBF-level efficiency (SPL $0.95$ vs.\ $0.96$; detour $1.09$ vs.\ $1.04$) while requiring essentially the same mapping ratio as PF ($10.7\%$ vs.\ $10.3\%$). 
PF, despite identical sensing, yields substantially longer paths (detour $1.42$) and lower SPL ($0.77$), indicating that the improvement comes from how sensed patches are converted into a local navigation field, not from sensing more of the map.

\paragraph{RL does not close the gap under matched sensing.}
Deep RL variants consume \emph{more} mapping (typically $14$--$18\%$) yet remain far behind GRL-SNAM. We observe three recurring behaviors:
(i) \textbf{Kin} policies often graze or collide (negative MinClear) and achieve low SPL; 
(ii) \textbf{Dyn} policies frequently stall or fail to complete (near-zero SPL, large detours); and 
(iii) \textbf{Coef} policies are strongest (TRPO/SAC-Coef), but still require higher mapping and operate at near-grazing clearances ($\approx 0.004$m) with only moderate SPL ($0.57$). 
Changing optimizer (PPO/TRPO/SAC) or parameterization (Kin/Dyn/Coef) shifts outcomes but does not reproduce the efficiency--safety trade-off achieved by GRL-SNAM.

\paragraph{Takeaway.}
Under identical stagewise sensing, GRL-SNAM is the only method that simultaneously attains high path efficiency (near CBF), low mapping (near PF), and strictly positive clearance, suggesting the main advantage is the Hamiltonian-structured supervision that learns a constraint-aware local energy landscape rather than a direct reward-maximizing policy.

\subsection{Comprehensive Navigation Comparison}
\label{subsec:comprehensive_nav}

\begin{figure}[h!]
    \centering
    \includegraphics[width=\linewidth]{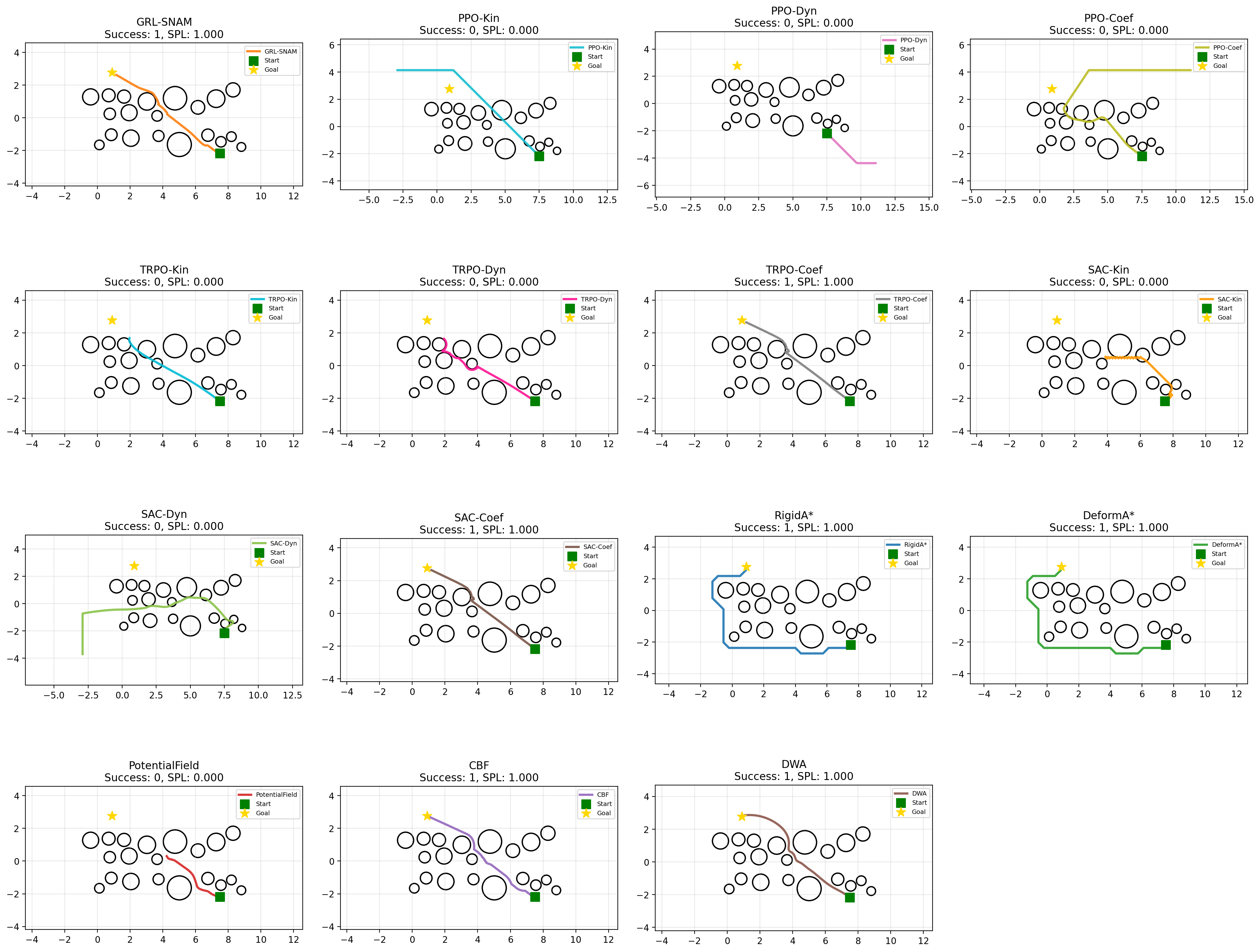}
    \caption{{\textbf{Qualitative rollouts on a representative Test-OOD bottleneck layout.} Start is marked by $\star$, goal by $\square$, and circles denote obstacles; each panel title reports Success and SPL. \textbf{Row 1:} GRL-SNAM and PPO (Kin/Dyn/Coef). \textbf{Row 2:} TRPO (Kin/Dyn/Coef) and SAC-Kin. \textbf{Row 3:} SAC (Dyn/Coef) and global planners (RigidA*, DeformA*). \textbf{Row 4:} classical local methods (Potential Field, CBF, DWA). GRL-SNAM is the only method here that achieves a smooth, short \emph{squeeze-and-recover} trajectory through the bottleneck without grazing, while learning-based baselines either fail (Kin/Dyn) or succeed by boundary-hugging (Coef). Global planners succeed but are jagged/overly conservative, and local controllers highlight reactive artifacts (trapping for PF; oscillatory/arc-like adjustments for CBF/DWA).}}

    \label{fig:exp1_paths}
\end{figure}

\begin{figure}[h!]
    \centering
    \includegraphics[width=0.48\linewidth]{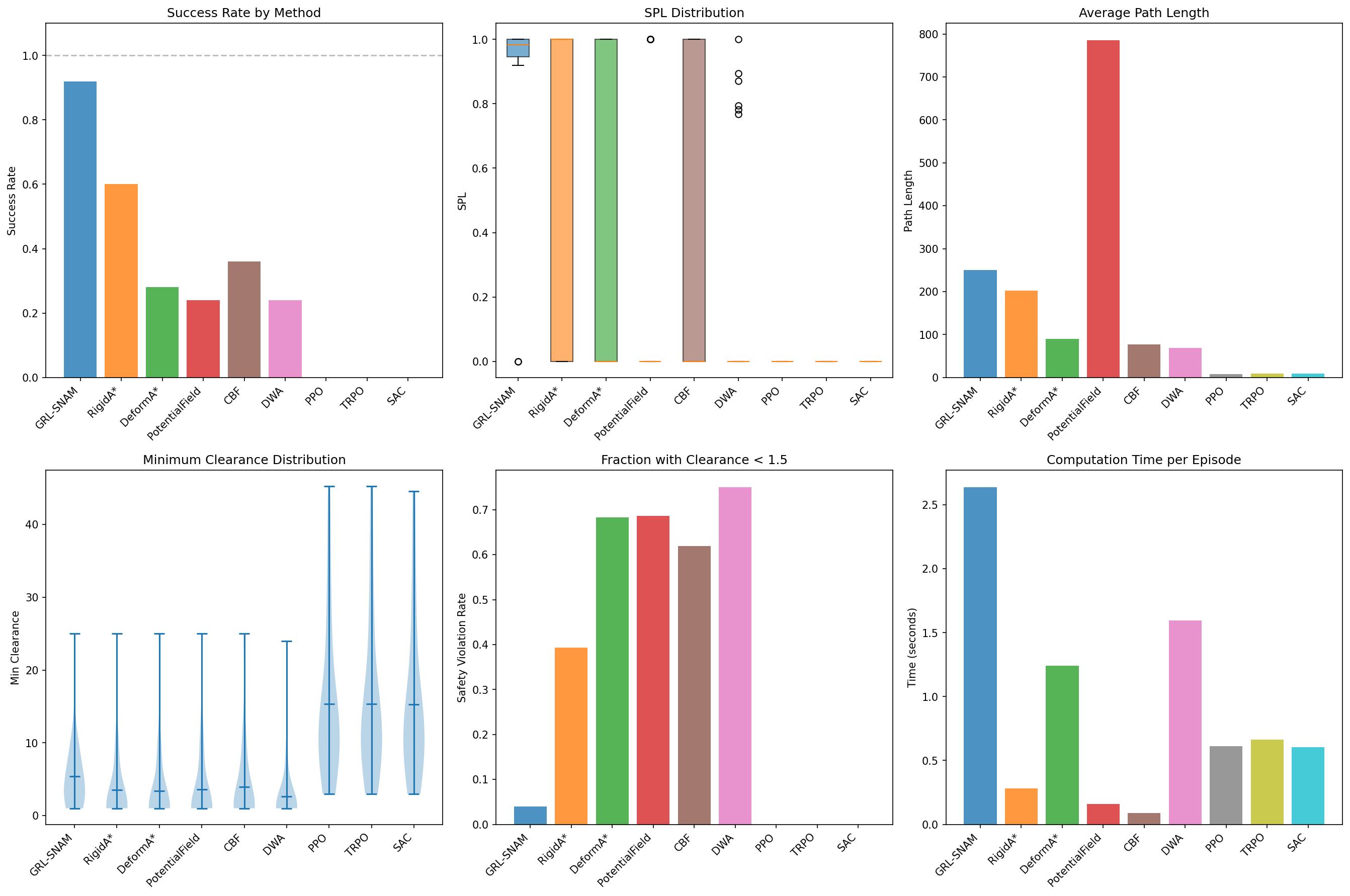}
    \hfill
    \includegraphics[width=0.48\linewidth]{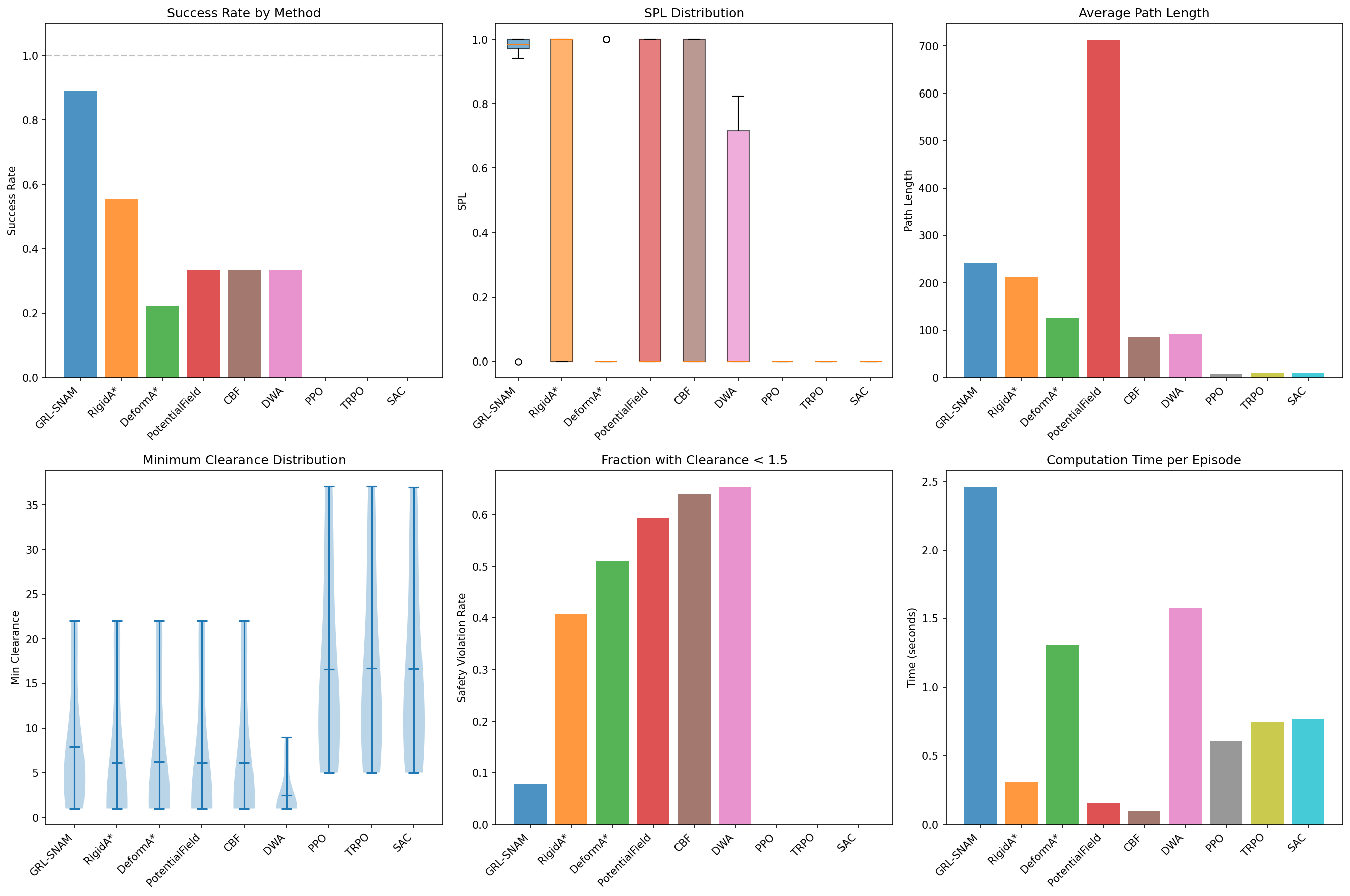}
    \caption{{\textbf{Test-ID (left) and Test-OOD (right) dashboard (aggregate over all trials).}
Each side contains six panels.
\emph{Top row (performance/efficiency):} (1) success rate; (2) SPL distribution (zeros correspond to failures; higher is better among successes); (3) average path length.
\emph{Bottom row (safety/cost):} (4) distribution of per-episode minimum clearance $\mathrm{clr}_{\min}=\min_n \mathrm{clr}_n$; (5) grazing rate $\mathbb{P}(\mathrm{clr}_{\min}\le d_{\text{thr}})$ with $d_{\text{thr}}{=}1.5$; (6) average computation time per episode.
GRL-SNAM achieves a strong joint trade-off across splits: high success with high SPL and short paths, while preserving positive clearance (low grazing rate) at moderate compute.}}

    \label{fig:exp1_comparison}
\end{figure}

We compare GRL-SNAM to (i) full-map global planners (Rigid A$^{\star}$, Deformable A$^{\star}$), 
(ii) stagewise local controllers (PF, CBF, DWA) with the same sensing window and stage manager, 
and (iii) deep RL policies trained under the same stagewise, short-rollout interface as GRL-SNAM (Kin/Dyn/Coef).
We evaluate on Test-ID and Test-OOD using success, SPL, detour $L/L_{\mathrm{ref}}$, minimum clearance, and smoothness.
Figure~\ref{fig:exp1_comparison} reports aggregate metrics and Figure~\ref{fig:exp1_paths} shows representative rollouts.

\paragraph{Performance across ID and OOD.}
GRL-SNAM maintains near-perfect success on both Test-ID and Test-OOD and achieves the highest SPL with low variance.
By contrast, full-map planners are either conservative (inflation-induced detours) or sensitive to hand-tuned deformation costs, and stagewise reactive methods (PF/CBF/DWA) degrade on Test-OOD due to local oscillations, stalls, or cul-de-sac behavior.

\paragraph{Aggregate trade-offs (readout of Fig.~\ref{fig:exp1_comparison}).}
{Fig.~\ref{fig:exp1_comparison} is organized as two dashboards: Test-ID (left) and Test-OOD (right).
Within each dashboard, the \emph{top row} reports task-level performance and efficiency:
(\textit{left}) success rate (fraction of episodes reaching the goal),
(\textit{middle}) the distribution of SPL (zeros indicate failures; values near one indicate near-shortest successful paths),
and (\textit{right}) average path length, which disambiguates whether high SPL comes from genuinely short paths versus a small number of efficient successes.
The \emph{bottom row} reports safety and cost:
(\textit{left}) the distribution of per-episode minimum clearance
$\mathrm{clr}_{\min}=\min_n \mathrm{clr}_n$ (lower tails indicate grazing/collisions),
(\textit{middle}) the “grazing rate” $\mathbb{P}(\mathrm{clr}_{\min}\le d_{\text{thr}})$ for a fixed threshold $d_{\text{thr}}$ (here $1.5$),
and (\textit{right}) average computation time per episode.

Across both Test-ID and Test-OOD, GRL-SNAM couples \emph{high success} with an SPL distribution concentrated near high values, while maintaining strictly positive clearance (small lower tail and low grazing rate).
In contrast, kinematic/dynamic RL baselines show a larger mass at low SPL (more failures and inefficient successes) and heavier low-clearance tails, while conservative/global planners and potential-field variants tend to keep large clearance but pay for it with reduced efficiency (lower SPL / longer paths) and, in some settings, additional failures from overly conservative detours or local minima.}

\paragraph{Qualitative behavior.}
{Fig.~\ref{fig:exp1_paths} shows a representative \emph{Test-OOD bottleneck} layout (start: $\star$, goal: $\square$; circles: obstacles). Each panel title reports \emph{Success} and \emph{SPL}.
\textbf{Row 1 (GRL-SNAM and PPO variants):}
GRL-SNAM (top-left) executes a single, smooth \emph{squeeze-and-recover} maneuver: it threads the narrow gap between the upper obstacle cluster and the large central obstacle, then recenters and reaches the goal with minimal detour (Success $=1$, SPL $=1$). In contrast, PPO-Kin and PPO-Dyn (top row, middle) fail in this bottleneck: the kinematic variant stalls short of the goal, while the dynamic variant drifts into an unproductive detour and times out (Success $=0$). PPO-Coef (top-right) illustrates that even when coefficients are learnable, the rollout can still fail on this OOD geometry (Success $=0$), reflecting sensitivity to tight clearances.

\textbf{Row 2 (TRPO variants and SAC-Kin):}
TRPO-Kin and TRPO-Dyn (second row, left/middle) again fail to traverse the bottleneck (Success $=0$), typically stopping early or clipping the corridor geometry. TRPO-Coef (second row, third) succeeds (Success $=1$) but the path \emph{hugs obstacle boundaries} through the constriction, indicating smaller clearance margins than GRL-SNAM. SAC-Kin (second row, right) also fails on this instance (Success $=0$).

\textbf{Row 3 (SAC variants and global planners):}
SAC-Dyn (third row, left) fails (Success $=0$) and exhibits a large, boxy detour pattern that does not resolve the bottleneck. SAC-Coef (third row, second) succeeds (Success $=1$) but, similar to TRPO-Coef, succeeds by skimming close to obstacles in the narrow passage. The global planners RigidA* and DeformA* (third row, right two panels) both succeed (Success $=1$) but produce \emph{piecewise-linear and/or conservative} routes with multiple right-angle turns and larger stand-off distances, yielding longer paths that avoid exploiting the short near-boundary squeeze.

\textbf{Row 4 (classical local controllers):}
Potential Field (bottom-left) fails (Success $=0$), consistent with a local-minimum/trapping failure mode near the bottleneck. By contrast, CBF and DWA (bottom-middle/right) both succeed (Success $=1$), but their rollouts show characteristic \emph{reactive behavior}: visible curvature/adjustments around obstacles and a less direct route than GRL-SNAM, reflecting oscillations or conservative arcs induced by local constraint satisfaction.}

Global planners succeed but generate jagged { or overly conservative paths that does not exploit near boundary squeezing path which can be shorter}.

\subsection{Learning {Strategies} in Dungeon Navigation: Deep RL vs.\ Hamiltonian Supervision}
\label{subsec:rl_baselines}

The dungeon point-agent task isolates the \emph{learning signal} under matched dynamics and observations: all methods share the same 2D continuous control, local occupancy encoding, and Transformer encoder; only the training objective differs.
We report (i) end-to-end, full-episode RL on the long-horizon MDP, and (ii) stagewise short-rollout training with the same observation/action interface used by GRL-SNAM.

\paragraph{Full-episode deep RL (long-horizon).}
We train PPO, TRPO, and SAC to reach a global goal within a 2000-step budget using only local observations.
Despite 5--8M environment transitions, success remains below 8\% and collision rates are $\sim$25\%, highlighting the difficulty of long-horizon credit assignment under partial observability.

\begin{table}[htbp]
\centering
\caption[Full-episode end-to-end RL baselines in the dungeon]{
Full-episode, end-to-end navigation baselines (PPO/TRPO/SAC) in the dungeon environment.
Despite substantial training budgets (5--8M environment steps), all three RL methods
fail to learn reliable goal-directed navigation under the dungeon’s partial/local observations:
success remains extremely low (1.5--7.2\%), collisions are frequent (23.5--28.3\%), and the
agent typically terminates far from the goal (mean residual distance 14--24\,m).
See Fig.~\ref{fig:dungeon_map} for qualitative evidence.%
}
\label{tab:full_episode_rl}
\small
\begin{tabular}{lcccc}
\toprule
Algorithm & Success (\%) & Mean goal dist.\ (m) & Collision (\%) & Env.\ training steps \\
\midrule
PPO  & 7.2 & 14.1 & 23.5 & 5M \\
TRPO & 3.8 & 23.7 & 26.1 & 6.5M \\
SAC  & 1.5 & 20.2 & 28.3 & 8M \\
\bottomrule
\end{tabular}
\end{table}

\paragraph{Stagewise short-rollout RL (matched interface).}
To remove interface confounds, we train the same RL algorithms on the \emph{same} stagewise short-rollout distribution and structured observations as GRL-SNAM: stage-relative state, stage-exit goal, and locally extracted obstacles, with actions $\mathbf{a}_t=[v_x,v_y]$.
RL improves over the full-episode setting, but still plateaus well below GRL-SNAM.

\begin{table}[htbp]
\centering
\caption[Stagewise short-rollout baselines vs.\ GRL-SNAM]{
Short-rollout, stagewise baselines vs.\ GRL-SNAM on the same dungeon-derived dataset and the same sensors.
While PPO/TRPO/SAC improve over full-episode training, they remain limited in reliability
(18.4--26.1\% success) and exhibit noticeably larger tracking error and residual goal distance.
In contrast, GRL-SNAM achieves 87.5\% success with substantially lower state error and goal distance, indicating that the learned GRL-SNAM surrogate produces actions that are both locally safe and globally progress-seeking. This advantage is displayed with the qualitative comparison in Fig.~\ref{fig:dungeon_map}.%
}

\small
\begin{tabular}{lcccc}
\toprule
Algorithm & Success (\%) & Mean state error (m) $\downarrow$ & Mean goal dist.\ (m) $\downarrow$ & Env.\ training\ steps \\
\midrule
PPO      & 26.1 & 1.8 & 1.2 & $\approx 3.2$M env.\ steps \\
TRPO     & 21.7 & 2.1 & 1.5 & $\approx 3.8$M env.\ steps \\
SAC      & 18.4 & 2.4 & 1.9 & $\approx 4.1$M env.\ steps \\
GRL-SNAM & \textbf{87.5} & \textbf{0.3} & \textbf{0.1} & 500k gradient steps \\
\bottomrule
\end{tabular}
\label{tab:short_rollout_rl}
\end{table}

\paragraph{Takeaway: objective dominates optimizer.}
With dynamics, sensing, and encoder fixed, swapping PPO/TRPO/SAC changes results modestly but does not close the gap.
The consistent improvement comes from \emph{Hamiltonian-structured supervision}: GRL-SNAM learns an energy model whose gradients encode goal attraction, barrier avoidance, and dissipation, rather than optimizing a sparse scalar reward.
This yields substantially higher success and lower terminal error in the same observation/action regime, indicating that the advantage is primarily due to the learning formulation rather than the choice of RL optimizer.

\subsection{Hamiltonian Field and Constraint Learning}
\label{subsec:hamiltonian_analysis}

We analyze the learned force fields, barrier structure, and replay contents on the hyperelastic ring under stagewise sensing.

\begin{figure}[htbp]
    \centering
\includegraphics[width=0.8\textwidth]{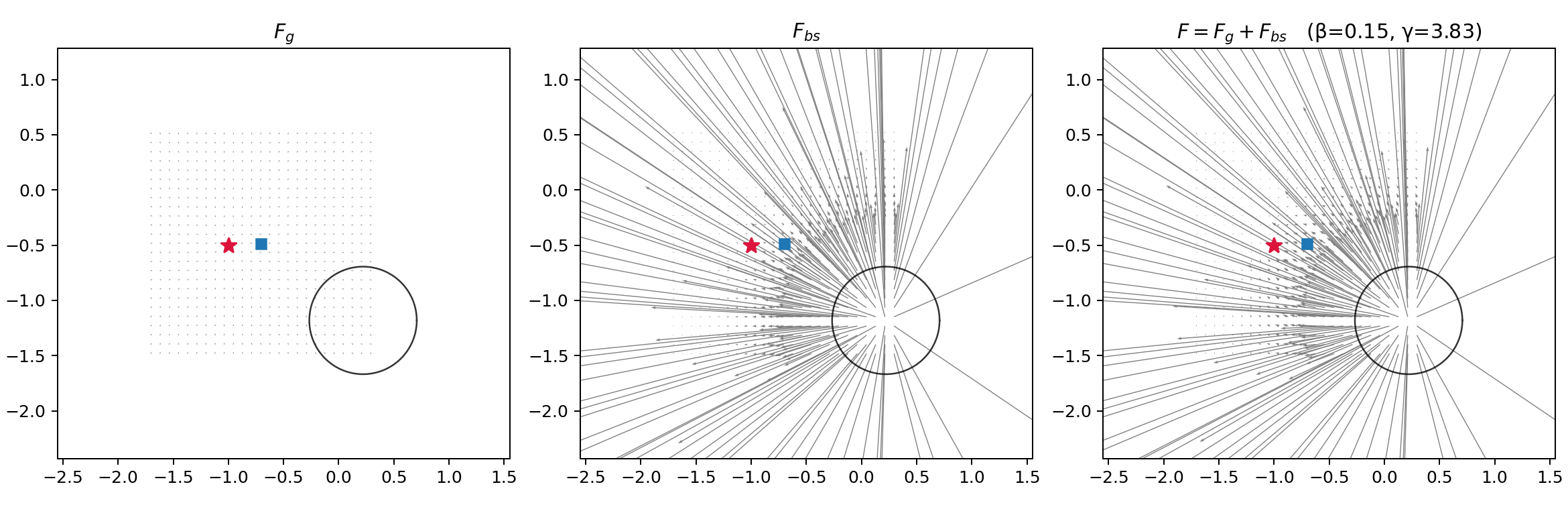}
\caption{\textbf{Learned Hamiltonian field decomposition.}
Shown are the constituent fields used by GRL-SNAM to generate local motion under partial, stagewise observations.
\textbf{Left:} $F_g$ points toward the current stage goal and captures progress, but ignores obstacles.
\textbf{Middle:} $F_{bs}$ is induced by locally sensed barriers and encodes avoidance/clearance, but lacks task direction.
\textbf{Right:} the deployed field $F=\beta F_g+\gamma F_{bs}$ combines both effects; the learned, context-dependent coefficients $(\beta,\gamma)$ modulate the relative strength of goal following vs.\ obstacle avoidance, producing a trajectory direction that threads the corridor while maintaining a safety margin near contact regions.}
\label{fig:vector_fields}
\end{figure}

\paragraph{Force-field decomposition.}
Figure~\ref{fig:vector_fields} visualizes the learned force field (negative differential of sub-modular potentials) as
goal attraction $F_g$, obstacle barriers $F_{bs}$, and their composition
\begin{equation}
F = \beta F_g + \gamma F_{bs}.
\label{eq:force_composition}
\end{equation}
$F_g$ alone is task-directed but obstacle-agnostic, while $F_{bs}$ encodes avoidance without a navigation objective.
Their composition produces smooth, clearance-preserving arcs around clutter via adaptive reweighting through $(\beta,\gamma)$.

\begin{figure}[htbp]
    \centering
    \includegraphics[width=0.6\textwidth]{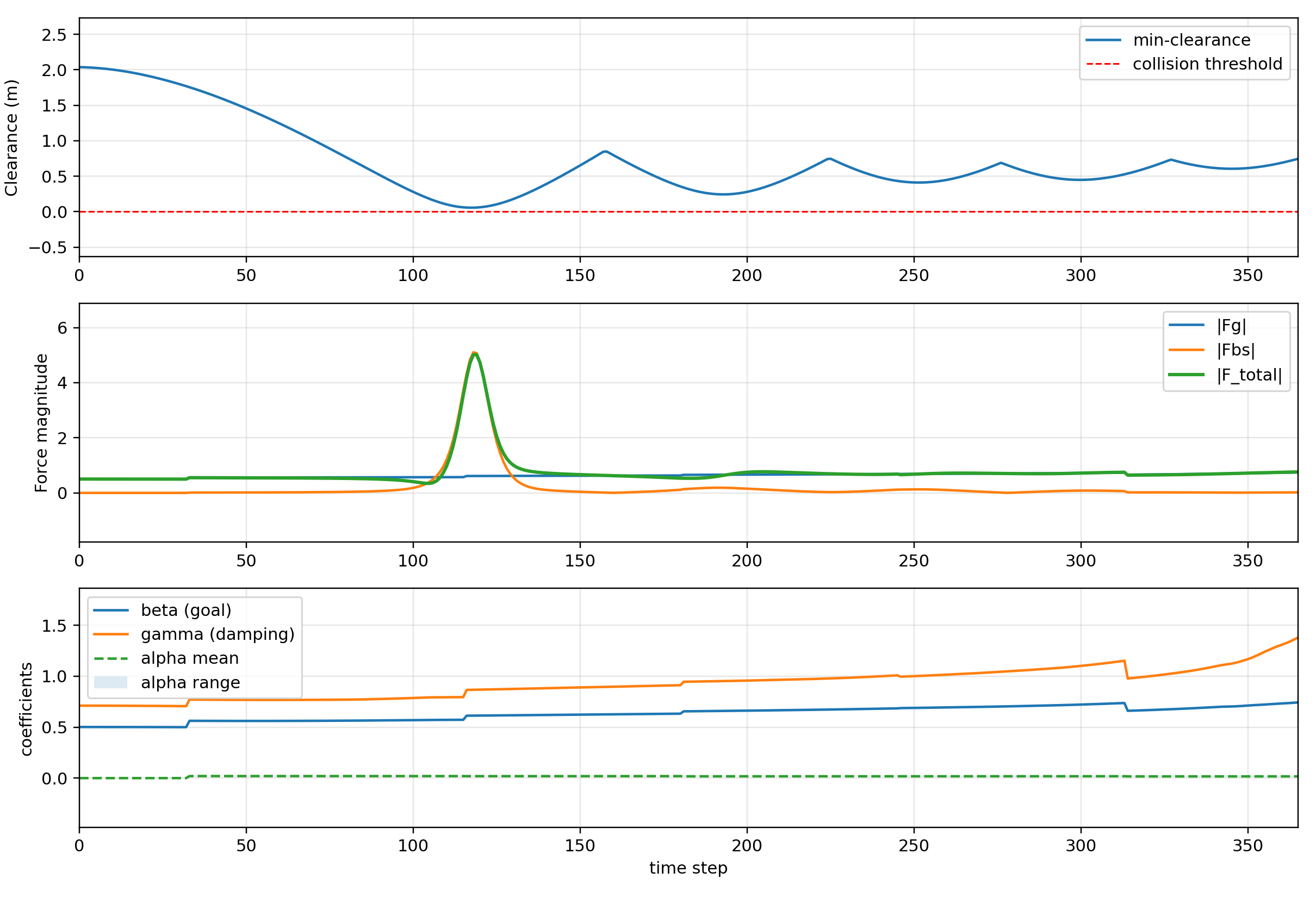}
    \caption{\textbf{Temporal analysis of Hamiltonian adaptation.}
    Top: minimum clearance. Middle: $|F_g|$ and $|F_{bs}|$. Bottom: coefficients $(\beta,\gamma,\alpha)$.
    Coefficients change smoothly as the robot enters/exits clutter, rebalancing goal and barrier terms.}
    \label{fig:time_series}
\end{figure}

\paragraph{Online coefficient adaptation.}
Along a representative rollout (Fig.~\ref{fig:time_series}), clearance decreases near bottlenecks and recovers after passage (top).
The magnitudes $|F_g|$ and $|F_{bs}|$ rebalance accordingly (middle), while coefficients $(\beta,\gamma,\alpha)$ vary smoothly over time (bottom), indicating adaptation occurs through the Hamiltonian parametrization rather than ad hoc action scaling.

\begin{figure}[htbp]
    \centering
    \includegraphics[width=0.55\textwidth]{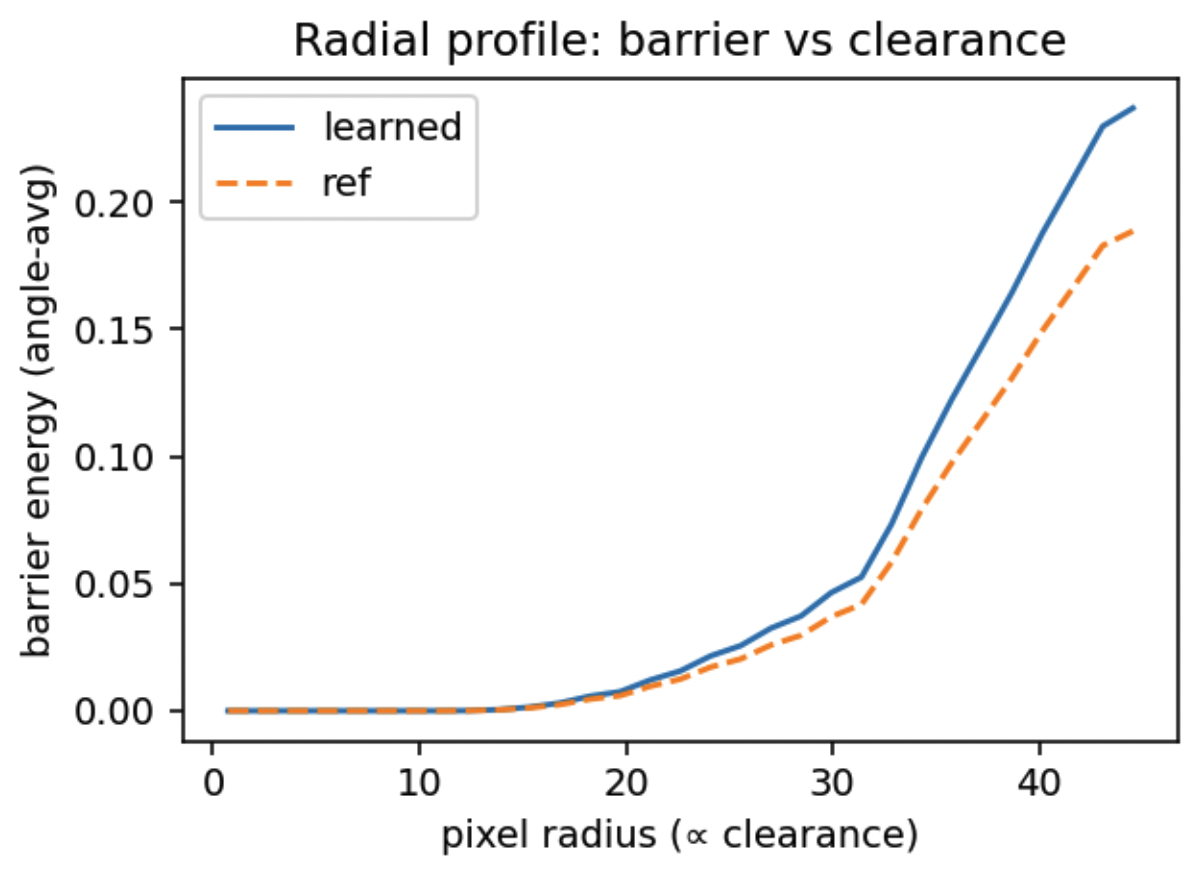}
    \caption{\textbf{Radial profile of barrier energy vs.\ clearance.}
    Angle-averaged barrier energy around obstacles for the learned barrier (solid) and analytic reference (dashed).
    The learned profile matches the reference trend and sharpens near contact. { The difference between two curves indicates Our method can learn to adapt the energy landscape based on its current environment context.}}
    \label{fig:barrier_profile}
\end{figure}

\paragraph{Empirical barrier learning.}
Figure~\ref{fig:barrier_profile} compares the learned obstacle barrier energy to the simulator’s analytic reference using an angle-averaged radial profile.
The learned barrier is near-zero at large clearance and rises sharply near contact, closely tracking the reference while slightly softening at intermediate distances and steepening near contact, consistent with additional deformation/friction costs captured in the rollout data.

\paragraph{Stagewise reuse and bootstrapped refinement.}
Stagewise parameterization induces a family of local policies (sensor/frame/shape controllers) that share parameters but adapt $(\beta,\gamma,\alpha)$ by stage.
As new windows are visited, local rollouts refine the Hamiltonian for those regions; revisits exploit this refinement without recomputing global plans.
Empirically, trajectories tighten toward least-cost paths over time while maintaining positive clearance.

\begin{figure}[htbp]
    \centering
    \includegraphics[width=0.75\linewidth]{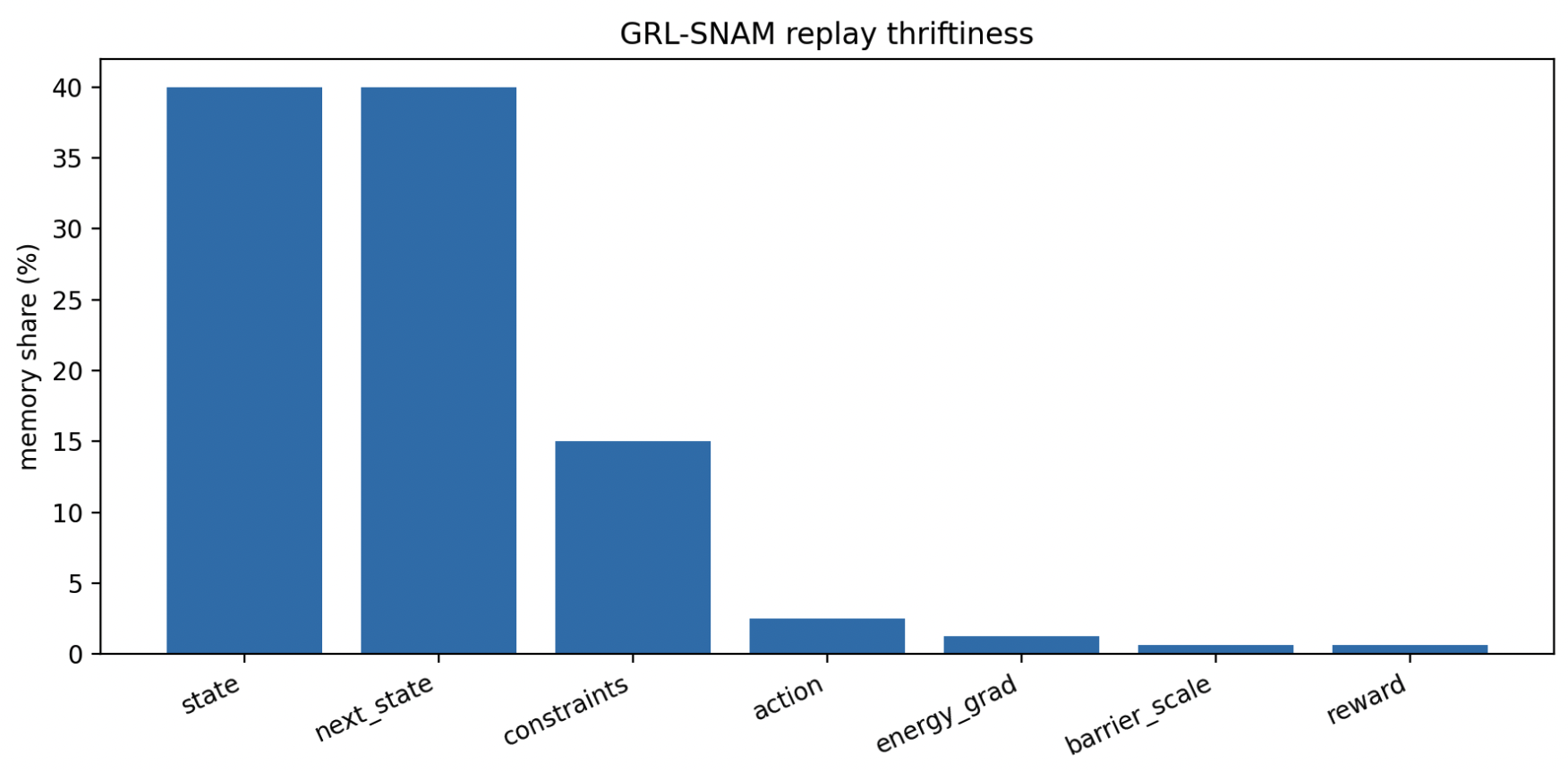}
    \caption{\textbf{Replay thriftiness in GRL-SNAM.}
    Memory share of buffer fields: $\sim$80\% local transitions $(q_t,q_{t+1})$, $\sim$15\% compact constraint/contact sets, and only a few percent for actions, gradients, barrier scales, or rewards. { GRL-SNAM is memory efficient as it only maintains the local state exhaustively while using a low dimensional network structure which effectively encodes the energy landscape.}}
    \label{fig:replay_buffer}
\end{figure}

\paragraph{Replay buffer thriftiness.}
Figure~\ref{fig:replay_buffer} shows buffer memory is dominated by local transitions $(q_t,q_{t+1})$ (about 80\%), with compact constraint/contact summaries (about 15\%); actions, gradients, barrier scales, and rewards occupy only a few percent.
The buffer therefore stores only what is needed for Hamiltonian updates—local dynamics and local constraints—without caching full maps or dense cost volumes.

\paragraph{Closed-loop Hamiltonian refinement.}
Stagewise interaction supplies short rollouts that populate the buffer (Fig.~\ref{fig:replay_buffer});
buffer updates adjust coefficients and barrier parameters, reshaping the local landscape (Figs.~\ref{fig:vector_fields}--\ref{fig:barrier_profile});
the updated landscape then induces the next round of dynamics.
This yields a closed-loop procedure that repeatedly refines $\mathcal{H}_\theta$ from local experience.

\subsection{Ablations and Robustness}
\label{subsec:ablations_robustness}

We finally probe how specific design choices in GRL-SNAM affect performance, and how well the learned Hamiltonian controller tolerates sensing and dynamics perturbations.

\paragraph{Effect of loss components.}
Recall that our navigation surrogate is trained with a weighted multi-term loss
\begin{equation}
\mathcal{L} = w_{\text{traj}} \mathcal{L}_{\text{traj}} + w_{\text{vel}} \mathcal{L}_{\text{vel}} + w_{\text{friction}} \mathcal{L}_{\text{friction}} + w_{\text{multi}} \mathcal{L}_{\text{multi}},
\label{eq:total_loss}
\end{equation}
where $\mathcal{L}_{\text{traj}}$ and $\mathcal{L}_{\text{vel}}$ supervise trajectory and velocity matching, $\mathcal{L}_{\text{friction}} = \|\gamma - \gamma_o\|_2^2$ aligns the learned damping with the stagewise reference, and $\mathcal{L}_{\text{multi}}$ trains near-contact robustness via perturbed short rollouts.
We ablate the last two terms while keeping environments and architectures fixed.

\begin{table}[htbp]
\centering
\caption{\textbf{Ablation of loss terms} on the hyperelastic ring benchmark (Test-ID/OOD).
Arrows denote desired direction. Icons summarize consistent qualitative trends across several seeds.}
\small
\setlength{\tabcolsep}{6pt}
\begin{adjustbox}{valign=c, width=\textwidth}
\begin{tabular}{lcccccc}
\toprule
\textbf{Variant} & \textbf{Collisions} $\downarrow$ & \textbf{MinClr} $\uparrow$ & \textbf{Barrier Viol.} $\downarrow$ & \textbf{Progress/SPL} $\uparrow$ & \textbf{Smoothness} $\downarrow$ & \textbf{Observed behavior}\\
\midrule
$w_{\text{fric}} = 0,\ w_{\text{multi}} = 0$    & \textcolor{red}{High} ($\times$) & \textcolor{red}{$<0$} & \textcolor{red}{High} & \textcolor{red}{Poor} & \textcolor{red}{Poor} & Penetrates obstacles \\
$w_{\text{fric}} = 0,\ w_{\text{multi}} = 0.5$   & \textcolor{green!50!black}{Low}~($\checkmark$) & \textcolor{green!50!black}{High} & \textcolor{green!50!black}{Low} & \textcolor{orange}{Low} & \textcolor{green!50!black}{OK} & Very slow, conservative \\
$w_{\text{fric}} = 0.1,\ w_{\text{multi}} = 0$   & \textcolor{green!50!black}{None}~(\checkmark) & \textcolor{green!50!black}{High} & \textcolor{green!50!black}{Low} & \textcolor{green!50!black}{High} & \textcolor{green!50!black}{Best} & Smooth, stable, fast \\
$w_{\text{fric}} = 0.1,\ w_{\text{multi}} = 0.5$ & \textcolor{green!50!black}{None}~($\checkmark$) & \textcolor{orange}{Slightly lower} & \textcolor{green!50!black}{Low} & \textcolor{green!50!black}{High} & \textcolor{green!50!black}{Good} & Stable; tighter margins \\
\bottomrule
\end{tabular}
\end{adjustbox}

\label{tab:ablation_fric_multi}
\end{table}

Table~\ref{tab:ablation_fric_multi} shows that:
(i) without $\mathcal{L}_{\text{friction}}$ or $\mathcal{L}_{\text{multi}}$, the learned field is under-damped and oscillatory, leading to barrier violations and penetrations;
(ii) $\mathcal{L}_{\text{multi}}$ alone enforces safety but at the cost of extremely slow motion, as the learner over-weights near-contact failures and suppresses progress;
(iii) $\mathcal{L}_{\text{friction}}$ alone is sufficient to eliminate penetrations and produce smooth, fast trajectories; and
(iv) combining both terms yields the best overall trade-off, with slightly tighter clearances but robust, stable motion in clutter.
These trends are consistent with the Hamiltonian interpretation: $\mathcal{L}_{\text{friction}}$ learns an appropriate dissipation term, preventing “ringing’’ of the barrier energy, while $\mathcal{L}_{\text{multi}}$ selectively shapes the field near constraints.

\paragraph{Robustness to sensing and dynamics perturbations.}
We next evaluate GRL-SNAM under controlled perturbations to sensing and dynamics.
We vary position jitter and radius estimation error in the local obstacle extractor, randomly drop or hallucinate obstacles, and perturb the damping coefficient $\gamma$ and applied velocities.
Each start–goal pair is rolled out across a grid of perturbation levels, yielding a total of $N = n_{\text{env}} \times n_{\text{trials}} \times n_{\text{perturbations}}$ runs.

\begin{table}[htbp]
\centering
\caption{Robustness of GRL-SNAM to sensing noise and dynamics perturbations.
Perturbation settings are reported as (sensing noise, damping scale). { Even when noise level in sensor and motion is very high (Severe Noise), our method is still robust and achieve 87\% overall success rate, which is better comparing against RL methods.}}
\label{tab:robustness}
\vspace{0.5em}
\begin{tabular}{lcccc}
\toprule
Perturbation Level & Success (\%) & SPL $\uparrow$ & Min. Clearance (m) $\uparrow$ & Collisions $\downarrow$ \\
\midrule
Nominal (0.0, 1.0)      & 98.7 & 0.82 & 0.36 & 0.3 \\
Mild Noise (0.05, 0.9)  & 91.3 & 0.79 & 0.33 & 0.7 \\
Severe Noise (0.10, 0.7) & 87.1 & 0.72 & 0.29 & 1.1 \\
\bottomrule
\end{tabular}
\end{table}

As shown in Table~\ref{tab:robustness}, GRL-SNAM maintains high success rates ($\geq 87\%$) and positive clearance even under severe noise and dynamics mismatch.
SPL and clearance degrade gracefully as perturbations increase, while collision counts remain low.
Rather than failing catastrophically when constraints or dynamics are mis-specified, the learned Hamiltonian field re-weights goal and barrier terms in response to distorted local observations (cf.\ Figure~\ref{fig:time_series}), preserving a meaningful safety–progress trade-off.

\paragraph{Takeaways.}
The performance of
(i) the friction and multi-start losses play distinct, interpretable roles in learning dissipation and near-contact robustness in the Hamiltonian surrogate;
(ii) performance is not brittle to moderate changes in sensing or dynamics, because GRL-SNAM adapts by reshaping the energy landscape rather than overfitting to a single nominal model; and
(iii) these properties are emergent from the same empirical Hamiltonian-modification pipeline analyzed in Section~\ref{subsec:hamiltonian_analysis}, reinforcing our claim that GRL-SNAM provides a stable, constraint-aware alternative to standard reward-based policies for stagewise navigation.

\section{Conclusion}
We introduce GRL-SNAM, a {geometric} reinforcement learning framework that leverages Hamiltonian structure to couple sensing, planning, and deformation into a unified energy-based policy. Our formulation enables stable, feedforward navigation updates and achieves near-optimal path quality with minimal mapping effort in challenging deformable-robot tasks. The results highlight that incorporating geometric priors into RL can yield both efficiency and robustness, even under noisy sensing and out-of-distribution layouts. Future work will extend the approach to richer sensing modalities and more complex environments, with the goal of validating its scalability to real robotic systems.

\section*{Acknowledgement}
This research was supported in part  from the Peter O’Donnell Foundation, the Jim Holland- Backcountry Foundation and in part from a grant from the Army Research Office accomplished under Cooperative Agreement Number W911NF-19-2-0333.
\newpage

\bibliography{refs, iclr2026_conference}
\bibliographystyle{iclr2026_conference}

\appendix
\newpage
\appendix

\section{Identifiability of GRL-SNAM}
\label{app:proof:and:lemma}
\subsection{From optimal control to reduced Hamiltonian dynamics }
\label{app:ocp_to_ham}
\begin{lemma}[Forward: autonomous pure state constraints induce a (reduced) Hamiltonian flow]
\label{thm:forward_state_constrained_pmp_reduced_ham}
Fix a scenario $\mathcal{E}$ and consider the autonomous, control-affine system on an open set
$\mathcal{Q}\subset\mathbb{R}^n$,
\begin{equation}
\dot q(t)=f(q(t)) + A(q(t))u(t),\qquad q(0)=q_0,\qquad q(T)=q_T,
\label{eq:ocp_dyn}
\end{equation}
with pure state inequality constraints
\begin{equation}
d_i(q(t))\ge 0,\qquad i=1,\dots,m,\qquad t\in[0,T],
\label{eq:ocp_state_constraints}
\end{equation}
and objective
\begin{equation}
\min_{u(\cdot)\in\mathcal{U}} \ J(q, u)
:=\int_0^T L(q(t),u(t))\,dt,
\label{eq:ocp_cost}
\end{equation}
where the admissible control set $\mathcal{U}$ consists of measurable $u(\cdot)$ with
$u(t)\in U\subset\mathbb{R}^d$ a.e.

Assume:
\begin{enumerate}
\item $f$ and $A$ are locally Lipschitz on $\mathcal{Q}$.
\item $U$ is nonempty, compact, and convex; $L(\cdot,\cdot)$ is continuous and $C^1$ in $(q,u)$.
\item (Existence) There exists an optimal pair $(q^\star(\cdot),u^\star(\cdot))$ satisfying
\cref{eq:ocp_dyn} and \cref{eq:ocp_state_constraints}.
\item (Regularity / qualification) Each $d_i\in C^2(\mathcal{Q})$ and along boundary arcs the active constraint
gradients satisfy a standard constraint qualification (e.g.\ linear independence of
$\{\nabla d_i(q^\star(t))\}_{i\in\mathcal{A}(t)}$ and a feasible-direction condition), so that the
state-constrained maximum principle applies.
\end{enumerate}

Define the \emph{control Hamiltonian} as \textit{Fenchel Coupling}
\begin{equation}
\mathcal{H}_c(q,p,u):=\langle f(q)+A(q)u, p\rangle-L(q,u).
\label{eq:def_control_ham}
\end{equation}
Then there exist:
(i) a costate $p:[0,T]\to\mathbb{R}^n$ of bounded variation(\textbf{BV}, right-continuous with left limits),
and (ii) nonnegative finite Radon measures $\{\mu_i\}_{i=1}^m$ on $[0,T]$,
such that the following hold:

\smallskip
\noindent\textbf{(i) Maximum condition (a.e.).}
For almost every $t\in[0,T]$,
\begin{equation}
u^\star(t)\in \arg\max_{u\in U}\ \mathcal{H}_c\big(q^\star(t),p(t),u\big).
\label{eq:max_condition}
\end{equation}
\noindent\textbf{(ii) State equation.}
For a.e.\ $t$,
\begin{equation}
\dot q^\star(t)=\nabla_p \mathcal{H}_c\big(q^\star(t),p(t),u^\star(t)\big)
= f(q^\star(t))+A(q^\star(t))u^\star(t).
\label{eq:state_eq_from_ham}
\end{equation}
\noindent\textbf{(iii) Adjoint equation in measure form.}
As an equality of $\mathbb{R}^n$-valued measures on $[0,T]$,
\begin{equation}
dp(t)= -\nabla_q \mathcal{H}_c\big(q^\star(t),p(t),u^\star(t)\big)\,dt
\;-\;\sum_{i=1}^m \nabla d_i\big(q^\star(t)\big)\,\mu_i(dt).
\label{eq:adjoint_measure}
\end{equation}
Equivalently, for all $0\le t_0<t_1\le T$,
\begin{equation}
p(t_1^+)-p(t_0^-)
= -\int_{t_0}^{t_1}\nabla_q \mathcal{H}_c\big(q^\star(t),p(t),u^\star(t)\big)\,dt
\;-\;\sum_{i=1}^m \int_{(t_0,t_1]} \nabla d_i\big(q^\star(t)\big)\,\mu_i(dt).
\label{eq:adjoint_integral}
\end{equation}
\noindent\textbf{(iv) Complementarity / support.}
Each $\mu_i$ is supported on the active set:
\begin{equation}
\mathrm{supp}(\mu_i)\subset \{t\in[0,T]: d_i(q^\star(t))=0\},
\qquad\text{i.f.f.}\qquad
\int_{[0,T]} d_i(q^\star(t))\,\mu_i(dt)=0.
\label{eq:complementarity_measure}
\end{equation}
\noindent\textbf{(iv) Reduced Hamiltonian.} Assume in addition that the running cost decomposes as
\begin{equation}
L(q,u)=\varphi(u)+V(q),
\qquad\text{with $\varphi$ proper, closed, strictly convex, and $C^1$, and $V\in C^1(\mathcal Q)$.}
\label{eq:cost_split_B_statecon}
\end{equation}
Then the maximizer $u^\star(q,p)$ in \cref{eq:max_condition} is unique (up to the compact control constraint $U$),
and the reduced PMP Hamiltonian depends only on $(q,p)$:
\begin{equation}
\label{eq:def_reduced_ham_pmp_statecon}
H_{\mathrm{PMP}}(q,p)
:=\sup_{u\in U}\mathcal H_c(q,p,u)
=\sup_{u\in U}\big\{p^\top A(q)u-\varphi(u)\big\}+p^\top f(q)-V(q).
\end{equation}
Hence \cref{eq:state_eq_from_ham} and \cref{eq:adjoint_measure} can be restated without explicit dependency on $u^\star$.
\end{lemma}

\begin{proof}
We present the derivation in steps and refer readers Theorem 4.2 in \cite{hartl1995survey} for further discussions. The proof contains major steps including reducing the general state-constrained PMP statement
to the present autonomous, pure-state-constraint setting and then eliminating $u$ via conjugacy.

\smallskip
\noindent\textbf{Step 1 (Apply the state-constrained maximum principle).}
Under assumptions (1)--(4), the state-constrained maximum principle provides a costate of bounded variation and
a bounded-variation multiplier associated with the state constraints, together with an integral adjoint relation.
Specializing to \emph{pure} state constraints (no mixed constraints) yields an identity of the form
\cref{eq:adjoint_integral} except that the state-constraint multiplier appears as a BV function $\zeta_i(\cdot)$.
Writing its Stieltjes increment as $d\zeta_i(t)$ and using the BV decomposition, define the nonnegative measure
$\mu_i := -d\zeta_i$; then $\mu_i\ge 0$ and \cref{eq:adjoint_integral} follows.

\smallskip
\noindent\textbf{Step 2 (Maximum condition from needle variations).}
Fix a Lebesgue point $t$ for $(q^\star,u^\star)$ and consider a needle variation
$u_\varepsilon(\cdot)$ that equals $u^\star(\cdot)$ except on a small interval $(t,t+\varepsilon)$ where it is set
to a constant $\bar u\in U$.
Using (A1)--(A2), the first variation of $J$ exists and is linear in $\bar u-u^\star(t)$, while feasibility of
\cref{eq:ocp_state_constraints} is handled by the measure multipliers $\mu_i$.
The necessary condition that the first variation is nonnegative for all $\bar u\in U$ implies
\cref{eq:max_condition} for almost every $t$.

\smallskip
\noindent\textbf{Step 3 (Adjoint equation from state variations + integration by parts).}
Let $\delta q$ be an admissible first-order state variation induced by admissible control variations.
Write the first variation of the Lagrangian functional associated with \cref{eq:ocp_dyn} using a multiplier $p(\cdot)$,
and integrate by parts on each interval of continuity of $p$.
The terms involving the state constraints appear through Stieltjes integrals against $d\zeta_i$, i.e.,
through $\int \nabla c_i(q^\star(t))\,d\zeta_i(t)$; converting to $\mu_i=-d\zeta_i$ yields \cref{eq:adjoint_measure}.

\smallskip
\noindent\textbf{Step 4 (Complementarity/support).}
Because each $\zeta_i(\cdot)$ is constant on interior intervals where $c_i(q^\star(t))>0$,
its Stieltjes increment vanishes there; hence $\mu_i=-d\zeta_i$ only charges times when the constraint is active.
This is exactly \cref{eq:complementarity_measure}.

\smallskip
\noindent\textbf{Step 5 (Reduced Hamiltonian by Legendre--Fenchel conjugacy).}
~Define the \emph{reduced Hamiltonian} (pointwise supremum over $u$)
\begin{equation}
H_{\mathrm{PMP}}(q,p)
:=\sup_{u\in U}\Big\{p^\top A(q)u-\varphi(u)\Big\}+p^\top f(q)-V(q).
\end{equation}
By strict convexity of $\varphi$, the maximizer $u^\star(q,p)$ is unique and satisfies the first-order condition
$\nabla\varphi(u^\star)=A(q)^\top p$ (when $U=\mathbb{R}^d$; for compact $U$ use the argmax form).
Hence $u^\star(q,p)=\nabla\varphi^\ast(A(q)^\top p)$, and Danskin's theorem yields
\[
\begin{aligned}
\nabla_p H_{\mathrm{PMP}}(q,p)&=f(q)+A(q)u^\star(q,p), \\
\nabla_q H_{\mathrm{PMP}}(q,p)&=\big(\nabla_q f(q)\big)^\top p+\big(\nabla_q A(q)\,u^\star(q,p)\big)^\top p-\nabla V(q).    
\end{aligned}
\]
Substituting $u^\star(t)=u^\star(q^\star(t),p(t))$ into
\cref{eq:state_eq_from_ham}--\cref{eq:adjoint_measure} gives the \emph{reduced} constrained Hamiltonian flow
\begin{equation}
\begin{aligned}
\dot q^\star(t)&=\nabla_p H_{\mathrm{PMP}}\big(q^\star(t),p(t)\big), \\ 
dp(t)&= -\nabla_q H_{\mathrm{PMP}}\big(q^\star(t),p(t)\big)\,dt -\sum_{i=1}^m \nabla d_i(q^\star(t))\,\mu_i(dt),    
\end{aligned}
\end{equation}
which is precisely the claimed Hamiltonian outcome of the state-constrained optimal control problem.
\end{proof}

\begin{corollary}[Reduced Hamiltonian via a quadratic control penalty]
\label{cor:mechanical_reduced_hamiltonian}
Consider the control--affine dynamics $\dot q=f(q;\mathcal E)+A(q;\mathcal E)u$ and assume the { cost-to-go functional} splits as
$L(q,u;\mathcal E)=\varphi(u) + {V}(q;\mathcal E)$ with a quadratic effort penalty
\(
\varphi(u)=\tfrac12 u^\top \Phi u
\)
 for some $\Phi\succ 0$. Assume $f(\cdot;\mathcal E)\equiv 0$ and $A(q;\mathcal{E})\equiv A$.
Then the reduced Hamiltonian (i.e.\ $\sup_u\{p^\top(f+A u)-L\}$) equals
\[
H(q,p;\mathcal E)
=\tfrac12\,p^\top\!\big(A(q;\mathcal E)\Phi^{-1}A(q;\mathcal E)^\top\big)p -V(q;\mathcal E).
\]
In particular, if we define the (inverse) mass by
\[
M(q;\mathcal E)^{-1}:=A(q;\mathcal E)\Phi^{-1}A(q;\mathcal E)^\top\in\mathbb S_{++}^2,
\]
then \(H(\cdot,\cdot;\mathcal E)\in\mathscr H\), where
\[
\mathscr{H}_{PMP}:=\Big\{\,H(q,p;\mathcal{E})=\tfrac12\,p^\top M^{-1}p{-}V(q;\mathcal{E})
\ \Big|\ M\in \mathbb{S}_{++}^2 \Big\}.
\]
\end{corollary}

\begin{proof}
By definition of the PMP reduced Hamiltonian,
\[
H_{PMP}(q,p;\mathcal E)=\sup_{u}\Big\{p^\top(f(q;\mathcal E)+A(q;\mathcal E)u)-\varphi(u)- V(q;\mathcal E)\Big\}.
\]
Setting $f\equiv 0$ reduces the supremum to$\sup_u\{(A^\top p)^\top u-\tfrac12 u^\top \Phi u\}{-V}$.
The maximizer satisfies $\Phi u^\star=A^\top p$, hence $u^\star=\Phi^{-1}A^\top p$ and the supremum value is
$\tfrac12 p^\top A\Phi^{-1}A^\top p { - V}$. Since $\Phi\succ 0$ and $A(\cdot;\mathcal E)$ has full rank, the induced
matrix $A\Phi^{-1}A^\top$ is positive definite, so $M^{-1}\in\mathbb S_{++}^2$ and $H\in\mathscr H$.
\end{proof}
{
\paragraph{Barrier relaxation of state-constraint OCPs}\label{rem:barrier_relaxation}
In the implementation one replaces hard constraints $d_i(q)\ge 0$ by a smooth barrier/penalty term
$B_\mu(q):=\mu \sum_i b(d_i(q))$ and optimizes an \emph{unconstrained} problem with augmented running cost
$L_\mu(q,u,t)=L(q,u,t)+B_\mu(q)$.
This does not require introducing BV multipliers in the PMP dynamics; instead it modifies the potential (equivalently, modifies $\nabla_q H_{PMP}$) by adding $\nabla_q B_\mu$. From the inverse-design viewpoint, learning barrier weights corresponds to selecting a particular absolutely continuous approximation of the constraint measure $d\zeta$ (as a standard exact-penalty / interior-point interpretation). For our GRL-SNAM navigator, we provide additional statements (c.f. \cref{prop:barrier_to_state_constrained_complete} and \cref{cor:ipc_from_logbarrier}) to address the approximability of our choice of Hamiltonians.
}

\begin{proposition}[Barrier relaxation $\mu\downarrow 0$ yields a state-constrained limit]
\label{prop:barrier_to_state_constrained_complete}
Fix $\mathcal E$ and consider the deterministic state-constrained OCP
\begin{equation}
\begin{aligned}
\label{eq:OCP_hard}
\min_{u(\cdot)}&\ J(u):=\int_0^T L(q(t),u(t))\,dt\\
 \text{s.t.}&\quad 
\dot q=f(q)+A(q)u,\ \ q(0)=q_0,\ q(T)=q_T,\ \ d_i(q(t))\ge 0\ \forall t,\ i=1,\dots,m .
\end{aligned}
\end{equation}
Assume:
\begin{enumerate}[(H1)]
\item $U\subset\mathbb R^d$ is compact and convex, $u(\cdot)\in L^\infty([0,T];U)$.
\item $f,A$ are globally Lipschitz on $\mathbb R^n$; $L(\cdot,\cdot)$ is continuous and convex in $u$,
and there exist $c_0,c_1>0$ such that $L(q,u)\ge -c_0+c_1\|u\|^2$.
\item Each $d_i\in C^2(\mathbb R^n)$ and the feasible set of \cref{eq:OCP_hard} is nonempty.
Moreover, (\emph{Slater}) there exists a strictly feasible pair $(\bar q,\bar u)$ satisfying $d_i(\bar q(t))\ge \delta$
for all $t\in[0,T]$ and all $i$, for some $\delta>0$.
\item (\emph{Existence of an optimal solution}) \cref{eq:OCP_hard} admits at least one optimal pair $(q^\star,u^\star)$.
\end{enumerate}
For $\mu>0$, define the \emph{log-barrier relaxed} (unconstrained) problems
\begin{equation}
\begin{aligned}
\label{eq:OCP_barrier}
\min_{u(\cdot)}&\ J_\mu(u)
:=\int_0^T \Big( L(q(t),u(t)) + \mu\sum_{i=1}^m \alpha_i\,b(d_i(q(t)))\Big)\,dt \\
\text{s.t.}&\quad
\dot q=f(q)+A(q)u,\ q(0)=q_0,\ q(T)=q_T,
\end{aligned}
\end{equation}
where $\alpha_i>0$ are fixed weights and $b(s)=-\log s$ (so \cref{eq:OCP_barrier} is defined only on trajectories with
$d_i(q(t))>0$ for all $t$). Let $(q^\mu,u^\mu)$ be a minimizer of \cref{eq:OCP_barrier}.

Then:
\begin{enumerate}[(a)]
\item (\emph{Strict feasibility}) For every $\mu>0$, $d_i(q^\mu(t))>0$ for all $t\in[0,T]$ and all $i$.
\item (\emph{Subsequence convergence}) For any sequence $\mu_k\downarrow 0$, there exists a subsequence (not relabeled) and
a feasible pair $(q^0,u^0)$ for \cref{eq:OCP_hard} such that
\[
q^{\mu_k}\to q^0 \ \text{uniformly on }[0,T],
\qquad
u^{\mu_k}\rightharpoonup^\ast u^0 \ \text{in }L^\infty([0,T];\mathbb R^d).
\]
\item (\emph{Optimality of cluster points}) Any cluster point $(q^0,u^0)$ obtained in (b) is optimal for \cref{eq:OCP_hard}, and
the optimal values satisfy $\lim_{\mu\downarrow 0}\inf J_\mu = \inf J$.
\end{enumerate}

Moreover, assume in addition:
\begin{enumerate}[(H5)]
\item (\emph{Uniform multiplier tightness}) The measures $\mu_i^\mu(dt):=\lambda_i^\mu(t)\,dt$ defined by
\begin{equation}
\label{eq:lambda_def}
\lambda_i^\mu(t):=\mu\,\alpha_i\,\frac{1}{d_i(q^\mu(t))}\ge 0
\quad\Longleftrightarrow\quad
\mu\,\alpha_i\,\nabla b(d_i(q^\mu(t))) = -\lambda_i^\mu(t)\,\nabla d_i(q^\mu(t))
\end{equation}
have uniformly bounded total mass:
$\sup_{\mu>0}\mu_i^\mu([0,T])<\infty$ for each $i$.
\end{enumerate}
Then, along the subsequence in (b), $\mu_i^{\mu_k}\overset{\ast}{\rightharpoonup} \mu_i$ as finite Radon measures,
and the limiting pair $(q^0,u^0)$ admits a costate $p$ and measures $\{\mu_i\}$ satisfying the state-constrained PMP in measure form.
In particular, complementarity holds:
\begin{equation}
\label{eq:measure_complementarity_limit}
\int_{[0,T]} d_i(q^0(t))\,\mu_i(dt)=0,\qquad \mathrm{supp}(\mu_i)\subset\{t:\ d_i(q^0(t))=0\}.
\end{equation}
\end{proposition}

\begin{proof}
\textbf{Step 1 (Existence and strict feasibility for the barrier problems).}
Fix $\mu>0$. By (H3) there exists at least one strictly feasible trajectory, hence $J_\mu$ is finite on a nonempty set.
Since $b(s)=-\log s$ diverges to $+\infty$ as $s\downarrow 0$ and is undefined for $s\le 0$,
any admissible minimizer must satisfy $d_i(q^\mu(t))>0$ for all $t$ and $i$, proving (a).
Existence of a minimizer for \cref{eq:OCP_barrier} follows from standard deterministic OCP existence arguments:
(H1) provides weak-$^\ast$ compactness of controls in $L^\infty$; (H2) gives existence/uniqueness and continuous dependence
of $q$ on $u$ (via global Lipschitz dynamics), and lower semicontinuity of the integral cost under weak-$^\ast$ convergence
(using convexity in $u$). Thus a minimizing sequence has a convergent subsequence whose limit attains the infimum.

\textbf{Step 2 (Compactness of $\{(q^\mu,u^\mu)\}$ as $\mu\downarrow 0$).}
Let $\bar u$ be the strictly feasible Slater control from (H3), and $\bar q$ its trajectory.
For any minimizer $(q^\mu,u^\mu)$,
\[
J_\mu(u^\mu)\le J_\mu(\bar u)=\int_0^T L(\bar q(t),\bar u(t))\,dt + \mu\sum_i \alpha_i\int_0^T b(d_i(\bar q(t)))\,dt
\le C_0 + \mu C_1,
\]
where $C_0,C_1<\infty$ are constants (since $d_i(\bar q(t))\ge\delta>0$ implies $b(d_i(\bar q(t)))$ is bounded).
By (H2), $L(q,u)\ge -c_0+c_1\|u\|^2$ yields a uniform $L^2$ bound on $u^\mu$ (hence also boundedness in $L^\infty$ because $u^\mu(t)\in U$).
Therefore $\{u^\mu\}$ is bounded in $L^\infty$, so by Banach--Alaoglu there exists a subsequence $\mu_k\downarrow 0$
with $u^{\mu_k}\rightharpoonup^\ast u^0$ in $L^\infty$.
By Lipschitz dynamics and bounded controls, the corresponding states $q^{\mu_k}$ are equi-Lipschitz and uniformly bounded,
so by Arzel\`a--Ascoli, $q^{\mu_k}\to q^0$ uniformly for some absolutely continuous $q^0$.
Passing to the limit in the ODE (continuous dependence of solutions on controls) gives that $q^0$ is the trajectory induced by $u^0$.
Since each $d_i(q^{\mu_k}(t))>0$ and $d_i$ is continuous, the uniform limit implies $d_i(q^0(t))\ge 0$ for all $t$,
so $(q^0,u^0)$ is feasible for \cref{eq:OCP_hard}. This proves (b).

\textbf{Step 3 (Value convergence and optimality of cluster points).}
We prove $\lim_{\mu\downarrow 0}\inf J_\mu = \inf J$ and that any cluster point is optimal.

\emph{(i) {Lim-inf} inequality.}
For the minimizing sequence $u^{\mu_k}$,
\[
\inf J_{\mu_k}=J_{\mu_k}(u^{\mu_k}) \ge \int_0^T L(q^{\mu_k}(t),u^{\mu_k}(t))\,dt = J(u^{\mu_k}),
\]
since the barrier term is nonnegative.
By convexity of $L$ in $u$ and uniform convergence $q^{\mu_k}\to q^0$, the integral functional $u\mapsto\int_0^T L(q[u](t),u(t))dt$
is weak-$^\ast$ lower semicontinuous, hence
\[
\liminf_{k\to\infty} \inf J_{\mu_k} \ \ge\ \liminf_{k\to\infty} J(u^{\mu_k})\ \ge\ J(u^0)\ \ge\ \inf J.
\]

\emph{(ii) {Lim-sup} inequality.}
Fix any feasible control $u$ for \cref{eq:OCP_hard} with trajectory $q$ (so $d_i(q(t))\ge 0$).
Let $(\bar q,\bar u)$ be the Slater strictly feasible pair with margin $\delta>0$.
Define a perturbed control
\[
u^{(\epsilon)} := (1-\epsilon)u+\epsilon \bar u,\qquad \epsilon\in(0,1).
\]
Then $u^{(\epsilon)}\in U$ a.e.\ by convexity of $U$.
Let $q^{(\epsilon)}$ be the induced trajectory.
By continuous dependence of the ODE solution on controls, $q^{(\epsilon)}\to q$ uniformly as $\epsilon\downarrow 0$.
Moreover, since $(\bar q,\bar u)$ is strictly feasible and $d_i$ is continuous, there exists $\epsilon_0>0$ such that
for all $\epsilon\in(0,\epsilon_0)$, $q^{(\epsilon)}$ is \emph{strictly feasible}, i.e.\ $\min_{t\in[0,T]} d_i(q^{(\epsilon)}(t))>0$
for all $i$ (intuitively: the perturbation moves the trajectory into the interior; formally, use uniform convergence plus
$d_i(\bar q(t))\ge\delta$ and continuity of $d_i$).
Hence $J_\mu(u^{(\epsilon)})$ is well-defined for all $\mu>0$.

Now choose a sequence $\epsilon=\epsilon(\mu)\downarrow 0$ such that
\[
\epsilon(\mu)\downarrow 0
\qquad\text{and}\qquad
\mu\,\log\!\big(1/\epsilon(\mu)\big)\to 0\quad(\mu\downarrow 0),
\]
e.g.\ $\epsilon(\mu)=\exp(-\sqrt{1/\mu})$.
Then $q^{(\epsilon(\mu))}\to q$ uniformly and $J(u^{(\epsilon(\mu))})\to J(u)$ by continuity of $L$ and dominated convergence theorem.
For the barrier term, strict feasibility implies $d_i(q^{(\epsilon(\mu))}(t))\ge c\,\epsilon(\mu)$ for some $c>0$
on $[0,T]$ (possibly after shrinking $\epsilon_0$); thus
\[
0\le \mu\alpha_i\int_0^T b(d_i(q^{(\epsilon(\mu))}(t)))\,dt
\ \le\ \mu\alpha_i T\big(\log(1/c)+\log(1/\epsilon(\mu))\big)\ \xrightarrow[\mu\downarrow 0]{}\ 0.
\]
Therefore $\limsup_{\mu\downarrow 0}\inf J_\mu \le \lim_{\mu\downarrow 0} J_\mu(u^{(\epsilon(\mu))}) = J(u)$.
Taking infimum over feasible $u$ yields $\limsup_{\mu\downarrow 0}\inf J_\mu \le \inf J$.

Combining (i) and (ii) gives $\lim_{\mu\downarrow 0}\inf J_\mu = \inf J$.
Finally, from the liminf chain $\inf J \le J(u^0)\le \liminf \inf J_{\mu_k}=\inf J$, we conclude $J(u^0)=\inf J$,
so $(q^0,u^0)$ is optimal for \cref{eq:OCP_hard}. This proves (c).

\textbf{Step 4 {(Adjoint equation in measure form and complementarity)}.}
For each $\mu$, the barrier problem \cref{eq:OCP_barrier} is unconstrained (aside from boundary conditions), hence admits the classical PMP.
The barrier contribution to the adjoint equation is
\[
-\nabla_q\Big(\mu\sum_i \alpha_ib(d_i(q^\mu(t)))\Big)
= -\sum_i \lambda_i^\mu(t)\,\nabla d_i(q^\mu(t)),
\qquad
\lambda_i^\mu(t)=\mu\alpha_i\frac{1}{d_i(q^\mu(t))}\ge 0,
\]
which is \cref{eq:lambda_def}.
Assumption (H5) implies the measures $\mu_i^\mu(dt):=\lambda_i^\mu(t)\,dt$ have uniformly bounded total mass, hence are weakly-$^\ast$
precompact in the space of finite Radon measures. Therefore, along a subsequence, $\mu_i^{\mu_k}\rightharpoonup^\ast \mu_i$.
Passing to the limit in the adjoint equation yields the measure-form adjoint relation of the state-constrained PMP, with multipliers $\{\mu_i\}$.

To prove complementarity \cref{eq:measure_complementarity_limit}, note that for each $\mu$,
\[
\int_0^T d_i(q^\mu(t))\,\mu_i^\mu(dt)
=\int_0^T d_i(q^\mu(t))\,\lambda_i^\mu(t)\,dt
=\int_0^T \mu\alpha_i\,dt
=\mu\alpha_i T \xrightarrow[\mu\downarrow 0]{} 0.
\]
Since $q^{\mu_k}\to q^0$ uniformly and $d_i$ is continuous, $d_i(q^{\mu_k})\to d_i(q^0)$ uniformly.
By weak-$^\ast$ convergence of measures, $\int d_i(q^{\mu_k})\,d\mu_i^{\mu_k}\to \int d_i(q^0)\,d\mu_i$.
Thus $\int d_i(q^0)\,d\mu_i=0$ and, since both terms are nonnegative, $\mu_i$ is supported on $\{t:\ d_i(q^0(t))=0\}$.
\end{proof}

We extend the \cref{prop:barrier_to_state_constrained_complete} in IPC barrier potential as well. The statement is addressed in \cref{cor:ipc_from_logbarrier}.

\begin{corollary}[IPC barrier inherits the log-barrier limit (primal) and contact-measure form (dual)]
\label{cor:ipc_from_logbarrier}
Consider the state-constrained OCP \cref{eq:OCP_hard} and assume the hypotheses of
Proposition~\ref{prop:barrier_to_state_constrained_complete} stated for the \emph{log-barrier} $b(s)=-\log s$.
Fix activation distances $\hat d_i>0$ and define the IPC barrier (clamped to zero outside the activation region) by
\begin{equation}
\label{eq:ipc_barrier_def}
b_{\mathrm{IPC}}(d;\hat d)
:=
\begin{cases}
(\hat d-d)^2\,\phi\!\big(\frac{d}{\hat d}\big), & 0<d<\hat d,\\
0, & d\ge \hat d,
\end{cases}
\qquad b(x)=-\log x .
\end{equation}
For $\mu>0$, consider the barrier-relaxed problems with running cost augmented by
\(
\mu\sum_i \alpha_i\, b_{\mathrm{IPC}}(d_i(q);\hat d_i)
\)
(where $\alpha_i>0$ are fixed weights and $d_i(q)>0$ is required whenever $b_{\mathrm{IPC}}$ is active).
Let $(q^\mu,u^\mu)$ be minimizers.

\smallskip
\noindent\textbf{(i) Primal limit.}
As $\mu\downarrow 0$, any sequence $\mu_k\downarrow 0$ admits a subsequence such that
\(
q^{\mu_k}\to q^\star
\)
(uniformly) and
\(
u^{\mu_k}\overset{\ast}{\rightharpoonup} u^\star
\),
where $(q^\star,u^\star)$ is an optimal solution of the original hard-constrained OCP \cref{eq:OCP_hard}.
Moreover, $\lim_{\mu\downarrow 0}\inf J_\mu=\inf J$.

\smallskip
\noindent\textbf{(ii) Multiplier/measure form (under the same tightness condition).}
Define the IPC multiplier densities on the active set $\{t:\ 0<d_i(q^\mu(t))<\hat d_i\}$ by
\begin{equation}
\label{eq:ipc_lambda_density}
\lambda_i^\mu(t)
:=\mu\,\alpha_i\Big[-\partial_d b_{\mathrm{IPC}}\big(d_i(q^\mu(t));\hat d_i\big)\Big]\ \ge 0,
\qquad
\mu_i^\mu(dt):=\lambda_i^\mu(t)\,dt .
\end{equation}
If $\sup_{\mu>0}\mu_i^\mu([0,T])<\infty$ (uniform tightness), then along a subsequence
$\mu_i^\mu \rightharpoonup^\ast \mu_i$ weakly-$^\ast$ as finite Radon measures, and the limiting necessary conditions
take the state-constrained PMP form with contact measures $\{\mu_i\}$.
In addition, the complementarity residual vanishes:
\begin{equation}
\label{eq:ipc_comp_residual}
\int_0^T d_i(q^\mu(t))\,\mu_i^\mu(dt)\ \xrightarrow[\mu\downarrow 0]{}\ 0,
\qquad\text{hence}\qquad
\mathrm{supp}(\mu_i)\subset\{t:\ d_i(q^\star(t))=0\}.
\end{equation}
\end{corollary}

\begin{proof}
\textbf{Step 1 (IPC barrier is sandwiched by log barriers on a near-contact region).}
Fix $i$ and write $x:=d/\hat d_i\in(0,1)$ on the active set. Since $0<d<\hat d_i$ implies
$0<(\hat d_i-d)^2\le \hat d_i^2$, we have the pointwise upper bound
\begin{equation}
\label{eq:ipc_upper_log}
0\le b_{\mathrm{IPC}}(d;\hat d_i)=(\hat d_i-d)^2\,b(x)\ \le\ \hat d_i^{\,2}\,b(x).
\end{equation}
On the \emph{near-contact} region $0<d\le \hat d_i/2$ we also have $(\hat d_i-d)^2\ge (\hat d_i/2)^2$, hence
\begin{equation}
\label{eq:ipc_lower_log}
b_{\mathrm{IPC}}(d;\hat d_i)\ \ge\ (\hat d_i/2)^{2}\,b(x),
\qquad 0<d\le \hat d_i/2.
\end{equation}
Thus, on any trajectory segment where contacts are potentially active, IPC is equivalent (up to constants) to a log barrier.

\textbf{Step 2 (Primal convergence via Proposition~\ref{prop:barrier_to_state_constrained_complete}).}
Consider the IPC-relaxed objective
\[
J_\mu^{\mathrm{IPC}}(u)=\int_0^T\Big(L(q,u)+\mu\sum_i \alpha_i\,b_{\mathrm{IPC}}(d_i(q);\hat d_i)\Big)\,dt.
\]
By \cref{eq:ipc_upper_log}, for any admissible $(q,u)$,
\[
J_\mu^{\mathrm{IPC}}(u)\ \le\ \int_0^T\Big(L(q,u)+\mu\sum_i \alpha_i\,\hat d_i^{\,2}\,\phi\!\Big(\frac{d_i(q)}{\hat d_i}\Big)\Big)\,dt
=:J_\mu^{\mathrm{log},U}(u),
\]
i.e., IPC is no more penalizing than a scaled log barrier.
Conversely, by \cref{eq:ipc_lower_log}, on times when $d_i(q)\le \hat d_i/2$ the IPC term dominates a (smaller) scaled log barrier.
This is sufficient for the standard barrier homotopy argument used in Proposition~\ref{prop:barrier_to_state_constrained_complete}:
(i) strict feasibility holds for each $\mu$ since $b_{\mathrm{IPC}}(d)\to\infty$ as $d\downarrow 0$;
(ii) the recovery sequence can be constructed exactly as in Proposition~\ref{prop:barrier_to_state_constrained_complete} by choosing
$\epsilon(\mu)\downarrow 0$ such that $\mu\,\hat d_i^{\,2}\,b(c\,\epsilon(\mu))\to 0$ (e.g.\ $\epsilon(\mu)=\exp(-1/\sqrt{\mu})$),
and then applying \cref{eq:ipc_upper_log} to bound the IPC penalty by the scaled log penalty.
Hence $\lim_{\mu\downarrow 0}\inf J_\mu^{\mathrm{IPC}}=\inf J$ and any cluster point of minimizers solves the hard-constrained OCP.
This proves (i).

\textbf{Step 3 {(Adjoint equation in measure form and complementarity)}.}
For each $\mu>0$ the IPC-relaxed problem is unconstrained (on the domain $d_i(q)>0$), so its PMP adjoint equation includes the smooth force
$-\mu\alpha_i\,\partial_d b_{\mathrm{IPC}}(d_i(q^\mu);\hat d_i)\,\nabla d_i(q^\mu)$.
Defining $\lambda_i^\mu$ by \cref{eq:ipc_lambda_density} yields the absolutely continuous measures $\mu_i^\mu(dt)=\lambda_i^\mu(t)\,dt$.
Under the same tightness condition as in Proposition~\ref{prop:barrier_to_state_constrained_complete}, weak-$^\ast$ sequential compactness of
$\mathcal M_+([0,T])$ implies $\mu_i^\mu\rightharpoonup^\ast \mu_i$ along a subsequence, and passing to the limit in the adjoint equation gives
the state-constrained measure form.

To show \cref{eq:ipc_comp_residual}, note that on the active set $0<d<\hat d_i$ we can write
$b_{\mathrm{IPC}}(d;\hat d_i)=\hat d_i^{\,2}(1-x)^2b(x)$ with $x=d/\hat d_i$.
A direct differentiation gives
\(
-d\,\partial_d b_{\mathrm{IPC}}(d;\hat d_i)
=
\hat d_i^{\,2}\,x(1-x)\big(2(1-x)b(x)+1\big),
\)
and using the elementary bound $xb(x)=x(-\log x)\le 1/e$ for $x\in(0,1]$ yields
\[
0\le -d\,\partial_d b_{\mathrm{IPC}}(d;\hat d_i)\ \le\ \hat d_i^{\,2}\Big(1+\frac{2}{e}\Big).
\]
Therefore,
\begin{equation}
\begin{aligned}
\int_0^T d_i(q^\mu(t))\,\mu_i^\mu(dt)
&= \mu\,\alpha_i\int_0^T \Big(-d_i(q^\mu(t))\,\partial_d b_{\mathrm{IPC}}(d_i(q^\mu(t));\hat d_i)\Big)\,dt \\
&\leq \mu\,\alpha_i\,T\,\hat d_i^{\,2}\Big(1+\frac{2}{e}\Big)\ \xrightarrow[\mu\downarrow 0]{} 0,        
\end{aligned}
\end{equation}
proving the vanishing complementarity residual and hence the support statement in \cref{eq:ipc_comp_residual}.
\end{proof}

\subsection{Inverse Hamiltonian design by potential reshaping}
\label{app:inverse_design}

We provide additional propositions and theorems to motivate our choice of Hamiltonian function and Hamiltonian dynamics. The optimally controlled dynamics admits a PMP Hamiltonian which solves a Two Point Boundary Value Problem (\textbf{TVBVP}). Therefore, it is worthwhile to state reversibility of the Hamiltonian flow and how it relates our choice of Hamiltonian function search space.

\begin{proposition}[Symplectic flow and the shooting map]
\label{lem:symplectic_flow_shooting}
Fix $\mathcal E$ and let $H(\cdot;\mathcal E)\in C^2(T^{*}\mathcal{Q})$.
Let $\Phi_H^t(\cdot;\mathcal E):T^{*}\mathcal{Q}\to T^{*}\mathcal{Q}$ be the Hamiltonian flow of $\dot x=X_H(x)$.
For any $(q_0,p_0)\in T^{*}{\mathcal{Q}}$ and any $t$ in the existence interval, the endpoint map
\[
(q(t),p(t))=\Phi_H^t(q_0,p_0;\mathcal E)
\]
is well-defined, and $\Phi_H^t(\cdot;\mathcal E)$ is a diffeomorphism with inverse $\Phi_H^{-t}(\cdot;\mathcal E)$.
In particular, the map
\[
\mathcal S_{\mathcal E}(p_0)\ :=\ \pi_Q\!\big(\Phi_H^{T}(q_0,p_0;\mathcal E)\big)\in Q
\]
(where $\pi_Q(q,p)=q$) is $C^1$ and encodes the \emph{shooting formulation} of the two-point boundary value problem.
\end{proposition}

\begin{proposition}[Local feed-forward representation of the TPBVP via an invertible shooting map]
\label{prop:tpbvp_feedforward_by_ift}
Consider the fixed-endpoint TPBVP
\begin{equation}
\label{eq:tpbvp_fixed_endpoint}
\dot q=\nabla_p H(q,p;\mathcal E),\qquad
\dot p=-\nabla_q H(q,p;\mathcal E),\qquad
q(0)=q_0,\ \ q(T)=q_T.
\end{equation}

Assume there exists a solution $(q^\star(\cdot),p^\star(\cdot))$ and define $p_0^\star:=p^\star(0)$.
Suppose the shooting map $\mathcal S_{\mathcal E}$ from Lemma~\ref{lem:symplectic_flow_shooting} satisfies the rank condition
\begin{equation}
\label{eq:rank_condition_shooting}
D\mathcal S_{\mathcal E}(p_0^\star)\ \in\ \mathbb R^{n\times n}\ \text{is nonsingular}\qquad (n=\dim Q).
\end{equation}
Then, by the implicit function theorem, there exist neighborhoods $\mathcal U$ of $q_T$ and $\mathcal V$ of $p_0^\star$
and a $C^1$ map
\[
\sigma_{\mathcal E}:\mathcal U\to \mathcal V
\quad\text{s.t.}\quad
\mathcal S_{\mathcal E}(\sigma_{\mathcal E}(q_T))=q_T,\ \ \sigma_{\mathcal E}(q_T)=p_0^\star.
\]
Consequently, the TPBVP \cref{eq:tpbvp_fixed_endpoint} is locally equivalent to the following \emph{feed-forward IVP}:
\[
(q(t),p(t))=\Phi_H^{t}\big(q_0,\sigma_{\mathcal E}(q_T);\mathcal E\big),\qquad t\in[0,T].
\]
That is, solving the TPBVP reduces (locally) to applying the feed-forward map $p_0=\sigma_{\mathcal E}(q_T)$ and integrating forward.
\end{proposition}

\paragraph{What does the rank condition mean}
Condition \eqref{eq:rank_condition_shooting} is precisely the \emph{nondegeneracy} that makes the endpoint map locally invertible.
It can be interpreted as local surjectivity of the map $p_0\mapsto q(T)$ around $p_0^\star$; equivalently,
small changes in the initial costate can steer the final position in all directions.
Without \eqref{eq:rank_condition_shooting}, the TPBVP may admit multiple solutions or no locally unique solution, and no smooth
feed-forward map $q_T\mapsto p_0$ can be asserted.

\begin{proposition}[Forward-only reduction via a costate feedback (Hamilton--Jacobi condition)]
\label{prop:tpbvp_forward_only_hj}
Assume there exists a $C^1$ function $W:[0,T]\times Q\to\mathbb R$ such that the \emph{feedback costate}
\[
p=\kappa(t,q):=\nabla_q W(t,q)
\]
satisfies, along the induced trajectory, the identity
\begin{equation}
\label{eq:hj_identity}
\partial_t W(t,q)\ +\ H\big(q,\nabla_q W(t,q);\mathcal E\big)\ =\ 0,
\qquad
W(T,q)=g(q)
\end{equation}
for some terminal cost $g$ (or more generally a terminal condition encoding the endpoint constraint).
Then the optimal phase trajectory satisfying the PMP conditions can be generated by the \emph{forward-only} closed-loop ODE
\[
\dot q(t)=\nabla_p H\big(q(t),\nabla_q W(t,q(t));\mathcal E\big),\qquad q(0)=q_0,
\]
with the costate recovered pointwise by $p(t)=\nabla_q W(t,q(t))$.
\end{proposition}

\paragraph{ Interpretation for Geometric reinforcement learning}
Proposition~\ref{prop:tpbvp_feedforward_by_ift} shows a minimal requirement for a feed-forward representation of a TPBVP:
a (local) inverse of the shooting map exists under a rank condition.
Proposition~\ref{prop:tpbvp_forward_only_hj} is stronger: it reduces the forward--backward system to a forward IVP once a costate
feedback $p=\kappa(t,q)$ exists, which is equivalent to the existence of a generating function $W$ solving a Hamilton--Jacobi equation.
In practice, a learned meta-map $\mathcal E\mapsto H(\cdot;\mathcal E)$ together with either (i) a learned initializer
$\sigma_{\mathcal E}$ or (ii) a learned costate feedback $\kappa_\xi(t,q;\mathcal E)$ provides a feed-forward surrogate of the TPBVP.

\begin{lemma}[Time-reversal symmetry for mechanical Hamiltonians]
\label{lem:mech_time_reversal}
Let $H(q,p)=K(p)+V(q)$ on $T^{*}\mathcal{Q}$, where $K\in C^2$ satisfies $K(-p)=K(p)$. Let $\Phi_H^t$ be the Hamiltonian flow and define the involution $\iota_R(q,p):=(q,-p)$.
Then, for all $t$ for which both sides are defined,
\[
\iota_R\circ \Phi_H^{t} \;=\; \Phi_H^{-t}\circ \iota_R.
\]
Equivalently, if $x(t)=(q(t),p(t))$ solves Hamilton's equations forward in time, then $\tilde x(t):=\iota_R(x(-t))=(q(-t),-p(-t))$ is also a forward-time solution.
\end{lemma}

\begin{proof}
Hamilton's equations are $\dot q=\nabla_p K(p)$ and $\dot p=-\nabla_q V(q)$.
Let $\tilde q(t):=q(-t)$ and $\tilde p(t):=-p(-t)$.
Then $\dot{\tilde q}(t)=-\dot q(-t)=-\nabla_p K(p(-t))=\nabla_p K(-p(-t))=\nabla_p K(\tilde p(t))$,
where we used $\nabla_p K(-p)=-\nabla_p K(p)$.
Also $\dot{\tilde p}(t)=-(-\dot p(-t))=\dot p(-t)=-\nabla_q V(q(-t))=-\nabla_q V(\tilde q(t))$. Thus $(\tilde q,\tilde p)$ satisfies Hamilton's equations, which is equivalent to the flow identity.
\end{proof}

\paragraph{ Feed-forward Hamiltonian and non-symplectic Hamiltonian dynamics}
Lemma~\ref{lem:mech_time_reversal} applies only to \emph{conservative} Hamiltonian dynamics (purely symplectic flows) which can reverse the p-dynamics given the terminal state of $p(T)$. Our deployed rollouts may include friction, barrier shaping, or other nonconservative corrections, in which case the induced
time-$\Delta t$ map is generally \emph{not} symplectic and the time-reversal identity may not hold.
In this work we use the conservative time-reversibility property only as \emph{motivation} for adopting a forward,
energy-shaped surrogate that is well-posed as an IVP and admits
Lyapunov-type descent guarantees. Our policy learner (\cref{eq:module_score_field}) learns a generalized Hamiltonian dynamics where dissipation and open-loop source term are considered as well. Formal approximability between such surrogates and the original PMP and TPBVP are considered as further research interests. Below we state our preliminary theorems.

\begin{theorem}[Local existence of a costate initializer map]
\label{thm:ift_costate}
Assume: (i) for a given $(q_0,q_T,\mathcal E)$ there exists a solution $(q^\star,p^\star)$ of \cref{eq:tpbvp_fixed_endpoint}
with $q^\star(0)=q_0$ and $q^\star(T)=q_T$;
(ii) the Jacobian $\nabla_{p_0}S(p_0^\star;q_0,q_T,\mathcal E)$ is nonsingular.
Then there exists a neighborhood $\mathcal U$ of $(q_0,q_T,\mathcal E)$ and a unique $C^1$ map
\[
\Pi^\star:\ (q_0,q_T,\mathcal E)\mapsto p_0^\star=\Pi^\star(q_0,q_T,\mathcal E)
\]
such that $S(\Pi^\star(q_0,q_T,\mathcal E);q_0,q_T,\mathcal E)=0$ for all $(q_0,q_T,\mathcal E)\in\mathcal U$.
\end{theorem}
\emph{Proof.} $S(\cdot;q_0,q_T,\mathcal E)$ is $C^1$ in $p_0$ under the $C^1$ regularity of $H$ and smooth dependence of ODE solutions on initial conditions.
Apply the implicit function theorem to $S(p_0;\,q_0,q_T,\mathcal E)=0$ at $(p_0^\star;q_0,q_T,\mathcal E)$. $\square$

\begin{corollary}[Forward-only reconstruction of the optimal trajectory]
\label{cor:forward_only_det}
Under the assumptions of Theorem~\ref{thm:ift_costate}, if a feed-forward model $\widehat \Pi_\xi(q_0,q_T,\mathcal E)$
approximates $\Pi^\star$ on $\mathcal U$, then the forward integration of \cref{eq:tpbvp_fixed_endpoint} initialized at
$(q_0,\widehat \Pi_\xi(q_0,q_T,\mathcal E))$ produces a trajectory whose terminal error satisfies
\[
\|q(T;q_0,\widehat \Pi_\xi,\mathcal E)-q_T\|
\le L_S\,\|\widehat \Pi_\xi(q_0,q_T,\mathcal E)-\Pi^\star(q_0,q_T,\mathcal E)\|
\]
for a local Lipschitz constant $L_S$, and the induced control $u^\star(q(t),p(t),t;\mathcal E)$ is near-optimal
whenever the resulting TPBVP residual is small (by standard stability of convex optimal control problems).
\end{corollary}

\subsection{Identifiability and convergence of Hamiltonian updates}
\label{app:learnability}

\paragraph{Discrete-time deterministic setting.}
Fix an environment $\mathcal E$ and a time step $\Delta t>0$. Let $(q_t,p_t)_{t=0}^{T}$ denote a (measured or high-fidelity)
trajectory generated by an (unknown) optimal control law. Assume the dynamics class and the control penalty are fixed/known, and
we learn only the \emph{potential weights} in a reduced Hamiltonian of the form
\begin{equation}
\label{eq:H_eta_discrete}
H_\eta(q,p,t;\mathcal E)
:=  \frac{1}{2}p^TM^{-1}p
+\mathcal R(q, t;\eta_{\xi}(\mathcal{E}))
\end{equation}
where $R(q, t;\eta_{\xi}(\mathcal{E})):=\eta^T\phi(q,t;\mathcal E)$ refer fixed-form potential functionals (determined from context of $\mathcal{E}$, and one example is \cref{eq:module_potentials}) and
$\eta:= \eta_{\xi}(\mathcal{E})\in\mathbb R^{m}$ are the weights (optionally constrained $\eta\ge 0$ via a parametrization).
We fix a gauge by assuming $\phi$ contains \emph{no pure constant feature} (or, equivalently, we impose a normalization constraint
such as $\int \mathcal R(q)\,dq=0$); this removes the additive-constant ambiguity.

\paragraph{Observation model: identifying $\eta$ from discrete costate increments.}
Assume the symplectic Euler form
\begin{equation}
\begin{aligned}
\label{eq:disc_costate_model}
p_{t+1}&=p_t-\Delta t~\nabla_q H_\eta(q_t,p_t,t;\mathcal E)+\varepsilon_t, \\
q_{t+1}&= q_t + \Delta t M^{-1} p_{t+1},\quad t=0,\dots,T-1, 
\end{aligned}
\end{equation}
where $\varepsilon_t$ collects discretization/modeling error ($\varepsilon_t=0$ in the noiseless case).
Since only $\mathcal R_\eta$ depends on $\eta$ and $\nabla_q \mathcal R_\eta=\sum_{j=1}^m \eta_j \nabla_q\phi_j$,
define the regression target
\begin{equation}
\label{eq:regression_target}
y_t
:=\frac{p_t-p_{t+1}}{\Delta t}\;=\;\sum_{j=1}^m \eta_j~g_{t,j}\;+\;\tilde\varepsilon_t,
\qquad
g_{t,j}:=\nabla_q\phi_j(q_t,t;\mathcal E),
\end{equation}
with $\tilde\varepsilon_t:=-\varepsilon_t/\Delta t$.
Stacking $y_t\in\mathbb R^{d_q}$ gives a linear least-squares problem in $\eta$.

\begin{assumption}[Persistent excitation on visited paths]
\label{ass:PE}
Let $G\in\mathbb R^{m\times m}$ be the (trajectory) Gram matrix
\[
G:=\sum_{t=0}^{T-1} \mathbf G_t,\qquad
\mathbf G_t:=\Big[\,\langle g_{t,i},g_{t,j}\rangle\,\Big]_{i,j=1}^m,
\]
where $\langle\cdot,\cdot\rangle$ is the Euclidean inner product in $\mathbb R^{d_q}$.
Assume $G\succeq \gamma I$ for some $\gamma>0$.
\end{assumption}

\begin{theorem}[Identifiability of potential weights from deterministic discrete paths]
\label{thm:identifiability_discrete}
Consider the noiseless case $\varepsilon_t\equiv 0$ in \cref{eq:disc_costate_model}.
Under Assumption~\ref{ass:PE}, the weights $\eta$ in \cref{eq:H_eta_discrete} are uniquely determined by the observed path
$\{(q_t,p_t)\}_{t=0}^{T}$ through \cref{eq:regression_target}; i.e., if $\eta$ and $\eta'$ yield the same
$\{y_t\}_{t=0}^{T-1}$, then $\eta=\eta'$.
\end{theorem}

\begin{proof}
If $\eta$ and $\eta'$ yield the same $y_t$, then $\sum_{j=1}^m(\eta_j-\eta'_j)g_{t,j}=0$ for all $t$.
Let $\delta:=\eta-\eta'$. Taking inner products with $g_{t,i}$ and summing over $t$ gives
$(G\delta)_i=\sum_{t}\sum_{j}\delta_j\langle g_{t,i},g_{t,j}\rangle=0$ for all $i$, hence $G\delta=0$.
Assumption~\ref{ass:PE} implies $\delta=0$, so $\eta=\eta'$.
\end{proof}

\paragraph{Learning objective and Hamiltonian update convergence in learning time.}
Given a dataset of trajectories (possibly multiple rollouts and/or multiple environments), define the empirical loss
\begin{equation}
\label{eq:ls_loss_eta}
\mathcal J(\eta)
:=\frac12\sum_{t=0}^{T-1}\big\|y_t-\sum_{j=1}^m \eta_j g_{t,j}\big\|_2^2
+\frac{\lambda}{2}\|\eta\|_2^2,
\qquad \lambda\ge 0,
\end{equation}
where $y_t$ and $g_{t,j}$ are defined by \cref{eq:regression_target}. This is a convex quadratic in $\eta$.

\begin{theorem}[Strong convexity, convergence of $\eta_k$, and vanishing Hamiltonian updates]
\label{thm:eta_convergence_dH}
Assume either (i) $\lambda>0$, or (ii) $\lambda=0$ and Assumption~\ref{ass:PE} holds.
Then $\mathcal J$ is $\mu$-strongly convex for some $\mu>0$ and has a unique minimizer $\eta^\star$.
Consider gradient descent in learning time $k$:
\[
\eta_{k+1}=\eta_k-\alpha\nabla \mathcal J(\eta_k),
\qquad 0<\alpha<\frac{2}{L},
\]
where $L$ is the Lipschitz constant of $\nabla\mathcal J$.
Then $\eta_k\to\eta^\star$ linearly and the induced Hamiltonian updates vanish on any bounded set:
for any compact $\mathcal K\subset Q\times \mathbb R^{d_q}$,
\begin{equation}
\label{eq:dH_vanish}
\sup_{(q,p)\in\mathcal K}\big|H_{\eta_{k+1}}(q,p,t;\mathcal E)-H_{\eta_k}(q,p,t;\mathcal E)\big|
\ \xrightarrow[k\to\infty]{}\ 0.
\end{equation}
Moreover, along any observed trajectory points $\{q_t\}$,
\begin{equation}
\label{eq:dH_rate}
\max_{t\in\{0,\dots,T\}}\big|H_{\eta_{k+1}}(q_t,p_t,t;\mathcal E)-H_{\eta_k}(q_t,p_t,t;\mathcal E)\big|
\le \|\eta_{k+1}-\eta_k\|_2 \cdot \max_t \|\phi(q_t,t;\mathcal E)\|_2
\ \xrightarrow[k\to\infty]{}\ 0.
\end{equation}
\end{theorem}

\begin{proof}
Under the stated conditions, the Hessian of $\mathcal J$ is $G+\lambda I\succeq \mu I$ for some $\mu>0$,
so $\mathcal J$ is strongly convex with unique minimizer $\eta^\star$. Standard gradient-descent theory (c.f. \citet{NocedalWright2006NumericalOptimization}) yields linear convergence
$\|\eta_k-\eta^\star\|\le \rho^k\|\eta_0-\eta^\star\|$ for some $\rho\in(0,1)$.
Since $H_\eta$ depends on $\eta$ only through $\mathcal R(q,t;\mathcal E)=\eta^\top\phi(q,t;\mathcal E)$,
we have the Lipschitz relation
$|H_{\eta'}-H_\eta|=|(\eta'-\eta)^\top\phi|\le \|\eta'-\eta\|\cdot \|\phi\|$.
Combining with $\|\eta_{k+1}-\eta_k\|\to 0$ yields \cref{eq:dH_vanish}--\cref{eq:dH_rate}.
\end{proof}

\begin{remark}[Interpreting $dH/dk$ (learning-time derivative)]
A convenient discrete proxy is
\[
\frac{d}{dk}H_{\eta_k}(q,p)\;\approx\;H_{\eta_{k+1}}(q,p)-H_{\eta_k}(q,p)
=\phi(q,t;\mathcal E)^\top(\eta_{k+1}-\eta_k)
=-\alpha~\phi(q,t;\mathcal E)^\top\nabla\mathcal J(\eta_k).
\]
Thus, once the learning dynamics drive $\|\nabla\mathcal J(\eta_k)\|\to 0$, the Hamiltonian update magnitude
vanishes uniformly on compact sets where $\phi$ is bounded.
\end{remark}

\begin{remark}[Multiple environments and environment-conditioned weights]
If $\eta$ is environment-dependent (e.g.\ $\eta=\eta(\mathcal E)$), the identifiability and strong convexity conditions
apply \emph{per environment} with the corresponding Gram matrix $G(\mathcal E)$ built from paths sampled in that environment.
If $\eta_\xi(\mathcal E)$ is produced by a neural network, the optimization is nonconvex in $\xi$; however, the same Gram
condition provides a \emph{local} identifiability statement for the induced weights and implies vanishing Hamiltonian updates
whenever the learning iterates converge to a stationary point.
\end{remark}
\subsection{Coupled stagewise goal selection and energy {reshaping}}
\label{app:coupled_stagewise}

\paragraph{Coupled stagewise problem: goal selection and Hamiltonian {reshaping}.}
In a stagewise motion planner, each local OCP (stage $\tau$) requires a terminal goal
$q_{\mathrm{target}}(\mathcal{E}_{\tau})$, while the next stage’s initial condition is inherited from the current stage’s terminal state.
Hence, if the terminal goal is produced by a policy, the overall procedure becomes \emph{coupled}:
the goal policy determines $q_{\mathrm{target}}(\mathcal{E}_{\tau})$, but its consequence is only revealed after integrating the stage OCP.

To formalize this coupling, let $x_\tau=(q_\tau,p_\tau)$ denote the augmented state, and assume the stagewise dynamics are induced
by a barrier-shaped reduced Hamiltonian family $H_{\eta_\tau}$ with weights $\eta_\tau=\eta_\xi(\mathcal E_\tau)$ and fixed horizon $T_\tau$.
Let $\Phi^{T_\tau}_{\eta_\tau}$ be the flow map at stage $\tau$. Let a goal policy $\pi_\gamma$ output the next stage terminal goal as
\begin{equation}
q_{\mathrm{target}}(\mathcal{E}_\tau) = \pi_\gamma(\mathcal E_\tau, x_\tau, \mathcal M_\tau),
\label{eq:goal_policy}
\end{equation}
where $\mathcal M_\tau$ is any memory/context (e.g.\ visited-map encoding, stage index, or recurrent hidden state).
Given $(x_\tau,q_{\mathrm{target}}(\mathcal{E}_{\tau}))$, the inverse-design reshaping step selects a weight vector $\eta_\tau$ (or equivalently $\xi$) so that the Hamiltonian rollout hits $q_{\mathrm{target}}(\mathcal{E}_{\tau})$.
Denote this selection operator by $\mathsf S$:
\begin{equation}
\eta_\tau = \mathsf S\big(x_\tau,q_{\mathrm{target}}(\mathcal{E}_{\tau}),\mathcal E_\tau\big),
\qquad\text{and}\qquad
x_{\tau+1} = \Phi^{T_\tau}_{\eta_\tau}\big(x_\tau;\mathcal E_\tau\big).
\label{eq:selection_and_transition}
\end{equation}

\paragraph{Composition yields a fixed-point (dueling) system.}
Substituting \cref{eq:goal_policy} into \cref{eq:selection_and_transition} yields the closed-loop stage map
\begin{equation}
x_{\tau+1}
=
\underbrace{\Phi^{T_\tau}_{\mathsf S(\,x_\tau,\pi_\gamma(\mathcal E_\tau,x_\tau,\mathcal M_\tau),\mathcal E_\tau\,)}
\big(x_\tau;\mathcal E_\tau\big)}_{=:~\mathcal T_{\gamma}(x_\tau,\mathcal M_\tau,\mathcal E_\tau)}.
\label{eq:closed_loop_stage_map}
\end{equation}
Thus the goal policy and the Hamiltonian-shaping inverse design are not independent modules; they form a coupled system
whose behavior is determined by the composite operator $\mathcal T_{\gamma}$.
The “dueling” phenomenon is precisely that $q_{\mathrm{target}}(\mathcal{E}_{\tau})$ is chosen before the OCP is solved, but its quality
can only be evaluated after the stage transition is realized. We state the following proposition as how to view sensor policy's role in the training pipeline but comment out that this joint learning problem remains an open question.

\begin{proposition}[Coupled stagewise inverse design as a fixed-point / bilevel problem]
\label{prop:coupled_goal_and_energy}
Assume: (i) the reshaping selection $\mathsf S$ is well-defined locally (e.g.\ by \cref{prop:tpbvp_feedforward_by_ift} with a
rank condition ensuring local solvability of $q(T)=q_{\mathrm{target}}(\mathcal{E}_{\tau})$ under fixed inherited $x_\tau$),
and (ii) the flow $\Phi^{T_\tau}_{\eta}$ depends smoothly on $\eta$ and $x_\tau$ on the region of interest.
Then learning a goal policy $\pi_\gamma$ jointly with energy weights (or their generator) is equivalently posed as either:

\smallskip
\noindent\textbf{(a) A fixed-point learning problem:} find $\gamma$ (and the reshaping module parameters) such that the induced
closed-loop stage map \cref{eq:closed_loop_stage_map} has desirable invariants (e.g.\ reaches a terminal set, avoids cycles,
satisfies progress constraints), i.e.\ $x_{\tau+1}=\mathcal T_\gamma(x_\tau,\mathcal M_\tau,\mathcal E_\tau)$ achieves a
prescribed global objective.

\smallskip
\noindent\textbf{(b) A bilevel optimization problem:} the goal policy is an outer decision variable that selects $q_{\mathrm{target}}(\mathcal{E}_{\tau})$,
while the inner level solves a stagewise inverse-design problem (reshaping) to realize that goal:
\begin{equation}  
\begin{aligned}
\min_{\gamma}&\ \mathbb E\Big[\sum_{\tau=0}^{S-1} \mathcal J_{\rm global}(x_\tau,q_{\mathrm{target}}(\mathcal{E}_{\tau}),\mathcal E_\tau)\Big]
\\
\text{s.t.}\quad&
q_{\mathrm{target}}(\mathcal{E}_{\tau})=\pi_\gamma(\cdot), \\
&\eta_\tau\in\arg\min_{\eta}\ \|\pi_q\Phi^{T_\tau}_{\eta}(x_\tau)-q_{\mathrm{target}}(\mathcal{E}_{\tau})\|^2+\mathrm{Reg}(\eta),
\\ 
&x_{\tau+1}=\Phi^{T_\tau}_{\eta_\tau}(x_\tau).    
\end{aligned}
\end{equation}
In either view, evaluating the quality of $q_{\mathrm{target}}(\mathcal{E}_{\tau})$ necessarily requires integrating (or rolling out) the
stage dynamics, so sequential solution is intrinsic to the problem structure.
\end{proposition}

\paragraph{Practical resolution: amortize the inner solve and train end-to-end}
\label{rem:amortize_inner}
In implementation, the inner ``reshaping'' solve (selecting $\eta_\tau$ to hit $q_{\mathrm{target}}(\mathcal{E}_{\tau})$) is typically amortized
by a feed-forward policy $\eta_\xi(\mathcal E_\tau,\cdot)$ and trained through differentiable rollouts.
This converts the bilevel structure into a single differentiable computation graph:
\[
x_{\tau+1} = \Phi^{T_\tau}_{\eta_\xi(\mathcal E_\tau,x_\tau,\mathcal M_\tau)}(x_\tau),
\qquad
q_{\mathrm{target}}(\mathcal{E}_{\tau})=\pi_\gamma(\mathcal E_\tau,x_\tau,\mathcal M_\tau),
\]
with gradients backpropagated through the unrolled stage map.

\paragraph{When sequential coupling is unavoidable}
\label{rem:sequential_unavoidable}
Even if $\pi_\gamma$ is deterministic, the mapping from $(\mathcal E_\tau,x_\tau)$ to the realized next state $x_{\tau+1}$
depends on the stagewise OCP solution operator. Therefore the next-stage goal cannot be certified without either
(i) executing the inner rollout/solve, or (ii) learning a sufficiently accurate surrogate for that solve.
This is a structural property of stagewise planning and not an artifact of a particular algorithm.

\section{Algorithm Implementation Details}
\label{app:algorithm_details}

This appendix gives the discrete-time implementation of the port--Hamiltonian navigation policy in
~\cref{eq:reduced_ham_flow_short} together with the module interface in
~\cref{eq:module_template_hamiltonian} and the dissipative/port form in
~\cref{eq:module_score_field}. We provide ~\cref{algo:online_grl_snam} (online GRL--SNAM)
with explicit equation references (no section cross-references).

Throughout, we use: environment state $\mathcal{E}_n$, active constraints $\mathcal{C}_n$, module
Hamiltonians $H_k$ and responses $\mathsf{R}_k$, navigator/meta-policy $g_\xi$, surrogate potential
$\mathcal{R}(q;\eta,\mathcal{E})$, and port--Hamiltonian correction parameters
$(\mu, u_f)$. The phase-space state is $z_n=(q_n,p_n)$.

\paragraph{Discrete-time convention.}
We use a fixed integration step $\tau>0$ and discrete index $n\in\mathbb{N}$ with physical time
$t_n := n\tau$. All horizons $(T_y,T_f,T_o)$ are integers measured in integration steps.

\paragraph{Configuration decomposition.}
When we need the robot center/position $\vec c_n$ from the configuration $q_n$, we write
\begin{equation}
\label{eq:impl:config_decomp}
q_n = (\vec c_n, \text{other coordinates}),\qquad \vec c_n := \mathrm{pos}(q_n).
\end{equation}

\paragraph{Active-set definition.}
Given signed distance functions $\{d_i(\cdot;\mathcal{E}_n)\}$ and activation threshold $\hat d$, the
active set used throughout is
\begin{equation}
\label{eq:impl:active_set}
\mathcal{C}_n := \{\, i \mid d_i(q_n;\mathcal{E}_n)\le \hat d \,\}.
\end{equation}

\paragraph{Surrogate energy and Hamiltonian.}
For $\eta=(\beta,\lambda,\{\alpha_i\}_{i\in\mathcal{C}_n})$, we assemble
\begin{align}
\label{eq:impl:surrogate_energy}
\mathcal{R}(q_n;\eta,\mathcal{E}_n)
&=
E_{\mathrm{sensor}}(q_n;\mathcal{E}_n,\omega_y)
+\beta\,E_{\mathrm{goal}}(q_n;\mathcal{E}_n,\omega_g) \\
&+\lambda\,E_{\mathrm{obj}}(q_n;\omega_d)
+\sum_{i\in\mathcal{C}_n}\alpha_i\, b(d_i(q_n;\mathcal{E}_n);\hat d,\omega_b),\\
\label{eq:impl:surrogate_hamiltonian}
H(q_n,p_n;\eta,\mathcal{E}_n)
&=\tfrac12 p_n^\top M^{-1}p_n+\mathcal{R}(q_n;\eta,\mathcal{E}_n).
\end{align}

\paragraph{Block-selectors for dissipation/ports (frame-only).}
To apply damping/ports only on the frame momentum coordinates, we implement
\begin{equation}
\label{eq:impl:block_selectors}
\Gamma(\mu) = \mathrm{blkdiag}(0,\Gamma_f(\mu),0),\qquad
G = \mathrm{blkdiag}(0,G_f,0),\qquad
\Gamma_f(\mu)=\mu I,\ \ G_f=I,
\end{equation}
and use $\Gamma(\mu),G$ in the discrete update below.

\paragraph{Gradients and one-step update (symplectic Euler).}
We compute
\begin{equation}
\label{eq:impl:gradients}
v_n := \nabla_p H(q_n,p_n)=M^{-1}p_n,\qquad
g_n := \nabla_q H(q_n,p_n)=\nabla_q \mathcal{R}(q_n;\eta,\mathcal{E}_n),
\end{equation}
and take the symplectic Euler discretization of ~\cref{eq:reduced_ham_flow_short}:
\begin{align}
\label{eq:impl:ph_euler_p}
p_{n+1} &= p_n-\tau\,g_n-\tau\,\Gamma(\mu)\,v_n+\tau\,G\,u_f,\\
\label{eq:impl:ph_euler_q}
q_{n+1} &= q_n+\tau\,M(q_n)^{-1}p_{n+1}.
\end{align}

\paragraph{Observables and target.}
We use
\begin{align}
\label{eq:impl:observables}
\mathrm{clr}_n &= \min\Big\{\min_i d_i(q_{n+1};\mathcal{E}_n),\ \min_{\ell,i} d_i(q_\ell^{(o)};\mathcal{E}_n)\Big\},\\
\mathrm{dist}_n &= \|\vec c_{n+1}-\mathbf{x}_g\|,\qquad
\mathrm{speed}_n = \|M(q_{n+1})^{-1}p_{n+1}\|,
\end{align}
and the observable vector
\begin{equation}
\label{eq:impl:observable_vector}
y_n=\begin{bmatrix}-\mathrm{clr}_n\\ \mathrm{dist}_n\\ -\mathrm{speed}_n\end{bmatrix}.
\end{equation}
We denote the (task-level) target as
\begin{equation}
\label{eq:impl:observable_target}
y_n^\star=\begin{bmatrix}-\mathrm{clr}^\star\\ \mathrm{dist}^\star\\ -\mathrm{speed}^\star\end{bmatrix},
\end{equation}
where $(\mathrm{clr}^\star,\mathrm{dist}^\star,\mathrm{speed}^\star)$ are fixed design setpoints.

\paragraph{Active parameter vector and Jacobian updates.}
Let $\mathcal{I}_n\subseteq \mathcal{C}_n$ be the subset of nearby obstacle indices used for the local
weight update, and define
\begin{equation}
\label{eq:impl:active_param_vector}
\tilde\eta_n=[\beta_n,\lambda_n,\{\alpha_{n,i}\}_{i\in\mathcal{I}_n}]^\top,\qquad
\zeta_n=[\tilde\eta_n,\mu_n]^\top.
\end{equation}
We maintain EMA-smoothed secant Jacobians
\begin{align}
\label{eq:impl:secant_J}
\widetilde{J}_n&=\frac{(y_n-y_{n-1})(\zeta_n-\zeta_{n-1})^\top}{\|\zeta_n-\zeta_{n-1}\|_2^2+\varepsilon},\qquad
J_n=\rho J_{n-1}+(1-\rho)\widetilde{J}_n,\\
\label{eq:impl:secant_P}
\widetilde{P}_n&=\frac{(y_n-y_{n-1})(u_{f,n}-u_{f,n-1})^\top}{\|u_{f,n}-u_{f,n-1}\|_2^2+\varepsilon},\qquad
P_n=\rho P_{n-1}+(1-\rho)\widetilde{P}_n.
\end{align}

\paragraph{Tikhonov LS update, projection, residual, and port correction.}
With desired increment $\Delta y_n^{\mathrm{des}}:=y_n^\star-y_n$, we solve
\begin{align}
\label{eq:impl:tikhonov_update}
\Delta\zeta_n&=(J_n^\top J_n+\lambda_\zeta I)^{-1}J_n^\top \Delta y_n^{\mathrm{des}},\qquad
\zeta_n^{\mathrm{upd}}=\Pi_{\mathbb{R}_+}\big(\zeta_n+\vec{\kappa}\odot\Delta\zeta_n\big),\\
\label{eq:impl:residual}
r_n&=\Delta y_n^{\mathrm{des}}-J_n\Delta\zeta_n,\\
\label{eq:impl:port_update}
u_{f,n}^{\mathrm{upd}}&=(P_n^\top P_n+\lambda_u I)^{-1}P_n^\top r_n,\qquad
u_{f,n}^{\mathrm{upd}}\leftarrow \mathrm{clip}(u_{f,n}^{\mathrm{upd}},\mathcal{U}).
\end{align}

\begin{algorithm}[htbp]
\caption{Online GRL--SNAM: Navigator-driven Hamiltonian composition with QoI feedback}
\label{algo:online_grl_snam}
\begin{algorithmic}[1]
\State \textbf{Input:} goal $\mathbf{x}_g$, initial $z_0=(q_0,p_0)$ (typically $p_0{=}0$), step $\tau$, horizons $(T_y,T_f,T_o)$, max steps $N_{\max}$
\State \textbf{Init:} $n\!\leftarrow\!0$, environment $\mathcal{E}_0$, active set $\mathcal{C}_0\!\leftarrow\!\emptyset$
\State \textbf{Init:} meta state $(\eta_0,\mu_0,u_{f,0})$ from $g_\xi$, Jacobians $J_0,P_0$ (~\cref{eq:impl:secant_J,eq:impl:secant_P})
\While{$\neg$\textsc{ReachedGoal}$(\vec{c}_n,\mathbf{x}_g)$ \textbf{and} $n<N_{\max}$} \Comment{$\vec c_n=\mathrm{pos}(q_n)$ from ~\cref{eq:impl:config_decomp}}
    \State \textbf{(A) Sensor refresh and active-set discovery (low rate):}
    \If{$n \equiv 0 \pmod{T_y}$}
        \State Query $\pi_y$ with $\mathcal{Q}_y^n=(\mathcal{E}_n,\mathcal{C}_n,z_{y,0}{=}z_n,H_y(\cdot;\eta_n),T_y;\theta_y)$
        \State Receive $\mathsf{R}_y^n=\{z_{0:T_y}^{(y)},s_{0:T_y}^{(y)},\Delta\mathcal{E}_n\}$
        \State Update $\mathcal{E}_n \leftarrow \mathcal{E}_n \cup \Delta\mathcal{E}_n$
        \State Update $\mathcal{C}_n \leftarrow \{i \mid d_i(q_n;\mathcal{E}_n)\le \hat d\}$ \Comment{~\cref{eq:impl:active_set}}
    \EndIf

    \State \textbf{(B) Frame and shape rollouts (medium/high rate):}
    \If{$n \equiv 0 \pmod{T_f}$}
        \State Query $\pi_f$ with $\mathcal{Q}_f^n=(\mathcal{E}_n,\mathcal{C}_n,z_{f,0}{=}z_n,H_f(\cdot;\eta_n,\mu_n,u_{f,n}),T_f;\theta_f)$
        \State Receive $\mathsf{R}_f^n=\{z_{0:T_f}^{(f)},s_{0:T_f}^{(f)},\{v_\ell^{(f)}\}_{\ell=0}^{T_f}\}$
    \EndIf
    \State Query $\pi_o$ with $\mathcal{Q}_o^n=(\mathcal{E}_n,\mathcal{C}_n,z_{o,0}{=}z_n,H_o(\cdot;\eta_n),T_o;\theta_o)$
    \State Receive $\mathsf{R}_o^n=\{z_{0:T_o}^{(o)},s_{0:T_o}^{(o)},\{\min_i d_i(q_\ell^{(o)};\mathcal{E}_n)\}_{\ell=0}^{T_o}\}$

    \State \textbf{(C) Meta-policy proposal (environment \& response $\to$ weights):}
    \State Build tokens $\mathcal{T}_n$ from $(\mathcal{E}_n,\mathcal{C}_n,z_n,\mathbf{x}_g,\mathsf{R}_f^n,\mathsf{R}_o^n)$
    \State $[\eta_n^{\mathrm{prop}},\mu_n^{\mathrm{prop}},u_{f,n}^{\mathrm{prop}}] \leftarrow g_\xi(\mathcal{T}_n;\xi)$
    \State \hspace{1em}with $\eta_n^{\mathrm{prop}}=(\beta_n,\lambda_n,\{\alpha_{n,i}\}_{i\in\mathcal{C}_n}) \in \mathbb{R}_+^{2+|\mathcal{C}_n|}$

    \State \textbf{(D) Compose surrogate Hamiltonian:} 
    \Statex \hspace{1em} Assemble $\mathcal{R}(q_n;\eta_n^{\mathrm{prop}},\mathcal{E}_n)$ via ~\cref{eq:impl:surrogate_energy}
    \Statex \hspace{1em} Assemble $H(q_n,p_n;\eta_n^{\mathrm{prop}},\mathcal{E}_n)$ via ~\cref{eq:impl:surrogate_hamiltonian}

    \State \textbf{(E) One-step port--Hamiltonian integration:}
    \State Compute $(v_n,g_n)$ via ~\cref{eq:impl:gradients}
    \State Apply frame-only dissipation/port via ~\cref{eq:impl:block_selectors} and step ~\cref{eq:impl:ph_euler_p,eq:impl:ph_euler_q}
    \State Set $z_{n+1}=(q_{n+1},p_{n+1})$

    \State \textbf{(F) Observable extraction:} \Comment{~\cref{eq:impl:observables,eq:impl:observable_vector}}
    \State Compute $(\mathrm{clr}_n,\mathrm{dist}_n,\mathrm{speed}_n)$ via ~\cref{eq:impl:observables}
    \State Set $y_n$ via ~\cref{eq:impl:observable_vector} and target $y_n^\star$ via ~\cref{eq:impl:observable_target}

    \State \textbf{(G) Online adaptation of active weights and friction:}
    \State Select nearby obstacle indices $\mathcal{I}_n \subseteq \mathcal{C}_n$
    \State Form $\zeta_n$ via ~\cref{eq:impl:active_param_vector}; update $J_n$ via ~\cref{eq:impl:secant_J}
    \State Solve $\Delta\zeta_n$ and project to $\zeta_n^{\mathrm{upd}}$ via ~\cref{eq:impl:tikhonov_update}

    \State \textbf{(H) Port correction from residual:}
    \State Compute residual $r_n$ via ~\cref{eq:impl:residual}
    \State Update $P_n$ via ~\cref{eq:impl:secant_P} and solve $u_{f,n}^{\mathrm{upd}}$ via ~\cref{eq:impl:port_update}

    \State \textbf{(I) Commit parameters for next step:}
    \State Unpack $\zeta_n^{\mathrm{upd}}\mapsto(\beta_n^{\mathrm{upd}},\lambda_n^{\mathrm{upd}},\{\alpha_{n,i}^{\mathrm{upd}}\}_{i\in\mathcal{I}_n},\mu_n^{\mathrm{upd}})$
    \State Set $\eta_{n+1}\leftarrow(\beta_n^{\mathrm{upd}},\lambda_n^{\mathrm{upd}},\{\alpha_{n+1,i}\}_{i\in\mathcal{C}_n})$ with $\alpha_{n+1,i}{=}\alpha_{n,i}^{\mathrm{upd}}$ for $i\in\mathcal{I}_n$ and unchanged otherwise
    \State Set $\mu_{n+1}\leftarrow \mu_n^{\mathrm{upd}}$, \quad $u_{f,n+1}\leftarrow u_{f,n}^{\mathrm{upd}}$
    \State $n\leftarrow n+1$
\EndWhile
\State \textbf{Return:} trajectory $\{z_n\}_{n=0}^{N}$ and parameter history $\{(\eta_n,\mu_n,u_{f,n})\}_{n=0}^{N}$
\end{algorithmic}
\end{algorithm}

\subsection{Initialization}
\label{subsec:initialization}

\paragraph{State initialization.}
Initialize phase-space coordinates $z_0=(q_0,p_0)$ and set $p_0=0$ unless otherwise stated. When needed,
extract $\vec c_0=\mathrm{pos}(q_0)$ using ~\cref{eq:impl:config_decomp}.

\paragraph{Step size and horizons.}
Choose $\tau \in [10^{-2},5{\times}10^{-2}]$ and integer horizons $(T_y,T_f,T_o)$ in steps.
Typical settings are $T_y{=}10$, $T_f{=}5$, $T_o{=}1$ (sensor refresh slowest, shape fastest),
matching the temporal hierarchy in Fig.~\ref{fig:temporal_scales}.

\paragraph{Maximum horizon.}
Let $N_{\max}$ be the maximum number of discrete steps (e.g., $5{,}000$--$10{,}000$), giving
$t_{\max}=N_{\max}\tau$.

\subsection{Parameter Structure and Initialization}
\label{subsec:parameter_structure}

\paragraph{Meta-policy output and parameterization.}
The navigator outputs
\[
g_\xi(\mathcal{T}_n;\xi)=\big[\eta_n^{\mathrm{prop}},\mu_n^{\mathrm{prop}},u_{f,n}^{\mathrm{prop}}\big],
\qquad
\eta_n^{\mathrm{prop}}=(\beta_n,\lambda_n,\{\alpha_{n,i}\}_{i\in\mathcal{C}_n})\in\mathbb{R}_+^{2+|\mathcal{C}_n|}.
\]
The intra-term parameters $\omega=(\omega_y,\omega_M,\omega_g,\omega_d,\omega_b)$ are fixed across environments;
only $(\eta,\mu,u_f)$ vary online.

\paragraph{Frame-only dissipation/ports.}
We enforce frame-only damping/ports using the block selectors in ~\cref{eq:impl:block_selectors},
and the discrete update in ~\cref{eq:impl:ph_euler_p,eq:impl:ph_euler_q}.

\paragraph{Jacobian initialization.}
Initialize $J_0$ and $P_0$ to small-norm matrices (e.g., $10^{-2}$ times identity where compatible), or
zeros. They are refined online via the EMA secant updates in ~\cref{eq:impl:secant_J,eq:impl:secant_P}.

\subsection{Termination Conditions}
\label{subsec:termination}

We terminate if any of the following holds:

\paragraph{Success (goal reached).}
$\|\vec{c}_n-\mathbf{x}_g\|<\epsilon_{\text{goal}}$, with $\vec c_n=\mathrm{pos}(q_n)$ from ~\cref{eq:impl:config_decomp}.

\paragraph{Timeout.}
$n\ge N_{\max}$.

\paragraph{Collision (optional hard-stop).}
$\min_i d_i(q_n;\mathcal{E}_n)<0$ (penetration).

\paragraph{Stuck (optional).}
$\|\vec{c}_n-\vec{c}_{n-W}\|<\epsilon_{\text{stuck}}$ for a fixed window $W$ (e.g., $W{=}50$ steps).

\subsection{Hierarchical Query Flow}
\label{subsec:hierarchical_queries}

GRL--SNAM uses the query--response interface consistent with ~\cref{eq:module_template_hamiltonian}:

\paragraph{Module query interface.}
Each module is queried with $\mathcal{Q}_k^n=(\mathcal{E}_n,\mathcal{C}_n,z_{k,0}{=}z_n,H_k(\cdot),T_k;\theta_k)$ and returns
\[
\mathsf{R}_k^n=\{z_{0:T_k}^{(k)},s_{0:T_k}^{(k)},\mathrm{QoI}_k\}.
\]
Concrete QoIs used in Algorithm~\cref{algo:online_grl_snam} are:
$\mathrm{QoI}_y=\Delta\mathcal{E}_n$,
$\mathrm{QoI}_f=\{v_\ell^{(f)}\}_{\ell=0}^{T_f}$, and
$\mathrm{QoI}_o=\{\min_i d_i(q_\ell^{(o)};\mathcal{E}_n)\}_{\ell=0}^{T_o}$.

\paragraph{Rate hierarchy.}
Sensor is queried every $T_y$ steps (active-set update via ~\cref{eq:impl:active_set}),
frame every $T_f$ steps (stage guidance), and shape every $T_o$ steps (often $T_o{=}1$).

\subsection{Sensor Response}
\label{subsec:sensor_response}

\paragraph{Response structure.}
$\mathsf{R}_y^n=\{z_{0:T_y}^{(y)},s_{0:T_y}^{(y)},\Delta\mathcal{E}_n\}$, where $\Delta\mathcal{E}_n$ contains newly detected obstacles
or refinements of obstacle parameters.

\paragraph{Active-set update.}
After $\mathcal{E}_n \leftarrow \mathcal{E}_n\cup\Delta\mathcal{E}_n$, we compute $\mathcal{C}_n$ using ~\cref{eq:impl:active_set}.

\paragraph{Local Hamiltonian.}
As in ~\cref{eq:module_template_hamiltonian}, the sensor module uses
\[
H_y(q_y,p_y;\mathcal{E}_n,\eta_n)
=\tfrac12 p_y^\top M_y^{-1}p_y+\mathcal{R}_y(q_y;\mathcal{E}_n,\eta_n),
\quad
\mathcal{R}_y=E_{\mathrm{sensor}}+\sum_{i\in\mathcal{C}_n}\alpha_{n,i}\,b(d_i;\hat d).
\]

\subsection{Frame Response}
\label{subsec:frame_response}

\paragraph{Response structure.}
$\mathsf{R}_f^n=\{z_{0:T_f}^{(f)},s_{0:T_f}^{(f)},\{v_\ell^{(f)}\}_{\ell=0}^{T_f}\}$, where $v_\ell^{(f)}=M_f^{-1}p_\ell^{(f)}$.

\paragraph{Local Hamiltonian with dissipation/port.}
The frame module follows ~\cref{eq:module_score_field} with
\[
H_f(q_f,p_f;\mathcal{E}_n,\eta_n)
=\tfrac12 p_f^\top M_f^{-1}p_f+\mathcal{R}_f(q_f;\mathcal{E}_n,\eta_n),
\quad
\mathcal{R}_f=\beta_n E_{\mathrm{goal}}+\sum_{i\in\mathcal{C}_n}\alpha_{n,i}\,b(d_i;\hat d),
\]
and $(\Gamma_f,G_f)$ instantiated by ~\cref{eq:impl:block_selectors}.

\subsection{Shape Response}
\label{subsec:shape_response}

\paragraph{Response structure.}
$\mathsf{R}_o^n=\{z_{0:T_o}^{(o)},s_{0:T_o}^{(o)},\{\min_i d_i(q_\ell^{(o)};\mathcal{E}_n)\}_{\ell=0}^{T_o}\}$.

\paragraph{Local Hamiltonian.}
\[
H_o(q_o,p_o;\mathcal{E}_n,\eta_n)
=\tfrac12 p_o^\top M_o^{-1}p_o+\mathcal{R}_o(q_o;\mathcal{E}_n,\eta_n),
\quad
\mathcal{R}_o=\lambda_n E_{\mathrm{obj}}+\sum_{i\in\mathcal{C}_n}\alpha_{n,i}\,b(d_i;\hat d),
\]
with no dissipation/ports on the shape subsystem (i.e., $\Gamma_o=0$, $u_o=0$).

\subsection{Hamiltonian Assembly and Discrete Integration}
\label{subsec:symplectic_integration}

\paragraph{Energy terms.}
We use the surrogate energy and Hamiltonian in ~\cref{eq:impl:surrogate_energy,eq:impl:surrogate_hamiltonian}.

\paragraph{One-step update.}
We implement the symplectic Euler step in ~\cref{eq:impl:ph_euler_p,eq:impl:ph_euler_q}, using gradients from
~\cref{eq:impl:gradients} and frame-only selectors from ~\cref{eq:impl:block_selectors}.

\subsection{Observable Extraction}
\label{subsec:observable_extraction}

We compute observables using ~\cref{eq:impl:observables,eq:impl:observable_vector} and compare against the target
in ~\cref{eq:impl:observable_target}.

\subsection{Parameter Adaptation via Observable Feedback}
\label{subsec:parameter_adaptation}

We form the active parameter vector via ~\cref{eq:impl:active_param_vector}, update $J_n$ via ~\cref{eq:impl:secant_J},
and solve the projected Tikhonov update via ~\cref{eq:impl:tikhonov_update}. Inactive barrier weights $\alpha_{n,j}$ for
$j\notin\mathcal{I}_n$ remain unchanged.

\subsection{Port Correction from Residual}
\label{subsec:port_correction}

We compute the residual via ~\cref{eq:impl:residual}, update $P_n$ via ~\cref{eq:impl:secant_P}, and solve the port LS with
clipping via ~\cref{eq:impl:port_update}. A robust fallback is to set $P_n=\mathrm{diag}(0,0,\kappa_v)$ (ports primarily regulate speed).

\subsection{Output Format}
\label{subsec:output_format}

\paragraph{Trajectory and parameters.}
Return $\{z_n\}_{n=0}^{N}$ and the history $\{(\eta_n,\mu_n,u_{f,n})\}_{n=0}^{N}$ where $N$ is the final step. This enables post-hoc
inspection of how $(\beta,\lambda,\alpha_i)$ and $\mu$ evolve with the active set in ~\cref{eq:impl:active_set}, and when nonconservative
port forcing is invoked via ~\cref{eq:impl:port_update}.

\subsection{Additional hyperparameters in Table~\ref{tab:hyperparams}}
\label{subsec:hyperparam}

As a complement to Algorithm~\ref{algo:online_grl_snam}, we specify additional fixed parameters and training hyperparameters below.

\paragraph{Surrogate training (offline).}
The surrogate coefficient network $g_\xi$ is trained on short rollouts of horizon $H \in \{2,\ldots,6\}$ steps with loss
\[
\mathcal{L} = w_{\mathrm{traj}}\|o_H - o_H^*\|^2 + w_{\mathrm{vel}}\|v_H - v_H^*\|^2 + w_{\mathrm{fric}}\|\gamma - \gamma_o\|^2 + w_{\mathrm{multi}}\mathcal{L}_{\mathrm{ms}},
\]
where $(w_{\mathrm{traj}}, w_{\mathrm{vel}}, w_{\mathrm{fric}}, w_{\mathrm{multi}}) = (1.0, 1.0, 0.1, 0.5)$ and $\mathcal{L}_{\mathrm{ms}}$ is a multi-start robustness penalty sampling perturbed initial conditions near obstacles. We use learning rate $\eta \in [10^{-4}, 3 \times 10^{-4}]$ with Adam and gradient clipping at norm $5.0$.

\paragraph{IPC barrier.}
The log-barrier $b(d; \hat{d}) = -(d - \hat{d})^2 \log(d/\hat{d})$ activates at distance $\hat{d} \in [0.28, 1.0]$. Barrier values and gradients are clamped to $[0, 200]$ and $[-200, 200]$ respectively for numerical stability.

\paragraph{Stage planning.}
Stages have dimensions $(W_{\mathrm{stage}}, H_{\mathrm{stage}}) = (2.6, 2.0)$ with overlap ratio $\rho_{\mathrm{overlap}} = 0.3$. Obstacles are inflated by $r_{\mathrm{inflate}} \in [0.05, 0.4]$ to define the collision-free tube for widest-path exit selection. The margin factor for radius-aware integration is $0.5$.

\paragraph{Hyperelastic ring experiment.}
\emph{Dynamics:} The ring uses mass matrix $M = \mathrm{blkdiag}(M_o I_2, I, M_s)$ with $(M_o, I, M_s) = (1.5, 0.6, 1.0)$ and nominal damping $(\gamma_o, \gamma_\theta, \gamma_s) = (4.0, 1.6, 2.0)$. Area resistance uses bulk modulus $k_{\mathrm{bulk}} = 1.5$. 
\emph{Geometry:} The periodic cubic B-spline boundary has $n_{\mathrm{ctrl}} = 20$ control points and $K = 240$ curve samples, with nominal robot radius $r_{\mathrm{robot}} \in [0.30, 0.50]$.
\emph{Integration:} Symplectic Euler with timestep $\tau = 0.03\,\mathrm{s}$.

\paragraph{Dungeon navigation experiment.}
\emph{Dataset generation:} Maximum velocity $v_{\max} \in [2.5, 3.0]\,\mathrm{px/s}$, goal attraction gain $k_{\mathrm{goal}} \in [20, 25]$, barrier repulsion gain $k_{\mathrm{barrier}} \in [0, 1]$ (often $0$ when SDF provides implicit avoidance), and barrier activation $\hat{d} = 1.0\,\mathrm{px}$.
\emph{Integration:} Explicit Euler with timestep $\tau = 0.3\,\mathrm{s}$ and $4$ substeps for collision resolution.
\emph{Environment:} Obstacle distances are queried from a precomputed signed distance field $\mathrm{SDF}(x,y)$ via Euclidean distance transform. Stages are square windows of side $s_{\mathrm{stage}} = 60\,\mathrm{px}$, with waypoints inflated to maintain wall clearance $d_{\mathrm{wp}} \geq 1.5$--$4.0\,\mathrm{px}$.

\paragraph{Test-time adaptation.}
When enabled, online finetuning uses proximal regularization $\lambda_{\mathrm{prox}} = 10^{-3}$ anchored to the pretrained checkpoint, updating only the final projection heads. The secant-based controller selects $k_\alpha \in \{1, 2\}$ nearest obstacles for local $\alpha$-adjustment with per-head learning rates $(\kappa_\beta, \kappa_\gamma, \kappa_\alpha) = (0.25, 0.05, 0.4)$.

\end{document}